\documentstyle[12pt,psfig,twoside,suthesis]{report}

\newcommand{\given}{\mbox{ $|$ }}
\newcommand{\beqn}{\begin{equation}}
\newcommand{\eeqn}{\end{equation}}
\newcommand{\argmin}{\mbox{argmin}}

\begin{document}
\title{Natural Language Parsing\\ as\\ Statistical Pattern Recognition}
\author{David M. Magerman}
\principaladviser{Vaughan Pratt}
\firstreader{Nils Nilsson}
\secondreader{Jerry Hobbs\\(Dept. of Linguistics)}
\submitdate{February 1994}
\copyrightyear{1994}

\beforepreface
\prefacesection{Preface}

The debate between the relative merits of statistical modeling and
linguistic theory in natural language processing has been raging since
the days of Zellig Harris and his irreverent student, Noam Chomsky.  I
have never been shy about expressing my views on this issue.  I would
have liked nothing more than to declare in my dissertation that
linguistics can be completely replaced by statistical analysis of
corpora.  In fact, I intended this to be my thesis: given a corpus
parsed according to a consistent scheme, a statistical model can be
trained, {\em without the aid of a linguistics expert,} to annotate
new sentences with that same scheme.

The most important part of this thesis is that linguists need not
participate in the development of the statistical parser.  Using the
most obvious representations of the annotations in the parsed corpus,
the parser should automatically acquire disambiguation rules in the
form of probability distributions on parsing decisions.  In other
words, natural language parsing would be transformed from the
(never-ending) search for the perfect grammar into the simple task of
annotating enough sentences to train rich statistical
models.\footnote{This task is not so simple if ``enough sentences''
turns out to be, say, 10 trillion, but I will deal with that issue
later.}

With the guidance and support of the statistical modeling gurus in the
IBM Speech Recognition Group, I formulated and implemented a
statistical parser based on this thesis.  In experiments, it parsed a
large test set (1473 sentences) with a significantly higher accuracy
rate than a grammar-based parser developed by a highly-respected
grammarian.  The grammarian spent the better part of a decade
perfecting his grammar to maximize its score on the {\em
crossing-brackets} measure.\footnote{For the definition of the
crossing-brackets measure, see chapter~\ref{RESULTchapter}.} The
grammarian's score on this test set was 69\%.  The statistical parser,
trained on the same data to which the grammarian had access, scored
78\%.

While these results represent significant progress, they do not prove
my original thesis.  Despite the 78\% crossing-brackets score, only
about 35\% of the parses exactly matched the human annotations for
those sentences.  Ignoring part-of-speech tagging errors, just under
50\% of the parses were exactly correct.  All this means, of course,
is that I didn't solve the natural language parsing problem.  No big
surprise there (although the naive graduate student in me is a little
disappointed.)  But, in analyzing where the parser fails, there is a
glimmer of hope for a better solution.

Diagnosing parsing errors, especially in a parser that has hundreds of
thousands of parameters, is a tricky business.  But I suspect the main
problem with the parser is the lack of linguistic sophistication in
the disambiguation criteria made available to the statistical
models.\footnote{Actually, with 10,000 times more data, the parser
probably could have gotten by with the simplistic representations.
But if we could efficiently collect a half billion human-annotated
sentences, we probably wouldn't need automatic parsers.  We could just
use the annotators.  The discussion assumes we are limited to an
amount of data that we can reasonably collect for a new domain.} For
example, there was no morphological component to the parser.
Distributionally-determined word features provide some morphological
clues; however, these features are only reliable for the higher
frequency words.  As a result, conclusions about disambiguation for
some singular nouns were not carried over to their plural forms.  Some
untensed verbs were not related to their tensed counterparts,
preventing the parser from drawing conclusions about some attachment
decisions.

This error analysis leads me to conclude that linguistic input is
crucial to natural language parsing, but in a way much different than
it is currently being used.  Humans, as language processing experts,
are much more capable of identifying disambiguation criteria than they
are at figuring out how to apply them.  Grammarians can contribute to
statistical parser development not by writing large and unwieldy rule
bases, but instead by identifying the criteria by which a parser might
make disambiguation decisions.

\prefacesection{Acknowledgements}

Although I only began my graduate studies three and a half years ago,
the path to completing my doctoral thesis began much earlier.  There
are many people who contributed to this process, both technically and
personally.

First, I would like to thank my advisor, Dr. Vaughan Pratt, for
supporting me while I was at Stanford.  Dr. Pratt's master's thesis
was on a form of probabilistic natural language parsing, but since
completing his master's thesis work in the early 1970's, his research
has moved in many different directions.  I am sure there were many
other Stanford graduate students doing work which contributed more to
his current research projects, but he supported me, both technically
and financially, throughout my graduate studies.  If he had not
offered to act as my advisor after my first-year at Stanford, I would
not have lasted at Stanford very long.

I would also like to thank the other reading committee members, Drs.
Jerry Hobbs and Nils Nilsson, and the other friends and colleagues who
took the time to read the many drafts of my dissertation and who gave
me much-needed feedback: John Gillett, Dr. Ezra Black, and Adwait
Ratnaparkhi.

None of the work presented in this dissertation could have ever been
accomplished without the technical contributions of the IBM Language
Modeling Group.  Drs. Fred Jelinek, Bob Mercer, and Salim Roukos, the
managers in the language modeling group during the time I performed my
thesis research, contributed a great number of ideas to my research.
During the summer of 1992, Drs. Jelinek, Mercer, and Roukos, along
with Dr. John Lafferty, Adwait Ratnaparkhi, and Barbara Gates met with
me twice weekly to discuss progress in the development of SPATTER and
to brainstorm solutions to problems.  I greatly appreciate the
contributions of all of the members of the Language Modeling
Group.\footnote{This work was supported by a grant awarded jointly to
the IBM Language Modeling Group and the University of Pennsylvania
Computer Science Department (ONR contract No.  N00014-92-C-0189).}

Dr. Ezra Black, an extraordinary linguist and a good friend, did not
attend these meetings even though he was also a member of the IBM
Language Modeling Group during my time at IBM.  His parser was used as
a benchmark against which my thesis parser would be evaluated, and it
seemed inappropriate for him to contribute directly to the development
of the parser.  Nonetheless, he contributed to my thesis in many ways.
The design of my parser was based on that of his feature-based
grammar.  And he was always available to me to answer my naive
questions about language and the Lancaster treebank.  Even though he
didn't believe in my approach to the problem we were both trying
feverishly to solve, he never allowed his reservations to interfere
with his explanations.  And, in the end, I discovered that he was
right about many of the points on which we disagreed.

While I may have a degree in mathematics, and I've taken a course or
two in probability and statistics, when I arrived at IBM, I was quite
unskilled when it came to statistical modeling.  I would have remained
that way were it not for the hours and hours which the IBM Speech
Group members devoted to my education.  Drs. Stephen Della Pietra,
Vincent Della Pietra, and John Lafferty seemed to be on call 24 hours
a day to answer any and all of my questions about statistical
modeling.  And, they didn't just answer my questions, but also proved
their answers and explained the proofs until I understood them.  I owe
nearly all that I know about statistical modeling to their knowledge
and patience.

Although my technical knowledge developed at IBM, my interest in
natural language parsing began during my undergraduate days at the
University of Pennsylvania, under the tutelage of Dr. Mitch Marcus.  I
am greatly indebted to Dr. Marcus for introducing me to the NLP field,
and for teaching me the fundamentals of parsing.  I also thank him for
breaking the barriers imposed by my undergraduate status and allowing
me the opportunity to meet and discuss my work with the leading
researchers in statistical NLP, like Drs. Kenneth Church, Fernando
Pereira, and Don Hindle at Bell Laboratories, Dr. Stuart Shieber at
Harvard University and Dr. Fred Jelinek at IBM.

All of the researchers mentioned above contributed to my technical
knowledge of the field, and many of them supported me emotionally as
well during my undergraduate and graduate studies.  But the man to
whom I am most indebted for starting me down the path which has led to
this work is Dr. Max Mintz.  I first met Dr. Mintz as a student of his
in an undergraduate computer class.  After a class in the middle of
the semester, he invited me to his office.  For some reason, unknown
to me to this day, he offered me a deal no undergraduate computer
science student could turn down.  I needed only to name a research
area, any area I desired, and he would make sure I had an opportunity
to become involved with the research group at Penn working in that
area.  I am sure he hoped I would express interest in his robotics and
vision group.  But, instead of leading me toward this choice, he
deliberately biased me against his own group, so as not to appear to
be taking advantage of an impressionable youth such as I was.  In the
end, I chose natural language processing, and, true to his word, he
introduced me to the head of the Language and Information Computing
lab, Dr. Mitch Marcus.  Even though I was not working directly with
him, Dr. Mintz became my academic advisor, my mentor, and my friend.
Any problem I had, no matter how trivial or how difficult, he would
always drop what he was doing and try to solve it (and he almost
always succeeded).  I only hope that the work I am reporting in this
dissertation meets with his approval.

I would also like to thank my mother, my father (may he rest in
peace), and my sister for loving me, for supporting me, and for
encouraging me to be the best I could be in whatever endeavor I chose
to pursue.

Finally, I would like to thank my roommate for two of the years I was
at Stanford, fellow graduate student, lecturer, and my best friend,
Raymond Suke Flournoy.  Without Ray, I would surely have gone insane
long before I completed my dissertation.  I suspect that, without me,
he would be much closer to finishing his own dissertation, but I hope
he doesn't hold that against me.

\afterpreface

\chapter{Introduction}

{\em Automatic natural language (NL) parsing} is a central problem to
many natural language processing tasks.  The task of automatic NL
parsing is to design a computer program which identifies the
hierarchical constituent structure in a sentence.  In the early years
of artificial intelligence work, it was believed that this was a
relatively simple problem which would be solved quickly.
That was 30 years ago, and the problem is still a
thorn in the NL processing community's collective side.

Why is parsing a natural language so difficult?  The short answer is
simply: ambiguity.  A natural language sentence takes on different
meanings, depending on its context, the speaker, and many other
factors.  Ambiguity takes on many different forms in NL, such as
semantic ambiguity, syntactic ambiguity, ambiguity of pronominal
reference, to name just a few.

On closer inspection, the ambiguity resolution problem can be restated
as a {\em classification} problem.  Consider the prepositional phrase
attachment decisions in the following sentences:
\begin{verbatim} Print the file in the buffer. \end{verbatim}
\begin{verbatim} Print the file on the printer. \end{verbatim}
In these cases, the prepositional phrase can be attached to either the
nearest noun phrase, as in the first example, or to the higher verb
phrase, as in the second example:
\begin{verbatim} [V Print [N [N the file N] [P in the buffer P] N] V].
\end{verbatim}
\begin{verbatim} [V Print [N the file N] [P on the printer P] V].
\end{verbatim}
Given the entire sentence and perhaps the entire dialogue as context,
the parser's job is to {\em classify} the context as either one which
dictates the low N attachment or the high V attachment.

Traditionally, disambiguation problems in parsing have been solved by
enumerating possibilities and explicitly declaring knowledge which
might aid the disambiguation process.  This declarative knowledge
takes the form of semantic restrictions (e.g.\ Hirschman et.al.
\cite{PUNDIT}), free-form logical expressions (e.g.\ Alshawi et.al.
\cite{CLE}), or a combination of these methods (e.g.\ Black, Garside,
and Leech \cite{blackbook}).  Some have used probabilistic (e.g.\
Seneff \cite{Seneff:89}) or non-probabilistic (e.g.\ Hobbs et.al.
\cite{DIALOGIC}) weighting systems to accumulate disambiguation
decisions throughout the processing of a sentence into a single score
for each interpretation.

Each of these approaches has resulted in some degree of success in
accurately parsing sentences.  However, they all depend on the
intelligence and expertise of their developers to discover and
enumerate the specific rules or weights which achieve their results.
Most (if not all) of these systems were developed by a grammarian or
language expert examining sentence after sentence, modifying their
rules in some way to account for parsing errors or new phenomena.
This development process can take years, and there is no reason to
believe that the process ever converges.  Rule changes for new
sentences might undo fixes for old sentences, in effect causing more
recent sentences to take precedence over older ones.  And, most
important, there is no systematic way to reproduce this process.  A
researcher trying to reproduce these parsing results from scratch
would have no algorithm or systematic procedure to follow to discover
the same rules or weights, except for, perhaps, ``Look at a lot of
sentences.''

\section{Statement of Thesis}

This work addresses the problem of automatically discovering the
disambiguation criteria for all of the decisions made during the
parsing process.  These criteria can be discovered by collecting
statistical information from a corpus of parsed text.  Given the set
of possible features which can act as disambiguators, an
information-theoretic classification algorithm based on the contexts
of each decision made in the process of constructing a parse tree can
learn the criteria by which different disambiguation decisions should
be made.  Each candidate feature is a question about the context which
has a discrete, finite-valued answer.

The claim of this work is that statistics from a large corpus of
parsed sentences combined with information-theoretic classification
and training algorithms can produce an accurate natural language
parser without the aid of a complicated knowledge base or grammar.
This claim is justified by constructing a parser based on very limited
linguistic information, and comparing its performance to a
state-of-the-art grammar-based parser on a common task.

In this work, parsing is not viewed as the recursive application of
predetermined rewrite rules.  The parser developed for this work uses
a feature-based representation for the parse tree, decomposing the
parse tree into the words in the sentence, the part-of-speech tags for
each word, the constituent labels assigned to each node in the parse,
and the edges which connect the tree nodes.  Given the words as input,
statistical models are trained to predict each of the remaining
features.  A parse tree is constructed by generating values for each
of the features, one at a time, according to the distributions
assigned by the models.  Once a feature value is generated, that value
can be taken into account when determining the distributions of future
feature value assignments.

Each feature value assignment decision is modeled by a statistical
decision tree, which estimates the probability of each alternative
given the context.  Since the probability distribution of each
decision is conditioned on the entire context of the partial parse,
the order in which decisions are made affects the probability of the
parse.  Thus, the total probability of a parse tree is the sum over
all possible derivations of that parse tree, and the probability of a
derivation of a parse tree is the product of the probabilities of the
atomic decisions which resulted in the construction of the parse tree.

The decision tree models used in this work are constructed using the
CART algorithm as discussed in (Breiman et.al.), based on the counts
from a training corpus.  Then, an expectation-maximization (E-M)
algorithm is used to train the hidden derivation model, assigning
weights to the different derivations of the parse trees in the
training corpus, in order to maximize the total probability of the
corpus.  The resulting model is further improved by smoothing the
decision trees using another E-M algorithm, this time training hidden
parameters in the decision trees, maximizing the probability of the
parse trees in a new, held-out corpus.

One of the important points of this work is that statistical models of
natural language do not need to be restricted to simple,
context-insensitive n-gram models.  In fact, it should be clear that
in a problem like parsing, where long-distance lexical information is
crucial to disambiguate interpretations accurately, local models like
P-CFGs (probabilistic context-free grammars) or n-gram models are
insufficient.  And while it has been assumed that one could not
accurately train statistical models which consider large amounts of
contextual information using the limited amount of training data
currently available, this work illustrates that existing decision tree
technology can generate models which selectively choose elements of
the context which contribute to disambiguation decisions, and which
have few enough parameters to be trained using existing resources.

\section{Organization of Dissertation}

In Chapter~\ref{SURVEYchapter}, I attempt to put this work in the
context of previous work on statistical and non-statistical natural
language processing.  Then I introduce decision tree modeling in
Chapter~\ref{DTREEchapter}, describing the algorithms used in growing
and training decision trees and discussing some of the open questions
involved in decision tree modeling.
In Chapter~\ref{PRELIMchapter}, I report on some preliminary
experiments which explored the effectiveness of using
context-sensitive models and decision tree models in statistical
parsing.  Chapter~\ref{SPATTERchapter} introduces my thesis parser,
called SPATTER, defining the representations used in the parser and
stepping through the decoding algorithm.  Chapter~\ref{MODELchapter}
presents the specific models used in SPATTER and describes the
training process.  In Chapter~\ref{EVALUATIONchapter}, I discuss a few
methodological issues involved in evaluating the performance of
natural language parsers.  Experimental results using SPATTER follow
in Chapter~\ref{RESULTchapter}.  After exploring questions left
unanswered by this work in Chapter~\ref{OPENchapter}, I offer some
concluding remarks in Chapter~\ref{CONCchapter}.


\chapter{Related Work\label{SURVEYchapter}}

In many respects, the natural language processing task is the holy
grail of the artificial intelligence community.  It was one of the
earliest AI problems attempted, and its solution is one of the most
elusive.  The subtleties involved in understanding natural language,
from dealing with anaphora and quantifier scope to recognizing sarcasm
and humor, preclude superficial and knowledge-bereft solutions, which
characterize the majority of the early work in natural language
processing.

Automatic natural language parsing, as defined earlier, is a critical
component of any solution to the natural language understanding
problem.  Recent work in information extraction and text processing has
substituted finite-state pattern matching machinery for parsing
technology with great success.  However, these applications do not
require ``understanding'' as much as the identification and
regurgitation of critical information in a passage.  Disambiguation
decisions are less important to these tasks, since a program can make
many interpretation errors in a text and still correctly answer the
questions required by the task.  However, for a program to detect
subtle language usage and to use all of the information gained from a
text in intelligent activities, the complete disambiguation
capabilities of a parser are necessary.

In this chapter, I briefly survey early work in natural language
parsing, tracing the progression of the techniques employed.  Then I
discuss the paradigm shift in the speech recognition community in the
late 1970s from rule-based methods to statistical modeling, and the
impact of this paradigm shift on natural language processing in the
late 1980s.  Next I survey more recent work on the problem of
broad-coverage natural language parsing.  Finally, I discuss the
development of decision tree modeling, from early AI machine learning
to current speech recognition and natural language applications.

\section{Early Natural Language Parsing}

Automatic natural language processing research can be traced back to
the early 1950s, to Weaver's early work on machine translation (MT)
\cite{Weaver}.  The failure of superficial ``dictionary lookup''
solutions to the MT problem suggested the need for a higher level of
knowledge representation.

The work in the 1960s on natural language processing consisted
primarily of keyword analysis or pattern matching.  Systems such as
Green's BASEBALL\cite{BASEBALL}, Raphael's SIR\cite{SIR}, and Bobrow's
STUDENT\cite{STUDENT} search for simple patterns or regular
expressions which indicate useful information.  All information in the
text which does not conform to these patterns is ignored.  This
attribute makes pattern-matching systems more robust, but it also
makes them easy to identify, as they will happily process gibberish as
long as some subset of the input matches a known pattern.
Weizenbaum's ELIZA\cite{ELIZA} is a famous example of this
``technology,'' reviled in some corners of the community for falsely
encouraging the already widely-held belief that natural language
processing would be solved within a decade.

Chomsky's work in the late 1950s and early 1960s in transformational
grammars and formal language theory \cite{Chomsky1} \cite{Chomsky2}
provided much of the machinery for the next generation of natural
language processing research.  Context-free grammar parsers, such as
Lindsay's SAD-SAM\cite{SADSAM}, took advantage of Chomsky's
formalization to improve upon the simpler single-state and
finite-state models.  The SAD component of this system generates full
syntactic analyses for sentences, accepting a vocabulary of about 1700
words and a subset of English grammar.  This approach suffered from
some of the same deficiencies that exist in systems today, in
particular the limited coverage of the grammar and vocabulary.

In contrast to the decomposition of syntax and semantics in SAD-SAM,
Halliday's systemic grammar \cite{Halliday} proposed a formalism that
encoded the functional relationships in a sentence.  His theory was
illustrated in Winograd's blocks-world system, SHRDLU\cite{SHRDLU}.
SHRDLU demonstrated the effectiveness of functional representations on
a small problem, but it also implicitly revealed one of its
weaknesses.  Systemic grammar works when applied to the very
constrained blocks-world because the relationships among objects and
the possible actions could be completely and unambiguously specified.
A small number of predicates describe all actions and relations in the
blocks-world.\footnote{This is a bit of an oversimplification.  SHRDLU
implemented a set of predicates which were {\em defined} to be the
blocks-world.  There were many gaps in the representation of the
blocks-world.  While one could claim that predicates could be added to
SHRDLU to fill these gaps, this is the same as claiming that a grammar
would work if one only added the correct rules.} However, this is far
from true in the real world, and it is a daunting (if not impossible)
task to represent even a small subset of this real world knowledge in
a useful way.

The development of Augmented Transition Networks (ATNs) by Woods in
the early 1970s \cite{Woods} improved upon the power of regular
expressions and context-free grammars by augmenting a finite-state
automaton with register variables and functional constraints, allowing
an ATN to consider more contextual information when generating an
analysis while maintaining the computational simplicity of a
finite-state machine.  However, the use of ATNs also encouraged ad-hoc
design methodology, where each new application required a new ATN, and
the solution to one processing task did not guarantee a solution to
any others.

\section{Computational Grammatical Formalisms}

Perhaps in response to the ad-hoc nature of ATNs, in the early 1980s a
number of grammatical formalisms appeared which attempted to account
for the power of the functional augmentations of ATNs in a more formal
theoretical framework: Definite-Clause Grammar (DCG) \cite{DCG},
Functional Unification Grammar (FUG) \cite{FUG}, Lexical-Functional
Grammar (LFG) \cite{LFG}, Generalized Phrase Structure Grammar (GPSG)
\cite{GPSG}, and others.

Although these theories differ in their approach to language
processing and representation, they all have one attribute in common:
they are not really linguistic theories as much as {\em computational}
linguistic theories.  Chomsky's Transformational Grammar is a
linguistic theory which one can implement, and ATNs are computational
devices which encode some linguistic knowledge; but these new theories
unite linguistic theory and computational elegance.

For the purposes of this dissertation, each of these theories can be
viewed as augmented phrase structure grammars, where the augmentations
represent the long-distance dependencies and the subtleties of
language which are required for analysis and disambiguation of
text.\footnote{Excellent descriptions of these theories can be found
in Sells \cite{Sells} and Shieber\cite{Shieber}.} But does the
theoretical formalization of the augmentations solve the natural
language parsing problem more effectively than the more ad-hoc ATNs?

This question has not been conclusively answered.  These theories
provide superior representation schemes which allow a grammarian to
represent more aspects of language more efficiently and effectively
than ATNs do.  But they do not appear to ``solve'' the problem of
natural language parsing.  Implementations of these theories on a
grand scale have shown themselves to suffer from the same deficiencies
as the earlier ATNs, albeit to a lesser extent: language usage is too
varied to be represented completely in a rule base, and each language
processing task presents new problems which the previous ``solutions''
do not solve.

\section{Broad-coverage Parsing Systems}

After discovering that the ``success'' of early natural language
processing work was fleeting, NL researchers expanded their efforts to
solving larger, more general problems.  This work involved building
broad-coverage parsers of general language, and tuning them to focus
on a particular domain.  These systems consisted of a set of core
language rules which applied to any domain, with rules and
restrictions added to aid disambiguation and analysis for a specific
domain.

Some examples of this type of system development are Unisys' PUNDIT
system \cite{PUNDIT} and NYU's PROTEUS system \cite{PROTEUS}.  Both of
these systems are descendants of the Linguistic String Project (LSP)
\cite{Sager}, an early effort to develop grammar-based parsers for
sublanguages.  Both systems use a string grammar, consisting of a
context-free grammar backbone augmented with functional restrictions
on the application of the grammar's productions.  Although the
grammars in these two systems are similar, they handle ambiguity
resolution in very different ways.  The PUNDIT system uses a
recursive-descent design strategy with backtracking for its entire
processing pipeline.  The first syntactic analysis which passes
through the semantic and pragmatic components of the system without
error is accepted.  Since the system does not generate parses in an
intelligent order, the best analysis will not necessarily be the first
one generated, and it is possible for the systems to select a
suboptimal analysis, as long as it has an acceptable semantic and
pragmatic interpretation.  The PROTEUS system, on the other hand, uses
a hand-generated weighting strategy to rank syntactic analyses.
Heuristic scoring functions implement various preference mechanisms,
including preferring the closest attachment, disfavoring headless noun
phrases, and evaluating semantic selection.  The parser uses a
best-first search strategy to discover the highest-scoring analysis.

SRI's TACITUS system \cite{TACITUS} is another descendant of the
Linguistic String Project.  It uses the DIALOGIC parser, which is a
union of the LSP grammar and the DIAGRAM grammar, a grammar developed
for SRI's speech understanding research.  DIALOGIC is similar to the
PROTEUS parser in that it uses a sorted agenda parsing algorithm with
weighting system for disambiguation.  DIALOGIC performs some pruning
as well, advancing only the highest scoring analyses at each point in
the parsing process.  For sentences longer than 60 words, DIALOGIC
performs ``terminal substring parsing,'' segmenting the sentences into
substrings and parsing these substrings independently and trying to
paste together the partial analyses.

The TACITUS and PROTEUS systems were designed for an information
extraction task that only required recognizing and understanding key
pieces of information in a document.  Realizing this, the developers
of these systems implemented mechanisms to use partial information in
the event that a sentence could not be completely analyzed by their
grammar.  TACITUS included a relevance filter which allowed the system
to ignore sentences which it deemed statistically ``irrelevant'' to
the information extraction task.

These systems gave way to more refined information extraction systems,
such as SRI's FASTUS system \cite{FASTUS}, which abandons the
grammar-based parsing strategy in favor of a finite-state machine
approach, specifying flexible templates for identifying the critical
information necessary for accomplishing the information extraction
task.  Similarly, the grammar-based systems with complete syntactic,
semantic and pragmatic analysis designed for spoken language
applications, such as SRI's Core Language Engine (CLE) \cite{CLE} and
MIT's VOYAGER system \cite{VOYAGER}, have been dominated by newer
finite-state template-based systems.  These template-based systems,
using essentially the same technology as exhibited in Schank's SAD
analyzer, benefits from a data-driven design methodology to achieve
better coverage and accuracy.

These and many other broad-coverage, domain-specific natural language
parsing systems reengineered existing technology, augmenting it with
better heuristic strategies, providing better coverage and performance
than previous implementations.  The research community recognized
that, for some applications, complete understanding was not as
important as robustness in terms of overall performance.  This is
especially the case in information extraction and database query tasks.
However, while these systems performed {\em better} than earlier ones,
they seem to have ignored the original problem of natural language
understanding, where the subtleties of language usage can not be
ignored.

\section{The Toy Problem Syndrome}

Why did the NL community become sidetracked from its goal of NL
understanding?  Early NL processing research suffered from what I call
the {\em Toy Problem Syndrome}.  The Toy Problem Syndrome arises from
trying to solve a general class of problems by examining only a
single, simple example of the class.  The result is a partial solution
to the problem which is limited in scope and extensibility.  For
instance, the keyword analysis and pattern matching programs solved a
small part of the NL processing problem, but provided no mechanism to
solve the remainder of it.  The early ATN-based and grammar-based
parsers were developed to handle very constrained problems, and while
these methods worked on the toy problems they were designed for, they
have not been shown to work on larger ones.  Further, the rule bases
for these early systems, especially those using ATNs or systemic
grammars, are so domain-specific that developers of new systems using
the same research paradigm must essentially start from scratch, even
though the natural language used, English, is the same.

The information extraction and database query tasks on which much of
the NL community is currently working are not toy problems, based on
the definition above.  They are large and difficult problems which
cannot be solved by simple hacks like ELIZA.  However, the technology
which has been developed for these tasks is especially tailored for
the specific application being implemented.  For instance, the SRI
Template Matcher requires a set of templates for a domain and a
mapping from these templates to a database query language.  The system
coverage and performance depends on the extent to which these
templates can be translated into database queries.  Porting this
system to a new domain requires essentially starting from scratch,
designing new templates and writing new mappings from templates to
query code.  For some domains, this task may be difficult or
impossible.

\subsection{The Speech Recognition Revolution}

The speech community confronted the Toy Problem Syndrome in dealing
with the speech recognition (SR) problem in the 1970s.  Their solution
serves as an excellent model for the NL parsing community to emulate.

In 1971, the Advanced Research Projects Agency (ARPA) of the Defense
Department asked five speech research groups to build demonstration
systems to solve a simple speech recognition task \cite{SUR}.  The
systems were expected to recognize a 1000-word vocabulary from a
constrained domain reasonably quickly with less than a 10\% error
rate.  No other aspects of the system were constrained.  The goal of
the project was to achieve a breakthrough in speech recognition
technology.

In fact, what resulted from the ARPA speech effort was an exercise in
ad-hoc engineering.  The most extreme example of this is the HARPY
system \cite{HARPY}, developed at Carnegie-Mellon University. The
HARPY system used a precompiled network which computed all possible
sentences which HARPY could expect to recognize.  While this solution
satisfied the letter of their ARPA contract, it certainly violated its
spirit.  The speech recognition ``technology'' in HARPY was a brute
force approach which falls apart if the vocabulary is increased and
the domain enlarged.  With the HARPY system, the speech community was
no closer to automatic speech recognition than before the project
began.  All of the ARPA-sponsored systems suffered from the Toy
Problem Syndrome.

In the late 1970s and early 1980s, the speech community took a giant
leap towards a general solution to the SR problem which avoided the
Toy Problem Syndrome.  This revolution in speech technology can be
traced back to a seminar given by researchers at IDA in October, 1980,
on Hidden Markov Models (HMMs) \cite{IDA}.  A Markov process is
finite-state process for which the probability of going from one state
to another on a given input depends only on a finite history.  Hidden
Markov models are statistical models of a Markov process, where some
component of the model is ``hidden,'' i.e. not explicitly represented
in the data.  The hidden component of these models can be learned in
an unsupervised mode using algorithms from information theory.  The
speech community, in particular the IBM Speech Recognition group and a
company called Verbex, recognized HMMs as a solution to a critical
problem in speech processing: modeling the intermediate form of speech
input.

At the time, the speech recognition problem had been broken down into
two steps.  This first step is called the acoustic modeling problem.
Here, spoken language waveforms, converted to a sequence of
real-valued vectors which mathematically encode the important
characteristics of the input, are translated into a sequence of
phonemes, the linguistic representation for the building blocks of
words.  This was accomplished by a variety of rule-based methods.  In
the second step, the language modeling problem, these phonemes are
combined to form word sequences, again using rule systems.

The critical problem with the early speech systems was the brittleness
of their acoustic and language models.  Composed of hand-generated
rules, these models might have be adequate to a handle a single
speaker using a limited-vocabulary language with a low
perplexity\footnote{Perplexity is a measure of the average number of
words which can appear at any point in a sentence.} grammar, but they
never scaled up to larger vocabularies and general human speech.
There is no theoretical reason why a rule-based system could not be
designed to solve the problem; but no system ever approached the level
of coverage needed for general large-vocabulary speaker-independent
speech recognition.

If researchers could not adequately encode phonetic representations by
hand, HMMs offered an alternative.  The speech input and the sentence
output are the only givens of the problem.  The intermediate
representation, the phonemes, can viewed as ``hidden,'' and the whole
process can be interpreted as a hidden Markov process.  Using the
expectation-maximization algorithm from information theory, the
classes of ``phonemes'' can be discovered automatically instead of
encoded by hand.  In other words, information theory provides
techniques which, given written and spoken versions of the same text,
can generate statistical models for recognizing speech.  Porting this
technology to new domains and new speakers simply requires retraining
the models using text from the domain read by a speaker.  Eventually,
algorithms were perfected to combine speech from different speakers to
allow speaker-independent recognition.

Certainly HMMs are not a panacea.  The key issue in applying Markov
models to a problem is to determine if they are Markov processes.
Even if they are not, as long as the process depends mostly on the
most recent history then it is possible to represent a process
approximately using a Markov model.  However, for some problems, this
is not the case, as I illustrate later in the case of probabilistic
context-free grammars.  But it was found that the speech recognition
task {\em could} be reformulated as a Markov process, and this
reformulation soon led to a reliable solution to the general problem
of recognizing spoken language.

\section{Recent Work in Statistical NL}

Preliminary experiments in statistical and corpus-based NL parsing
have already begun to follow in the footsteps of the SR community.
This work has focused on syntactic analysis, such as part-of-speech
tagging and grammar induction, but some projects have begun involving
probabilistic understanding models and statistical machine
translation as well.

\subsection{Part-of-speech Tagging}

Statistical part-of-speech tagging has been a hot topic since the 1988
ACL paper by Church on HMM tagging \cite{Church:88}.  The problem in
part-of-speech tagging is to assign to each word in a sentence a
part-of-speech label which indicates the linguistic category (e.g.
noun, verb, adjective, etc.)  to which that word belongs in the
context of the sentence.  Some part-of-speech tag sets only have a few
dozen coarse distinctions, while others include hundreds of
categories, distinguishing temporal, mass, and location nouns, as well
as indicating the tenses and moods of verbs.  Actually, HMM tagging
was suggested a few years earlier during a lecture by Mercer at MIT,
which Church attended, and Merialdo published a more obscure paper on
the subject in 1986 \cite{Merialdo}.  At BBN, Weischedel
\cite{Weischedel} explored the behavior of HMM tagging algorithms when
trained on limited data, and reports experimental results using
various models designed to account for weaknesses of the simple HMM
trigram word-tag model.

Lafferty \cite{Lafferty:92} uses decision tree techniques similar to
those described in Chapter~\ref{DTREEchapter} in his paper on decision
tree part-of-speech tagging.  His work was an attempt to extend the
usual three-word window made available to a trigram part-of-speech
tagger.  By allowing a decision tree to select from a larger window
those features of the context which are relevant to tagging decisions,
he hoped to generate a more accurate model using the same number of
parameters as a trigram model.  His results, however, were not much
better than those of existing taggers.

Brill's dissertation work \cite{Brilldiss} explored using a corpus to
acquire a rule-based tagger automatically.  His tagger preprocessed a
corpus using a simple HMM tagger and, based on the correct tagging
provided by human taggers, learned a small set of rules which
corrected the output of the HMM tagger.  The tagger considered a
limited class of possible rules, and thus could explore the space of
rules completely, proposing only those that improved the overall
accuracy of the tagger on a sample corpus.

\subsection{Grammar Induction}

Much of the work in grammar induction has been a function of the
availability of parsed and unparsed corpora.  For instance, in 1990, I
published my undergraduate thesis with Marcus \cite{Magerman:90} on
parsing without a grammar using mutual information statistics from a
tagged corpus.  I originally intended to do this work on supervised
learning from a pre-parsed corpus, but no such corpus existed in the
public domain.  Thus, the earliest work on grammar induction involved
either completely unsupervised learning, or, using the Tagged Brown
Corpus, learning from a corpus tagged for parts of speech.


Following the same path of the speech community, a number of parsing
researchers (e.g. Black et.al. \cite{blackgram} \cite{blackbook},
Kupiec \cite{Kupiec}, and Schabes and Pereira \cite{Schabes:92}), have
applied the inside-outside algorithm, a special case of the
expectation-maximization algorithm for CFGs, to probabilistic
context-free grammar (P-CFG) estimation.  A P-CFG is a context-free
grammar with probabilities assigned to each production in the grammar,
where the probability assigned to a production,
$X\rightarrow~Y_1\ldots~Y_n,$ represents the probability that the
non-terminal category $X$ is rewritten as $Y_1\ldots~Y_n$ in the parse
of a sentence.

P-CFGs have been around since at least the early 1970s (e.g.
Pratt\cite{Pratt}); but using very large corpora and the
inside-outside algorithm, they can now be trained automatically,
instead of assigning the parameters by hand.  A P-CFG model can be
trained in a completely unsupervised mode, by considering all possible
parses of the sentences in a training corpus (e.g.
Baker\cite{Baker:75} and Kupiec \cite{Kupiec}), or it can be trained
in a constrained mode, maximizing the probability of the parse trees
in a parsed corpus (e.g. Black et.al. \cite{blackgram} and Schabes and
Pereira\cite{Schabes:92}).

More evidence that the availability of corpora has influenced the
direction of research is in the study of parsing tagged sentences
using statistical methods (Magerman and Marcus\cite{Magerman:90},
Brill \cite{Brill}, and Bod\cite{Bod}).  The motivation behind this
work was two-fold.  First, since part-of-speech taggers and tagged
corpora were readily available, it seemed reasonable to attempt to
parse a tagged corpus, under the assumption that part-of-speech
taggers would eventually be accurate enough to be used as automatic
pre-processors for text.  Second, statistical parsing based on words
required more sophisticated training methods than statistical parsing
based on tags, since analyzing words required estimating far more
parameters with the same amount of data.

Neither of these motivations is very compelling.  In order to use a
part-of-speech tagger as a pre-processor for a parser, the tagger must
be able to make disambiguation decisions which existing parsers cannot
make accurately.  Also, in the past few years, the technology for
estimating probability distributions in the face of sparse data has
been well-documented.  Finally, with greater access to very large
parsed corpora, training parsing models is possible even with more
direct estimation methods.

\subsection{Other Work in Statistical NL}

Statistical natural language research extends far beyond tagging and
parsing.  Work on language acquisition attempts to discover semantic
selection preferences (e.g. Resnik\cite{Resnik}) and verb
subcategorization information (e.g. Brent \cite{Brent:91}).  Schuetze
\cite{Schuetze} has developed a vector-based representation for
language which aids in word sense disambiguation.

A significant application of statistical modeling technology is the
Candide system \cite{Candide}, developed by the IBM Machine
Translation group.  Candide translates French to English using a
source-channel model, where it is assumed that the French sentence was
actually originally an English sentence passed through a noisy
channel.  The job of the system is to decode the message, i.e. recover
the English sentence that was intended by the French code.  While this
model may offend the francophile, it has resulted in a
state-of-the-art translation system.

Following this model, researchers at BBN are working on generating
semantic analyses for sentences using statistical models.  Their
Probabilistic Language Understanding Model \cite{PLUM} defines a
semantic language and attempts to translate the natural language
sentence into the semantic language.  A system developed at CRIN, a
Canadian natural language company, takes a similar approach to the
problem of database retrieval, using the SQL database query language
as their semantic language.

\def\argmin{\mathop{\rm arg\,min}}
\newcommand{\Prob}{\mbox{Pr}}

\newcommand{\Dtree}{{\cal D}}
\chapter{Statistical Decision Tree Modeling\label{DTREEchapter}}

Much of the work in this thesis depends on replacing human
decision-making skills with automatic decision-making algorithms.  The
decisions under consideration involve identifying constituents and
constituent labels in natural language sentences.  Grammarians, the
human decision-makers in parsing, solve this problem by enumerating
the features of a sentence which affect the disambiguation decisions
and indicating which parse to select based on the feature values.  The
grammarian is accomplishing two critical tasks: identifying the
features which are relevant to each decision, and deciding which
choice to select based on the values of the relevant features.

Statistical decision tree (SDT) classification algorithms account for
both of these tasks.  SDTs can be used to make decisions by asking
questions about the situation in order to determine what the best
course of action is to take, and with what probability it is the
correct decision.  For example, in the case of medical diagnosis, a
decision tree can ask questions about a patient's vital signs and test
results, and can propose possible diagnoses based on the answers to
those questions.  And, using a set of patient records which indicate
the correct diagnosis in each case, the SDT can estimate the
probability that its diagnosis is correct.  For a particular
decision-making problem, the SDT growing algorithm identifies the
features about the input which help predict the correct decision to
make.  Based on the answers to the questions which it asks, the
decision tree assigns each input to a class indicating the probability
distribution over the possible choices.

SDTs accomplish a third task which grammarians classically find
difficult.  By assigning a probability distribution to the possible
choices, the SDT provides a ranking system which not only specifies
the order of preference for the possible choices but also gives a
measure of the relative likelihood that each choice is the one which
should be selected.  A large problem composed of a sequence of
non-independent decisions, like the parsing problem, can be modeled by
a sequence of applications of a statistical decision tree model
conditioned on the previous choices.  Using Bayes' Theorem to combine
the probabilities of each decision, the model assigns a distribution
to the sequence of choices without making any explicit independence
assumptions.  Inappropriate independence assumptions, such as those
made in P-CFG models, seriously handicap statistical methods.

The decision tree algorithms used in this work were developed over the
past 15 years by the IBM Speech Recognition group.  The growing
algorithm is an adaptation of the CART algorithm in Breiman
et.al.\cite{CART}.  The IBM growing and smoothing algorithm were first
published in Lucassen's 1983 dissertation \cite{Lucassen}.  Bahl,
et.al., \cite{treelm} is an excellent discussion of these algorithms
applied to the language modeling problem.  For this dissertation, I
explored variations of these algorithms to improve the performance of
the decision trees on the parsing task.

In this chapter, I first introduce some terminology and concepts from
information theory which are used in decision tree modeling.  Next, I
discuss some representational issues involved in decision tree
parsing.  Then I present the basic decision tree algorithms, along
with the variations I used in my experiments.  Finally, I address some
of the problems associated with maximum-likelihood (M-L) decision tree
training.

\section{Information Theory}

The algorithms for growing and smoothing decision trees depend upon
the quantification of information.  Information theory, developed by
Shannon\cite{Shannon} and Wiener\cite{Wiener}, is concerned with the
compression of information when transmitted through a channel.
Information theory formalizes the notion of information in terms of
entropy.  In this section, I introduce some basic concepts from
information theory which are necessary to understand decision trees.
A more complete introduction to information theory can be found in
Cover and Thomas\cite{Cover}.

\subsection{Entropy}

Entropy is a measure of uncertainty about a random variable.  If a
decision, or random variable, $X$ occurs with a probability
distribution $p(x),$ then the entropy $H(X)$ of that event is
defined by
\beqn H(X) = -\sum_{x\in{\cal X}} p(x) \log_2 p(x). \eeqn
Since $x\log_2~x\rightarrow~0$ as $x~\rightarrow~0,$ it is
conventional to use the relation $0\log_2~0=0$ when computing entropy.

The units of entropy are bits of information.  This is because the
entropy of a random variable corresponds to the average number of bits
per event needed to encode a typical sequence of events sampled from
that random variable's distribution.

Consider how entropy behaves in extreme cases.  For instance, if a
random variable is uniformly distributed, i.e.
$p(x)=p=\frac{1}{|{\cal X}|},$ then
\beqn H(X) = -\sum_{x\in{\cal X}} p \log_2 p = -\log_2 p = \log_2 |{\cal X}|.
\eeqn
This is the case of maximum uncertainty, and thus maximum entropy.  At
the other extreme, when all of the probability mass is on one element
of ${\cal X},$ say $\hat{x},$ then $p(\hat{x})=1$ and $p(x)=0$ for all
$x\not=\hat{x}.$ Since $1\log_2~1=0$ and $0\log_2~0=0,$ then
\beqn H(X) = -\sum_{x\in{\cal X}} p(x) \log_2 p(x) = 0.\eeqn
This is the case of minimum entropy, since there is no uncertainty
about the future; $X$ takes on the value $\hat{x}$ every time.

\subsection{Perplexity}

Perplexity is a measure of the average number of possible choices
there are for a random variable.  The perplexity of a random variable
$X$ with entropy $H(X)$ is defined to be $2^{H(X)}.$  If $X$ is
uniformly distributed, then the perplexity of $X$ is
$2^{\log_2\frac{1}{|X|}},$ which reduces to $|X|$.

\subsection{Joint Entropy}

Joint entropy is the entropy of a joint distribution.  Given two
random variables $X$ and $Y,$ a joint probability mass function
$p_{X,Y}(x,y),$ the joint entropy of $X$ and $Y,$ $H(X,Y)$ is defined
as
\beqn H(X,Y) = -\sum_{x\in X} \sum_{y\in Y} p_{X,Y}(x,y) \log_2 p_{X,Y}(x,y).
\eeqn

\subsection{Conditional Entropy}

Conditional entropy is the entropy of a conditional distrition.  Given
two random variables $X$ and $Y,$ a conditional probability mass
function $p_{Y|X}(y|x),$ and a marginal probability mass function
$p_{Y}(y),$ the conditional entropy of $Y$ given $X,$ $H(Y|X)$ is
defined as
\beqn H(Y|X) = -\sum_{x\in X} \sum_{y\in Y} p_{X,Y}(x,y) \log_2 p_{Y|X}(y|x).
\eeqn

{}From probability theory, we know that
\beqn p_{Y}(y) = \sum_{x\in X} p_{X,Y}(x,y). \eeqn
Thus, if $X$ and $Y$ are independent, i.e.  $p_{Y|X}(y|x)=p_{Y}(y),$
then the conditional entropy of $Y$ given $X$ is just the entropy of
$Y:$
\begin{eqnarray}
H(Y|X) & = & -\sum_{x\in X} \sum_{y\in Y} p_{X,Y}(x,y) \log_2 p_{Y|X}(y|x) \\
 & = & - \sum_{x\in X} \sum_{y\in Y} p_{X,Y}(x,y) \log_2 p_{Y}(y) \\
 & = & - \sum_{y\in Y} [\sum_{x\in X} p_{X,Y}(x,y)] \log_2 p_{Y}(y) \\
 & = & - \sum_{y\in Y} p_{Y}(y) \log_2 p_{Y}(y) \\
 & = & H(Y).
\end{eqnarray}

\subsection{Relative Entropy, or Kullback-Liebler Distance}

Relative entropy, or the Kullback-Liebler distance, is a measure of
the distance between two probability distributions.  Given a random
variable $X$ and two probability mass functions $p(x)$ and
$q(x),$ the relative entropy of $p$ and $q,$
$D(p||q),$ is defined as
\beqn D(p||q) = \sum_{x\in{\cal X}} p(x)\log_2\frac{p(x)}{q(x)}.\eeqn

Note that the Kullback-Liebler ``distance'' is not really a distance
measure since, for one thing, it is not symmetric with respect to its
arguments.  The relative entropy function is generally used to measure
how closely a model $q$ correctly matches an empirical distribution
$p.$ If $p(x)=q(x)$ for all $x,$ then $D(p||q)=0.$ Statistical
training algorithms are generally structured as a search for a model
$q$ which minimizes the relative entropy function with respect to an
empirical distribution $p$ extracted from a training corpus.

\subsection{Mutual Information}

The mutual information of two random variables $I(X;Y)$ is defined as
the Kullback-Liebler distance between their joint distribution
and the product of their marginal distributions:
\beqn I(X;Y) = D(p_{X,Y}||p_{X}p_{Y}) = \sum_{x\in X} \sum_{y\in Y}
 p_{X,Y}(x,y) \log_2 \frac{p_{X,Y}(x,y)}{p_{X}(x)p_{Y}(y)} \eeqn
If $X$ and $Y$ are independent, i.e. $p_{X,Y}(x,y)=p_{X}(x)p_{Y}(y),$
then
\begin{eqnarray}
I(X;Y) & = & \sum_{x\in X} \sum_{y\in Y}
 p_{X,Y}(x,y) \log_2 \frac{p_{X,Y}(x,y)}{p_{X}(x)p_{Y}(y)}\\
 & = &  \sum_{x\in X} \sum_{y\in Y}
 p_{X,Y}(x,y) \log_2 \frac{p_{X}(x)p_{Y}(y)}{p_{X}(x)p_{Y}(y)}\\
 & = &  \sum_{x\in X} \sum_{y\in Y} p_{X,Y}(x,y) \log_2 1\\
 & = & 0\\
\end{eqnarray}
Thus, mutual information quantifies the dependence of two random
variables, with a value of 0 indicating independence.

\subsection{Cross Entropy}

Cross entropy is an estimate of the entropy of a distribution
according to a second distribution.  Given a random variable $X$ and
two probability mass functions $p(x)$ and $q(x),$ the cross
entropy of $p$ with respect to $q,$ $H(p,q),$ is
defined as
\beqn H(p,q) =  \sum_{x\in{\cal X}} p(x)\log_2 q(x). \eeqn

Note that the relative entropy of two distributions $p$ and $q$ is
equal to the cross entropy of $p$ and $q$ minus the entropy of X with
respect to $p.$

\begin{eqnarray}
D(p||q)
 & = & \sum_{x\in{\cal X}} p(x)\log_2\frac{p(x)}{q(x)}\\
 & = & \sum_{x\in{\cal X}} p(x)(\log_2 p(x) - \log_2 q(x))\\
 & = & \sum_{x\in{\cal X}} p(x)\log_2 p(x) - p(x)\log_2 q(x)\\
 & = & \sum_{x\in{\cal X}} p(x)\log_2 p(x) -
       \sum_{x\in{\cal X}} p(x)\log_2 q(x)\\
 & = & [- \sum_{x\in{\cal X}} p(x)\log_2 q(x)] -
       [- \sum_{x\in{\cal X}} p(x)\log_2 p(x)]\\
 & = & H(p, q) - H_p(X).\\
\end{eqnarray}

Thus, to minimize the relative entropy of a distribution $q$ with
respect to another distribution $p,$ it is sufficient to minimize
$q$'s cross entropy with respect to $p.$

\section{What is a Statistical Decision Tree?\label{SDTdef}}

A decision tree asks questions about an event, where the particular
question asked depends on the answers to previous questions, and where
each question helps to reduce the uncertainty of what the correct
choice or action is.  More precisely, a {\em decision tree} is an
{$n$}-ary branching tree in which questions are associated with each
internal node, and a choice, or class, is associated with each leaf
node.  A {\em statistical} decision tree is distinguished from a
decision tree in that it defines a conditional probability
distribution on the set of possible choices.

\subsection{Histories, Questions, and Futures}

There are three basic objects which describe a decision tree:
histories, questions, and futures.

A {\em history} encodes all of the information deemed necessary to
make the decision which the tree is asked to make.  The content of a
history depends on the application to which decision trees are being
applied.  In this work, a history consists of a partial parse tree,
and it is represented as an array of {$n$}-ary branching trees with
feature bundles at each node.  Thus, the history includes any aspect
of the trees in this array, including the syntactic and lexical
information in the trees, the structure of the trees, the number of
nodes in the trees, the co-occurrence of two tree nodes in some
relationship to one another, etc.

While the set of histories represents the state space of the problem,
the {\em questions,} and their possible answers, encode the heuristic
knowledge about what is important about a history.  Each internal node
in a decision tree is associated with a question.  The answers to
these questions must be finite-valued, since each answer generates a
new node in the decision tree.

A {\em future} refers to one of a set of possible choices which the
decision tree can make.  The set of choices is the {\em future
vocabulary.} For a decision tree part-of-speech tagger in which the
tree selects a part-of-spech tag for each word in a sentence, the
future vocabulary is the tag set.  Each leaf node of a decision tree
is associated with an element from the future vocabulary, indicating
which choice the decision tree recommends for events which reach that
leaf.

In this parsing work, decision trees are applied to a number of
different decision-making problems: assigning part-of-speech tags to
words, assigning constituent labels, determining the constituent
boundaries in a sentence, deciding the scope of conjunctions, and even
selecting which decision to make next.  Each of these decision-making
tasks has its own definition of a history, i.e. its own set of feature
questions, and its own future vocabulary.  The algorithms which are
described in the rest of this chapter illustrate how decision trees
use distributional information from a corpus of events to decrease the
uncertainty about the appropriate decision to make for each of these
problems.

\subsection{An Example}

I will illustrate decision trees using a simple example involving
determining the shape of an object.  Consider a world where there are
only three possible shapes: square, circle, and triangle.  Objects in
this world have only three measureable attributes: color (red, blue,
magenta, or yellow), height (in inches), and weight (in pounds).  You
are given 100 objects, each labeled with the values for its three
attributes and its correct shape.  From this data, you can create a
decision tree which can predict the shape of future objects based on
their attributes.

In this problem, the history consists of the three attribute values of
the object.  Examples of decision tree questions include: ``what is
the color of the object,'' ``is the object more than 1 pound,'' and
``is the object yellow or is it less than 5 inches in height?''  The
future is the shape of the object, and the future vocabulary is the
set $\{\mbox{square,~circle,~triangle}\}.$ The 100 example objects are
the training corpus which is used to select decision tree questions
and to determine empirical probability distributions.

Assume that the training corpus consists of 80 squares, 10 circles,
and 10 triangles.  Without knowing any information about the
attributes of the objects, we can already assign a distribution on the
shape of objects based on the empirical distribution of the training
corpus: $p(\mbox{square})=0.8,$ $p(\mbox{circle})=0.1,$ and
$p(\mbox{triangle})=0.1.$

\begin{figure}[tbhp]
\centerline{\psfig{figure=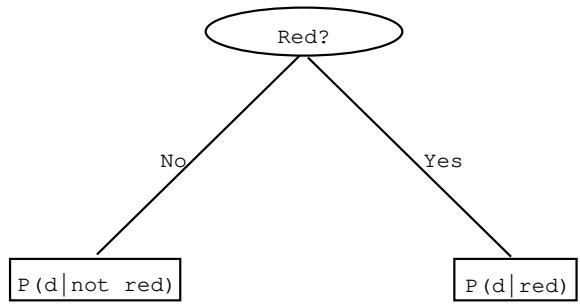,width=3in}}
\caption[Decision tree representing the red/not-red
distinction.]{Decision tree representing the red/not-red distinction.
Here $d$ is the random variable representing the shape of the
object.\label{Q2dtree}}
\end{figure}

Now, let's say that we know the color of these objects.  Specifically,
there are 70 red squares, 10 yellow circles, 10 blue triangles, and
10 blue squares.  If we asked the question ``is the object red?''
we would divide the data into two classes, or decision tree {\em
nodes,} indicating those objects which are red and those which are not
red.  Based on the training data, we know that 70\% of the objects are
red, and all of the red objects are squares, i.e.
$p(\mbox{square}|\mbox{red})=1.$ However, if the object is not red,
which happens 30\% of the time, then it might be a square, circle, or
triangle with equal probability $(p=\frac{1}{3}).$  The decision tree
consisting of this single question is shown in Figure~\ref{Q2dtree}.

Consider how much information we have gained, in terms of entropy
reduction, by asking this single question.  Before asking the
question, the entropy of the shape decision $D,$ was
\beqn H(D) = -\sum_{d\in D} p(d) \log_2 p(d) = - [0.8 \log_2 0.8 + 0.1
\log_2 0.1 + 0.1 \log_2 0.1 ] = 0.92, \eeqn
yielding a perplexity of 1.89.  The conditional entropy of the shape
decision given the answer to the redness question $R$ is
\beqn H(D|R) = -\sum_{r\in\{\mbox{red, not red}\}} \sum_{d\in D}
p_{R,D}(r,d) \log_2 p_{D|R}(d|r) = 0.2, \eeqn with a perplexity of
1.15.  Thus, this decision tree reduces the uncertainty about the
color of the object by 0.72 bits.  And instead of there being nearly 2
possible choices for each event on average, there is now closer to
only 1 choice.

\subsection{Binary Decision Trees\label{BDTexample}}

Decision trees are defined above as {$n$}-ary branching trees, but the
work described here discusses only binary decision trees, where only
yes-no questions are considered.

The main reason for using binary decision trees is that allowing
questions to have different numbers of answers complicates the
decision tree growing algorithm.  The trees are grown by selecting the
question which contributes the most {\em information} to the decision,
where information is defined in terms of entropy.  It is difficult to
compare the information provided by questions with different numbers
of answers, since entropy considerations generally favor questions
with more answers.

As an example of this, consider the case where histories come in
four colors, red, blue, yellow, and magenta.  The question set
includes the following questions:
\begin{enumerate}
\item What is the color of the history?
\item Is the color either blue or red?
\item Is the color red?
\item Is the color magenta?
\end{enumerate}

\begin{figure}[tbhp]
\centerline{\psfig{figure=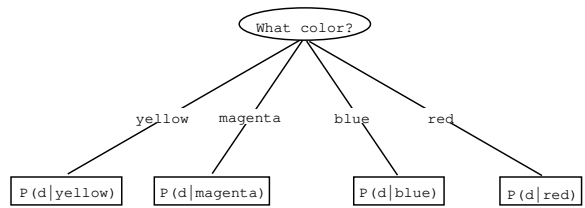,width=3in}}
\caption{A decision tree using {$n$}-ary questions.\label{Q1dtree}}
\end{figure}

\begin{figure}[tbhp]
\centerline{\psfig{figure=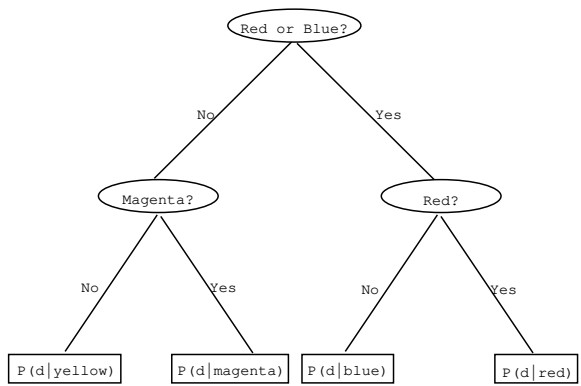,width=3in}}
\caption{A decision tree using only binary questions.\label{Q234dtree}}
\end{figure}

Question 1, with four values, provides the most information, and a
decision tree growing algorithm would certainly select it over the
other questions (Figure~\ref{Q1dtree}).  The decision tree could
effectively ask this question by asking a combination of binary
questions 2, 3, and 4 (Figure~\ref{Q234dtree}); but, it would never
choose this option over the single question.

Now, let's consider the situation where the only important feature of
the history is whether the history is red or not.  While question~3
achieves the same entropy reduction as question~1, question~1 divides
the histories into four different classes when only two are necessary.
This situation is referred to as {\em data fragmentation.} Since
magenta histories and blue histories behave similarly, if there are
very few (or no) magenta histories, then a decision tree which asks
question~1 (Figure~\ref{Q1dtree}) will have more difficulty
classifying the magenta history than one which asks question~3
(Figure~\ref{Q2dtree}).

Another reason for considering only binary questions for decision
trees is computational efficiency.  During the growing algorithm, the
histories at a node are sorted based on the answers to the question
selected for that node.  The case where there are only two possible
answers is simpler to implement efficiently than the general case.
Binary questions also speeds up the mathematical calculations, since
loops which range over all possible answers to questions can be
unraveled.

\subsection{Recasting N-ary Questions as Binary Questions}

It is very difficult to pose all questions about a decision in binary
terms.  In the previous example, it would be counterproductive to
expect a person to notice that blue, yellow, and magenta histories
behave one way and red histories behave another in the training data.

\begin{figure}[tbhp]
\centerline{\psfig{figure=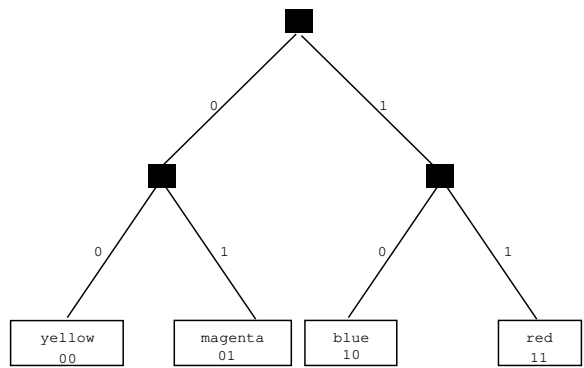,width=3in}}
\caption[A BCT for the {\em color} vocabulary.]{A binary
classification tree for the {\em color} vocabulary, along with its
corresponding binary encoding.\label{colorBCT}}
\end{figure}

An {$n$}-ary question can be recast as a sequence of binary questions
by creating a binary classification tree (BCT) for the {\em answer
vocabulary,} i.e. the set of answers to a given question.  A BCT for
the color questions in section~\ref{BDTexample} is shown in
Figure~\ref{colorBCT}.  BCTs can be acquired using an unsupervised
statistical clustering algorithm, as described in Brown
et.al.\cite{ngrams}; for smaller answer vocabularies, hand-generated
BCTs are a viable alternative.

The binary encoding of {$n$}-ary questions generates an implicit
binary representation for the answer vocabularies, as is labeled in
Figure~\ref{colorBCT}, where each bit corresponds to a decision tree
question.  This interpretation offers two possible difficulties.
First, since these questions are based on a hierarchical
classification tree, the $n$th bit does not necessarily have much
meaning without knowing the values of the first $n-1$ bits.  Also, if
the BCT is unbalanced, the values in the shallower parts of the BCT
will have fewer bits in their representations than those in the deeper
parts.  One could pad these shorter bit strings, but should they be
padded with 0s or 1s?

Both of these problems are solved using principles of information
theory.  Since the children at a given node in the BCT are unordered,
one can use a greedy algorithm to swap the order of the children to
maximize the amount of information contained in each bit.  This
procedure is called {\em bit-flipping.} This makes the padding issue
irrelevant, since regardless of which bit is initially assigned, it
will be flipped if more information is gained by doing so.

Even without bit-flipping, whether or not questions should be asked
out of order is not important.  If a question is meaningless without
knowing the answers to other questions first, the decision tree
growing algorithm will detect this situation and ask only the
meaningful questions.  The exception to this is when there is very
little data available to evaluate the relative value of questions,
which happens in the later stages of the growing algorithm.  {\em
Overtraining} can occur at this point, where coincidences in the
training data lead the algorithm to select questions which will be
uninformative on new data.  This is a general problem in decision
trees, as well as in most inductive algorithms.  It is addressed by
applying a smoothing algorithm using a second set of training data.

\section{Growing a Decision Tree}

In this section, I present the maximum-likelihood (M-L) decision
tree growing algorithm from Bahl et.al.\cite{treelm} and motivate the
modifications made to the algorithm for this work.

\subsection{Notation}

Let the random variables $X$ and $Y$ denote the history and future of
the events being modeled by the decision tree.  ${\cal X}$ is the set
of possible $X$ values and ${\cal Y}$ the set of possible $Y$ values.
Let $C$ denote a corpus of examples of the form $(x,y),$ where
$x\in{\cal X}$ and $y\in{\cal Y}.$

A decision tree $\Dtree$ assigns each history $x$ to a leaf node,
denoted by $l(x).$ ${\cal N}_\Dtree$ is the set of nodes in a decision
tree $\Dtree.$ $N_\Dtree(x)$ denotes the set of nodes along the path
from the root to $l(x),$ including $l(x).$ The $i$th ancestor of a
node $n$ is denoted by $a_i(n),$ where $i$ is the length of the path
from $n$ to $a_i(n).$ Thus, the parent of a node is denoted by
$a_1(n).$

A node $n$ can be interpreted as a subcorpus of a corpus $C$, where
the subcorpus is defined as the set of events in $C$ which visit the
node $n$ on the path from the root to a leaf:
\beqn n = \{ (x,y)\in C : n \in N_\Dtree(x) \}. \eeqn

A boolean question $q_i$ is denoted by two sets $Q_i^C$ and $\bar{Q}_i^C,$
where
\beqn Q_i^C=\{(x,y)\in C:\mbox{ the answer to question $q_i$ is yes for }x\}
\eeqn
and
\beqn \bar{Q}_i^C=\{(x,y): (x,y) \not\in Q_i^C\}. \eeqn
$q_i(x)$ is true if the answer to $q_i$ is yes for $x,$ and $q_i(x)$
is false if the answer to $q_i$ is no for $x.$ The question
$\bar{q}_i$ corresponds to the negation of question $q_i.$

The probability $\hat{p}_i^n(y|x)$ indicates the empirical conditional
probability\footnote{In general, $\hat{p}$ is used to refer to an
emiprical probability distribution, i.e. a distribution estimated
directly using the relative frequencies from a corpus.  On the other
hand, $\tilde{p}$ refers to a smoothed distribution.} that $Y=y$ given
that $n\in N(x)$ and $q_i(x)$ is true:
\beqn
\hat{p}_i^n(y|x)=\frac{|\{(x^\prime,y^\prime)\in~Q_i^n:y^\prime~=y\}|}{|Q_i|}.
\eeqn
Likewise,
\beqn
\bar{\hat{p}}_i^n(y|x)=\frac{|\{(x^\prime,y^\prime)\in~\bar{Q}_i^n:y^\prime~=y\}|}{|\bar{Q}_i|}.
\eeqn

\subsection{The Growing Algorithm}

\begin{figure}[tbhp]
\caption{Maximum likelihood decision tree growing
algorithm.\label{GrowAlgorithm}}

Begin with a single root node $n$ and with a training corpus $C.$
\begin{enumerate}
\item If the $y$ value is the same for all events in $n,$ i.e.
\beqn \exists y_n\in {\cal Y} : \forall (x,y)\in n, y=y_n,\eeqn
then $n$ is a {\em pure} node.  Designate $n$ a leaf node and quit.
\item For each question $q_i$ $(i = 1,2,\ldots,~m),$ calculate the
average conditional entropy $\bar{H}_n(Y|q_i):$
\begin{eqnarray}
\bar{H}_n(Y|q_i)
& = & \Prob\{(x,y)\in Q_i^n \} H(Y|q_i,n)+
     \Prob\{(x,y)\in\bar{Q}_i^n\}H(Y|\bar{q}_i,n)\\
& = & -\frac{|Q_i^n|}{|n|}
       \sum_{y\in {\cal Y}}
       \hat{p}^n(y|x\in Q_i^n)\log_2 \hat{p}^n(y|x\in Q_i^n)\\
&   & -\frac{|\bar{Q}_i^n|}{|n|}
       \sum_{y\in {\cal Y}}
       \hat{p}^n(y|x\in\bar{Q}_i^n)\log_2 \hat{p}^n(y|x\in\bar{Q}_i^n)\\
\end{eqnarray}
\item Determine the question $q_k$ which leads to the lowest entropy:
\beqn k = \argmin_{i}\bar{H}_n(Y|q_i).\eeqn
\item Calculate the reduction in entropy $R_n(k)$ achieved by asking
question $k$ at node $n:$
\beqn R_n(k) = H_n(Y) - \bar{H}_n(Y|q_i).\eeqn
\item If $R_n(k) \leq R_{min}(n),$ then designate $n$ a leaf node and quit.
\item Split node $n$ based on $q_k.$
\begin{enumerate}
\item Assign question $q_k$ to node $n.$
\item Create left and right children nodes $n_l$ and $n_r.$
\item Assign nodes to $n_l$ and $n_r$ such that $C_{n_l}=Q_i^n$ and
$C_{n_r}=\bar{Q}_i^n.$
\item Recursively apply this algorithm to $n_l$ and $n_r,$ removing
$q_k$ from the list of candidate questions.
\end{enumerate}
\end{enumerate}

\end{figure}

The basic M-L decision tree growing algorithm is
shown in Figure~\ref{GrowAlgorithm}.  The algorithm, starting with a
set of questions $\{q_1,q_2,\ldots,q_m\}$ and a training corpus $C,$
generates a decision tree which minimizes the expected conditional
entropy of the training data.

The main issue in applying a decision tree growing algorithm to a
problem is to decide on an appropriate {\em stopping rule.} The
stopping rule is the criterion by which the algorithm stops splitting
a node.

Stopping rules are motivated by the fact that as the number of events
at a node gets smaller, the decisions made based on the empirical
distribution of these events become less accurate.  This means that
not only are the probability distributions at these nodes called into
question, but also, since the conditional entropy values
$\bar{H}_n(Y|q_i)$ are estimated empirically from the events at a
node, the entire splitting process is suspect.  Significant splits
{\em might} occur using estimates from sparse data, but there is no
way to determine the value of a split without validating the decision
using more data.  More likely, splits which occur based on fewer
events will result in overtraining.

In the algorithm in Figure~\ref{GrowAlgorithm}, the stopping rule
dictates that a node should not be split if the entropy reduction
$R_n(k)$ achieved by asking the best question $q_k,$ is less than some
minimum value $R_{min}(n).$ This minimum value can be a constant, but
it also might be a function of the number of events at the node $n.$
One heuristic to follow is that the fewer events at a node, the higher
the $R_{min}(n)$ should be in order to consider the split
statistically significant.  One function used in experiments is the
product of the number of events at a node and the entropy reduction
achieved by the split, $|n|\cdot~R_{n}(k).$ The units of this function
are bit-events.

An alternative to a stopping rule is to grow the tree to completion
and then prune nodes based on the significance of splits.  For each
node $n$, consider the node's children, $n_l$ and $n_r.$ If either
node is not a leaf, apply the pruning algorithm recursively to the
non-leaf child(ren).  If both nodes are leaves after the pruning has
been applied recursively, then prune the children of $n$ if the split
at $n$ does not satisfy the stopping rule.  Results from experiments
involving this type of pruning algorithm are reported in
Chapter~\ref{RESULTchapter}.

In this work, decision trees are grown using an $R_{min}(n)$ value of
0, i.e. decision trees are grown until none of the questions cause a
reduction in entropy at a node.  To avoid overtraining and to
compensate for inaccurate probability estimates due to sparse data, an
expectation-maximization smoothing algorithm is applied to the
decision tree using a second training corpus.

\section{Training a Decision Tree Model\label{SmoothSection}}

The decision to grow trees to completion was made based on a
previously unpublished result comparing the test entropy of decision
tree models using various combinations of growing algorithms, stopping
rules, and training data set sizes.\footnote{These experiments were
performed during the summer of 1993 by members of the IBM Speech
Recognition group, Peter F. Brown, Bob Mercer, Stephen Della Pietra,
Vincent Della Pietra, Joshua Koppelman, and myself.} The experiments
were performed on the language modeling problem, predicting the class
of the next word given the previous words in the sentence.  The
variations included: asking all of the questions in a predefined order
vs.\ selecting the question order based on entropy reduction; growing
the tree to completion vs.\ applying a chi-squared test with
thresholds of 5, 10, and 15; and using different size training and
smoothing sets.  Regardless of the amount of training and smoothing
data used, the best results were achieved by growing the tree to
completion using entropy reduction as a question selection criterion.
Different problems may exhibit different behaviors with regards to
stopping rules, but in experiments involving applying a stopping rule
to the parsing decision tree models, the trees grown to completion
performed better on test data than those that were pruned.

The main reason the decision trees can be grown to completion without
overtraining is that, after the model is grown, it is smoothed using a
second, held-out training set.  The smoothing procedure does not
modify the structure of the decision tree or the questions asked at
any of the nodes.  Instead, it effectively adjusts the probability
distributions at the leaves of the decision tree to maximize the
probability of the held-out corpus.

If the leaf distributions were completely unconstrained during the
smoothing procedure, then the best model it could find would simply be
the M-L model determined by mapping each event in the
held-out data to a leaf node and computing the relative frequency of
the futures at each node.  But this would result in overtraining on
the smoothing data.  To avoid this, the smoothing procedure uses the
intuition behind stopping rules to uncover a model which, in a sense,
statistically unsplits nodes which should not have been split.

Stopping rules dictate that, if there is not sufficient confidence
that any question provides information about the future, then no
question should be asked.  Smoothing the model after growing
eliminates the need for making such harsh and irreversible decisions.
Each node $n$ in the decision tree is assigned a parameter $\lambda_n$
which loosely represents the confidence in the distribution at node
$n.$ The smoothed probability at a leaf, $\tilde{p}_n(y|x)$ is defined
recursively as
\beqn \tilde{p}_n(y|x) = \lambda_n p_n(y|x) + (1-\lambda_n)
\tilde{p}_{a_1(n)}(y|x). \eeqn
The smoothed probability of the root node is defined as:
\beqn
\tilde{p}_{\mbox{root}}(y|x) = \lambda_n p_{\mbox{root}}(y|x) +
(1-\lambda_n) \frac{1}{|{\cal Y}|}.
\eeqn
If it turns out that a node $n$ should not have been split, then the
smoothing algorithm can assign $\lambda_{n_l}=\lambda_{n_r}=0,$
effectively pruning the children of $n.$

\subsection{The Forward-Backward Algorithm for Decision Trees}
\begin{figure}[tbhp]
\centerline{\psfig{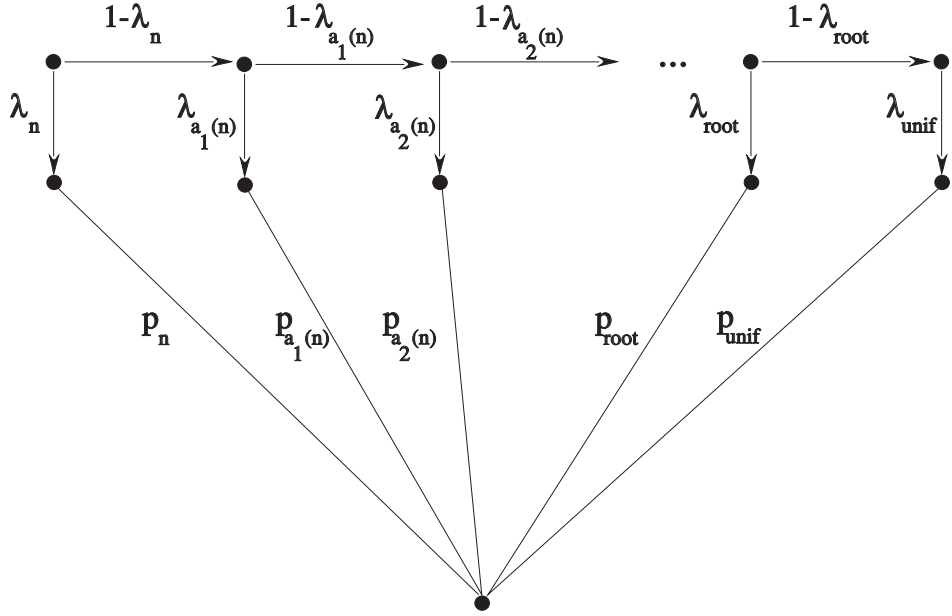}}
\caption[Finite-state machine for Forward-Backward
algorithm.]{Finite-state machine indicating the path for computing the
forward probability in the Forward-Backward algorithm.\label{FBFSM}}
\end{figure}

The Forward-Backward (F-B) algorithm \cite{Baum} can be used to search
for the optimal parameter settings for
$\Lambda=\{\lambda_1,\lambda_2,\ldots,\lambda_m\}.$ Given a held-out
corpus $C_h$, the F-B algorithm computes the probability of $C_h$
using $\Lambda$ in the forward pass, and then computes the updated
value vector $\Lambda^\prime$ in the backward pass.\footnote{This
smoothing algorithm was first published in Lucassen's dissertation in
1983 \cite{Lucassen}.  It was also mentioned briefly in Bahl
et.al.\cite{treelm}.}

The algorithm starts by assuming some initial value for $\Lambda.$
Consider the diagram in Figure~\ref{FBFSM} of a finite-state machine
representing the path from a leaf node to the root.  Imagine that a
history $x$ outputs its future $y$ according to some distribution
along the path from $l(x)$ to the root, or possibly it outputs its $y$
according to the uniform distribution.  Let $\alpha_n(x)$ represent
the probability of starting at state $l(x)$ and visiting state $n.$
Then
\beqn
\alpha_n(x)=\left\{
\begin{array}{ll}
{\displaystyle \prod_{\mbox{$d$ along path from $n$ to $l(x)$}}
 (1-\lambda_d)} & n\in N(x), n\not= l(x)\\
1 & n = l(x)\\
0 & n\not\in N(x)\\
\end{array}
\right.
\eeqn
Let $\Prob_\Lambda(y,n|x)$ be the probability of generating $y$ from state
$n$ on input $x.$  Then
\beqn \Prob_\Lambda(y,n|x) = \alpha_n(x)\lambda_n p_n(y|x).
\eeqn
Now, let $\beta_n(y|x)$ be the probability that $y$ was generated on
input $x$ from one of the states in $N(n).$  Then
\beqn
\beta_n(y|x) = \lambda_n p_n(y|x)+(1-\lambda_n)\beta_{a_1(n)}(y|x).
\eeqn
Let $\Prob_\Lambda^+(y,n|x)$ be the probability of visiting state $n$
but outputting $y$ from a state {\em other than $n$.}  Then
\beqn
\Prob_\Lambda^+(y,n|x) = \alpha_n(x)(1 - \lambda_n)\beta_{a_1(n)}(y|x).
\eeqn
Notice that $\beta_{l(x)}(y|x)$ is the probability of generating $y$
on the input $x.$

Let $\Prob_\Lambda(n|x,y)$ be the probability of having generated $y$
from $n$ given that $y$ was output.  Let $\Prob_\Lambda^+(n|x,y)$ be
the probability that $x$ was generated by some state along the path
from $n$ to the root other than $n.$  These are given by:
\begin{eqnarray}
\Prob_\Lambda(n|x,y) & = \frac{\Prob_\Lambda(y,n|x)}{\Prob_\Lambda(y|x)}
& = \frac{\alpha_{n}(x)\lambda_{n}p_n(y|x)}{\beta_{l(x)}(y|x)}\\
\Prob_\Lambda^+(n|x,y) & = \frac{\Prob_\Lambda^+(y,n|x)}{\Prob_\Lambda(y|x)}
&=\frac{\alpha_{n}(x)(1-\lambda_{n})\beta_{a_1(n)}(y|x)}{\beta_{l(x)}(y|x)}.\\
\end{eqnarray}

The F-B updates for the parameters are
\beqn
\lambda_n^\prime = \frac{\displaystyle \sum_{(x,y)\in C_h}\Prob_\Lambda(n|x,y)}
{\displaystyle \sum_{(x,y)\in C_h}\Prob_\Lambda(n|x,y) +
 \sum_{(x,y)\in C_h}\Prob_\Lambda^+(n|x,y)}.
\eeqn

\subsection{Bucketing $\lambda$'s}

Generally, less data is used for smoothing a decision tree than for
growing it.  This is best, since the majority of the data should be
used for determining the structure and leaf distributions of the tree.
However, as a result, there is usually insufficient held-out data for
estimating one parameter for each node in the decision tree.

A good rule of thumb is that at least 10 events are needed to estimate
a parameter.  However, since each event is contributing to a parameter
at every node it visits on its path from the root to its leaf, this
rule of thumb probably is insufficient.  I have required at least 100
events to train each parameter.

Since there will not be 100 events visiting each node, it is necessary
to {\em bucket} the $\lambda$'s.  The nodes are sorted by a primary
key, usually the event count at the node, and any number of secondary
keys, e.g. the event count at the node's parent, the entropy of the
node, etc.  Node buckets are created so that each bucket contains
nodes whose event counts sum to at least 100.  Starting with the node
with fewest events, nodes are added to the first bucket until it fills
up, i.e. until it contains at least 100 events.  Then, a second bucket
is filled until it contains 100 events, and so on.  Instead of having
a unique parameter for each node, all nodes in the same bucket have
their parameters tied, i.e. they are constrained to have the same
value.\footnote{For historical reasons, this method is referred to as
{\em the wall of bricks.}}

\section{Problems with M-L Decision Trees\label{DTREEproblems}}

The maximum-likelihood approach to growing and smoothing decision
trees described above is effective, but it has some very significant
flaws.

\subsection{Greedy Growing Algorithm}

The algorithm used for growing decision trees is a greedy algorithm.
While there is little doubt that the greediness of the search results
in suboptimal decision tree models, it is not clear how to limit the
extent of the damage done by the short-sighted procedure.

One possible solution is using cross-validation techniques, such as
jackknifing or validating against a second training set, to prevent
suboptimal splits.  Another idea is to increase the depth of the
search, so that at least it is a little less greedy.  Experiments
involving combinations of both of these ideas failed to yield
improvements on test data.

It is possible that for a given set of questions, there are many
decision trees which, combined with the smoothing algorithm, yield
effectively the same models in terms of test entropy, and the greedy
growing algorithm finds one of them.  But there is no way determine
this experimentally, since exhaustively searching the space of models
is computationally prohibitive.

\subsection{Data Fragmentation and Node Merging}

Another problem concerns unnecessary data fragmentation.  This problem
has more to do with the structure of decision trees than with the M-L
strategy itself.  Since decision trees are restricted to be trees, as
opposed to arbitrary directed graphs, there is no mechanism for
merging two or more nodes in the tree growing process.

In some cases, the best model might be found only by implementing some
form of node merging.  For instance, consider the case where it is
informative to know when either $q_1(x)$ is true or $q_2(x)$ is false.
While one can create a question $q_3(x)$ which represents this logical
combination of $q_1$ and $q_2,$ it is not feasible to consider all
possible logical combinations of questions.  In this case, the
decision tree might elect to ask $q_1,$ and when $q_1(x)$ is false, it
will then ask $q_2.$ This causes data fragmentation.  In this case,
the events where $q_1(x)$ is false and $q_2(x)$ is true behave
similarly, and should be combined into one node.  But they are divided
among two nodes, and there is no mechanism for reuniting them.

Node merging is an expensive operation, since any subset of the leaf
nodes at any point in the growing operation is a candidate for
merging.  It might be feasible to consider a small class of node
merging cases.  For instance, when a node has fewer than $k$ events,
the algorithm might try to merge it with a nearby node in the tree.
However, if nodes are allowed to become too small before they are
examined for merging, then the node merging may be ineffective, since
data fragmentation higher up in the tree may have already lead to
suboptimal node splits.

\subsection{Flaws in Smoothing}

There are numerous flaws in the smoothing procedure.\footnote{While
writing up this dissertation, I consulted members of the IBM research
group about some of the finer points of the algorithms I have
described in this chapter.  Without fail, each person I asked about
the smoothing algorithm remarked, aghast, ``Are we still using {\em
that} old smoothing algorithm?''  It is over a decade old, and the
only reason it is still in use is that no one has found anything that
works as well.  Of course, I'm not sure how hard they looked.} For
instance, the parameterization scheme only loosely represents the
information it is intended to reflect.  Consider a node $n$ which is
the parent of one leaf, $n_l,$ and another internal node, $n_r.$
Assume $n_l$ has very few events, and is not very significant, while
$n_r$ has many events.\footnote{While this may seem like an odd
occurence, given the growing procedure, it does happen in the case
where each question individually is not very informative, but a
combination of questions {\em is} informative.} Node $n$ is assigned
only one parameter, yet it plays two very different roles.
$\lambda_{n_l}$ will be set very close to 0; so, for events that reach
$n_l,$ node $n$ plays the role of a leaf, and its parameter is very
important.  However, assuming node $n_r$ is the root of large subtree,
node $n$ has very little impact on the smoothed distributions at the
leaves beneath $n_r.$ Thus, for events that visit $n_r,$ $\lambda_{n}$
is not very meaningful in the M-L framework.  But the updated value of
$\lambda_n$ will be affected by {\em all} held-out events that visit
$n,$ most of which are likely to visit $n_r.$

Another problem with the smoothing is a result of the bucketing
scheme.  Because of the way parameters are tied together, it is
possible to train and smooth a model and inadvertently assign an
plausible event a probability of 0.  This is a horrible condition in
any probabilistic model, since it results in an infinite entropy, and
thus causes any training algorithm which depends on entropy
calculations to fail.  It also illustrates the weakness of the
parameterization scheme.  Clearly, the $\lambda$'s are failing to
fulfill their role if the probability of an event is so grossly
underestimated.  It is likely that other probabilities are
inaccurately estimated as well, and it is only the 0's which stand
out.

\subsection{Maximizing the Wrong Function}

The most obvious and significant problem with the decision tree
algorithms described in this chapter is that the decision tree growing
algorithm is maximizing the wrong objective function!

It is always important to construct a model using the same probability
function which will be used when the model is applied to new data.
Otherwise, conclusions drawn about the training data are less likely
to be true when the model is applied in tests.

Notice that the parameterization scheme introduced in the smoothing
algorithm isn't considered in the growing.  The decision tree is grown
to minimize the expected entropy according to the empirical
distributions at the leaves.  But these distributions are {\em never}
used on their own to determine the probability of other data.

The correct growing algorithm would calculate the entropy reduction
for each question using the smoothed model instead of the unsmoothed
model.  However, this would require performing {\em at least} a few
iterations of the F-B algorithm {\em for every question} and {\em at
every node.} This calculation is prohibitive with current hardware.
And with the current bucketing scheme, the smoothing algorithm is not
completely defined until the entire tree is grown.  There is no
obvious way to correct the decision tree growing objective function
without redesigning the entire process.

\subsection{Final Comments}

Given the negative tone of this section, it is important to remind the
reader that maximum-likelihood decision trees have proven effective at
modeling parsing decisions, and they have been successfully applied to
problems in speech recognition and machine translation, as well.

The weaknesses revealed in the discussion above suggest that there are
some basic research issues in the area of decision tree modeling
which should be explored.  However, even with these flaws, decision
tree techniques provide more accurate and detailed models than the
statistical methods generally applied to NLP problems.  The ability to
ask arbitrary questions about a history and to smooth by backing off
one question at a time makes decision trees a preferable technology to
the simpler {$n$}-gram models with non-iterative smoothing.

\def\argmax{\mathop{\rm arg\,max}}
\newcommand{\subp}[1]{$\mbox{#1}_p$}
\newcommand{\ip}[1]{$\mbox{#1}_p^i$}
\newcommand{\fp}[1]{$\mbox{#1}_p^f$}

\chapter{Preliminary Experiments\label{PRELIMchapter}}

The development of the SPATTER parsing algorithm was motivated by
earlier successes in incorporating contextual information into a
statistical parsing model.  Until 1990, the state-of-the-art
statistical parsers used probabilistic context-free grammars, which
assign distributions on parse trees by assuming that context-free rule
applications were completely independent of the context in which they
were applied.  While this independence assumption simplifies the
mathematics of estimating P-CFG models, it is nonetheless
inappropriate for natural language context-free grammars.  Rule
applications are {\em not} independent of their context, and it seemed
likely that statistical models which considered contextual information
would allow more accurate analysis of language.

To test this hypothesis, I performed two preliminary experiments.  In
the first experiment, context-free rule productions extracted from the
Unisys string grammar were used in the development of a
context-sensitive statistical parser, called {$\cal P$}earl.  This
experiment illustrates the benefits of using limited lexical
information and nearby structural information in a parsing model. In
the second experiment, performed in collaboration with IBM Language
Modeling group members Ezra Black, Fred Jelinek, John Lafferty, Robert
Mercer, and Salim Roukos, a grammar-based parser developed by the IBM
group is adapted to use a generative statistical parsing model in
place of the simpler P-CFG model.  The new model uses statistical
decision tree algorithms to estimate models which incorporate more
contextual information than was feasible in the {$\cal P$}earl work,
including multiple words of lexical context.

In this chapter, I describe the relevant details of these two
experiments and discuss the conclusions drawn from them which
suggested the development of the SPATTER parser.  In describing the
second experiment, I also introduce the task domain which is used in
the experiments described in Chapter~\ref{RESULTchapter}.

\section{A Context-sensitive Probabilistic Parser}

The {$\cal P$}earl experiment was intended to test the hypothesis that
a statistical model which considered contextual information in
assigning a probability distribution over the space of parse trees
could be trained automatically to perform the task of disambiguation
in the parsing problem.  Beginning with a grammar which depended
heavily on rule-based disambiguating criteria, the experiment involved
removing the disambiguating rules and replacing them with an algorithm
which selected the sequence of rule applications which yielded the
highest probability parse according to a statistical model.

The product of this initial experiment was the {$\cal P$}earl parser,
developed at Unisys in collaboration with Dr. Mitch Marcus of the
University of Pennsylvania.  Preliminary experimental results using
{$\cal P$}earl were published in Magerman and Marcus\cite{Pearl}.  To
test the {$\cal P$}earl model further, and to correct an error in the
implementation of the {$\cal P$}earl model, the {$\cal P$}icky parser
was developed two years later at Stanford in collaboration with Dr.
Carl Weir of Unisys.  This work was published in Magerman and
Weir\cite{Picky}.  In this section, I describe the {$\cal P$}earl
model including the correction made in the {$\cal P$}icky work.

\subsection{The PUNDIT String Grammar}

The grammar to which this experiment was applied is the PUNDIT string
grammar developed at Unisys \cite{PUNDIT}.  The PUNDIT string grammar
consists of a set of context-free rules, referred to as the {\em
context-free backbone} of the grammar, and a set of restrictions on
the applications of these rules.  For instance, one restriction might
dictate that the rule VP~$\rightarrow$~V~PP could not be applied if
the PP (prepositional phrase) contained the preposition {\bf of} and
if the V (verb) was {\bf eat}.  This type of restriction is called a
{\em selectional} restriction. Some selectional restrictions are more
general than this example, referring to word classes instead of
specific words.  Some also consider more complex functions of the
context.

Restrictions also control the generation of {\em gaps} in a parse
tree.  Gaps, which originated in transformational grammar, are place
holders for words which have been moved or removed as a result of a
transformation, such as question inversion or subject raising.  For
instance, given the question
\beqn \mbox{Whom does Mary love?} \eeqn
the PUNDIT grammar would generate a gap after the word {\bf love,}
interpreting the gap as a noun phrase.  By inserting the gap, the
analysis of the verb phrase in the sentence
\beqn \mbox{Whom does Mary [VP love [NP NULL NP] VP] ?} \eeqn
is consistent with the analysis of the verb phrase in the
corresponding sentence without the question inversion:
\beqn \mbox{ Mary [VP loves [NP whom NP] VP] ?} \eeqn

\subsection{Replacing the Restrictions with Statistics}

One of the problems with the PUNDIT string grammar is that it is
tedious and time-consuming to develop the restriction set.  The
development of the knowledge base of selectional and gap restrictions
requires a grammarian to identify in example sentences the lexical and
structural co-occurrences that trigger or inhibit each rule.  This
task became a full-time job for the Unisys grammarian, since the
restrictions needed to be constantly improved for old domains and
modified significantly for new domains.

One approach to automating this process is to develop a probabilistic
model for the context-free backbone of the grammar which accomplishes
the task of the restrictions.  The restriction grammar effectively
acts like a probabilistic grammar whose model assigns a probability of
0 to some rules given certain contexts and assigns a uniform
probability over all other possibilities.  The initial problem with
this approach was that the P-CFG, the only existing form of
probabilistic grammar at the time, could not capture the
co-occurrences represented by the restrictions, since a P-CFG's
probabilistic model considers only the left-hand side of a rule when
determining a rule's probability.

Solving this problem required developing a more context-sensitive
probabilistic grammar model.  The co-occurrences encoded in the
restrictions {\em could} be represented using a probabilistic model
which conditions the probability of a rule production on lexical and
structural information from the context.  Instead of asking a
grammarian to recognize the important co-occurrences for each rule, a
statistical model could encode the relative frequencies of the
contexts in which a rule occurred in a set of sentences parsed
correctly using the grammar.  If the model conditioned its
probabilities on the same information that the restrictions used, then
it is conceivable that the model could represent the same information
contained in the restriction set, and the model could be trained
automatically using statistical training algorithms.

\subsection{A Context-Sensitive Parsing Model}

The {$\cal P$}earl model, a context-sensitive parsing model, considers lexical
and structural context when assigning a probability to a rule
application, in contrast to a P-CFG model which ignores the context in
which a rule is applied when assigning rule probabilities.  The {$\cal P$}earl
model estimates the probability of each parse $T$ given the words in
the sentence $S$, ${\cal P}(T|S)$, by assuming that each non-terminal
and its immediate children are dependent on the non-terminal's
siblings and parent and on the part-of-speech trigram centered at the
beginning of that rule:
\begin{equation}
{\cal P}(T|S) \simeq \prod_{A \in T} {\cal P}(A \rightarrow \alpha | C
\rightarrow \beta A \gamma \mbox{, } a_0 a_1 a_2) \\
\end{equation}
where $C$ is the non-terminal node which immediately dominates $A,$
$a_1$ is the part-of-speech associated with the leftmost word of
constituent $A,$ and $a_0$ and $a_2$ are the parts-of-speech of the
words to the left and to the right of $a_1,$ respectively.  See
Magerman and Marcus\cite{Pearl} for a more detailed description of
this model.

\subsection{Experimental Results}

Experiments were performed which tested whether or not the {$\cal P$}earl
parsing model could represent the same information encoded in the
restriction set.  In these experiments, the context-free backbone of
the PUNDIT grammar was used to parse sentences.  This unrestricted
grammar generates thousands of analyses for each sentence.  To avoid
generating all of the parses, the parser searches the space of
possible parses guided by the probability function, using an
agenda-based parsing algorithm that expands only the most probable
partial analyses.  These experiments evaluated for what percentage of
the test sentences the parser would be able to select a correct parse
from the space of grammatical parses.

In the original experiment, the {$\cal P$}earl parser was trained on 1,100
sentences from the Voyager direction-finding domain \cite{VOYAGER} and
tested on 40 test sentences from the same domain.  Of these 40
sentences, {$\cal P$}earl produced parse trees for 38 of them, and 35
of these parse trees were equivalent to the correct parse produced by
Pundit, for an overall accuracy rate of 88\%.

After the development of the {$\cal P$}icky parser, a more rigorous evaluation
of the {$\cal P$}earl parsing model was performed.  The {$\cal P$}icky parser
was trained on almost 1,000 sentences and was tested on 3 sets of 100
sentences which were held out from the rest of the corpus during
training.  The training corpus consisted of 982 correctly-parsed
sentences.  The average accuracy of the {$\cal P$}icky parser on these 3 data
sets was 89.3\%.

\subsection{Conclusions from the {$\cal P$}earl Experiment}

The {$\cal P$}earl parsing model illustrated the ability of a context-sensitive
probability model to represent the knowledge contained in a
restriction set generated over the course of many years by a
grammarian.  The results were surprising, since the grammarians at
Unisys considered it unlikely that the statistical models would be
able to predict the locations of the gaps, since it was presumed that
the generation of gaps was a semantic phenomenon and not a syntactic
one.  Nonetheless, the {$\cal P$}earl model proved itself capable of
identifying the gaps in the limited tests of its abilities.

There were two main limitations of the {$\cal P$}earl experiments.  The most
critical weakness of the work was the unreliability of the
experimental results.  The {$\cal P$}earl model was applied to a very simple
problem, and was tested on very little data.  Further, the test data
was viewed repeatedly during the experiments, and the same test data
was tested on over and over again.  It was possible that the
development process tuned the parser to the specific training set.  In
short, these experiments violated many of the rules of evaluations
discussed in Chapter~\ref{EVALUATIONchapter}.

The {$\cal P$}earl experiment also suffered from the oversimplicity of its
statistical model.  While the model considered far more contextual
information than a P-CFG, it ignored lexical information and
considered very little structural context.

{$\cal P$}earl was a reasonable first step in the direction of
context-sensitivity in parsing models.  And given the amount of
training data used, it probably would not have been possible to
estimate a more complex model using the resources available.  But, if
a parser using a context-sensitive probabilistic model were applied to
a more challenging domain, or to a corpus of general English, its
model would need to consider significant amounts of lexical and
structural information.

\section{A Decision Tree Parsing Model\label{HBGsection}}

The comparison of the {$\cal P$}earl model with the P-CFG model suggested that
increasing the context-sensitivity of a parsing model led to a higher
degree of parsing accuracy.  However, two questions remained.  First,
how much contextual information could effectively be incorporated into
a statistical model, given state-of-the-art statistical estimation
techniques?  And second, how would this form of statistical
disambiguation perform on a domain which was more difficult to parse.

Soon after the {$\cal P$}earl experiment, members of the IBM Language Modeling
group were developing the theoretical underpinnings of a class of
parsing models called {\em history-based grammar} (HBG) models.  HBG
models are based on the principle that the probability of any action
in the parsing process is potentially affected by any or all of the
actions which preceded it.

In order to answer the questions raised by the {$\cal P$}earl experiment, and
to explore the effectiveness of HBG models applied to parsing, an
experiment was devised to compare a {$\cal P$}earl-like HBG model with an
existing P-CFG model.

In this experiment, both the P-CFG and HBG models are applied to the
parses output by a broad-coverage, feature-based unification grammar.
The question which this experiment explores is: how much improvement
in parsing accuracy can be achieved by replacing a P-CFG model with a
lexically-sensitive HBG model?

This experiment differs from the {$\cal P$}earl experiment in three
significant ways.  First, the task domain, parsing computer manuals,
is more challenging than the Voyager domain, including longer
sentences, more complex structures, and more lexical and syntactic
ambiguity.  Second, the grammar to which the context-sensitive model
has already been probabilized.  Thus, there is a P-CFG to test the new
model against directly.  Finally, using the statistical decision tree
algorithms developed by the IBM group, it would be possible to
accurately estimate a probabilistic model using much more contextual
information than was previously attempted.

In this section, I define the history-based grammar model, briefly
discuss the grammar to which this experiment is applied, describe the
parsing model used in this experiment, report the test results, and
discuss the conclusions drawn from this experiment.

However, before beginning the discussion of history-based grammars, I
introduce the Lancaster Computer Manuals Treebank, which is the
treebank used in both the history-based grammar experiments as well as
in the experiments in Chapter~\ref{RESULTchapter}.

\def\lbrack{[}
\def\rbrack{]}
\def\amp{\&}
\subsection{The Lancaster Computer Manuals Treebank\label{TREEBANKsection}}

Training a context-sensitive decision tree parsing models requires a
very large treebank.  It is important that the treebank be annotated
accurately and consistently.  In evaluating a new parsing model, it is
also useful to use a data set to which an existing state-of-the-art
probabilistic parser has already been applied, so that previous
results can be used to gauge the progress made by using the new
methods introduced by the new work.

Using these guidelines, the domain I selected to use for the
history-based parsing experiments, as well as my dissertation
experiments, is the Computer Manuals domain, and the training and test
data I used is the Lancaster Computer Manuals Treebank.  Black,
Garside, and Leech \cite{blackbook} provides detailed reports on
experiments performed using his P-CFG, some of which is described in
this section.  And since I have access to both the training and test
data used in these previous experiments, I have the opportunity to
perform a direct comparison between the two parsing models, training
and testing on exactly the same sentences.

The treebank sentences were selected from 40 million words of IBM
computer manuals.  The sentences from the computer manuals which were
used in the treebank were selected in the following way.  First, the
3,000 most frequent tokens\footnote{The notion of a token is not
clearly defined in \cite{blackbook}.  However, since there are over
7,000 unique words in the treebank training set, a token is {\em not}
the same as a word.  I believe that each word is made up of one or
more tokens, and the extra 4,000 words in the training data are
multi-token words.} were identified in a corpus of 600,000 words from
10 manuals.  A sentence from a computer manual is used in the treebank
only if all of the tokens in the sentence are among these 3,000 most
frequent.  The treebank consists of a few million words of sentences
selected in this manner from the 40 million words of manuals.

The following sentence is a randomly selected example from the
computer manuals domain:
\begin{quotation}
It indicates whether a call completed successfully or
if some error was detected that caused the call to fail.
\end{quotation}
Figure~\ref{bracketing} shows the Lancaster treebank bracketing of
this sentence.

\begin{figure}[hbt]
\begin{tabbing}
\lbrack N It\_PPH1 N\rbrack  \\
\lbrack V \=indicates\_VVZ  \\
   \> \lbrack Fn \= \lbrack Fn\amp \= whether\_CSW  \\
   \>            \>             \> \lbrack N a\_AT1 call\_NN1 N\rbrack   \\
   \>            \>             \> \lbrack V completed\_VVD successfully\_RR
V\rbrack Fn\amp\rbrack  \\
   \>            \> or\_CC  \\
   \>            \> \lbrack Fn+ \= if\_CSW \\
   \>            \>             \> \lbrack N some\_DD error\_NN1 N\rbrack @ \\
   \>            \>             \> \lbrack V was\_VBDZ detected\_VVN V\rbrack
\\
   \>            \>             \> @\lbrack Fr \= that\_CST \\
   \>            \>              \>      \> \lbrack V \= caused\_VVD \\
   \>            \>              \>       \>          \> \lbrack N the\_AT
call\_NN1 N\rbrack  \\
   \>            \>              \>       \>          \> \lbrack Ti
to\_TO fail\_VVI Ti\rbrack V\rbrack Fr\rbrack Fn+\rbrack \\
   \>            \>              \>       \>          \> \ \ \ \ \ \ \ \ \ \ \
\ \ \ \ \ \ \ \ \ \ \ \ \ Fn\rbrack V\rbrack  .\_.
\end{tabbing}
\caption{Sample bracketed sentence from Lancaster Treebank.\label{bracketing}}
\end{figure}

The Lancaster treebank uses 195 part-of-speech tags and 19
non-terminal labels.  A complete list of the tags and labels is
included in Appendix~\ref{VOCABULARYappendix}.  The definitions of
these tags and labels are given in \cite{blackbook}.

\subsection{The History-based Grammar Model}

A generative probabilistic grammar model estimates the joint
probability of a derivation tree $T$ and the observed sentence, $S,$
denoted by $p(T,S).$ The P-CFG model estimates this probability by
assuming all derivational steps in $T$ are independent.  The
history-based grammar model makes the opposite assumption, namely that
the probability of each derivational step depends on all previous
steps.

\begin{centering}
\begin{figure}[tbhp]
\begin{eqnarray*}
   S & \rightarrow & ASB | AB \\
   A & \rightarrow & a \\
   B & \rightarrow & b\\
\end{eqnarray*}
\centerline{\psfig{figure=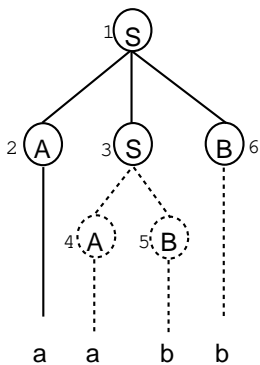}}
\caption{Grammar and parse tree for {\it aabb}.\label{tree}}
\end{figure}
\end{centering}

The history-based grammar model defines the context of a parse tree in
terms of the leftmost derivation of the tree.  Consider the
context-free grammar for ${a^nb^n}$ and the parse tree for the
sentence {\it aabb} shown in Figure~\ref{tree}. The leftmost
derivation of the tree $T$ in Figure~\ref{tree} is:
\begin{equation}
S \stackrel{r_1}{\rightarrow} ASB
  \stackrel{r_2}{\rightarrow} aSB
  \stackrel{r_3}{\rightarrow} aABB
  \stackrel{r_4}{\rightarrow} aaBB
  \stackrel{r_5}{\rightarrow} aabB
  \stackrel{r_6}{\rightarrow} aabb
\end{equation}
where the rule used to expand the $i$-th node of the tree is denoted
by $r_i$. Note that we have indexed the non-terminal nodes of the tree
with this leftmost order.  We denote by $t_i^-$ the sentential form
obtained just before we expand node $i$.  Hence, $t_3^-$ corresponds
to the sentential form ${\it a}SB$ or equivalently to the string
$r_1r_2$. In a leftmost derivation we produce the words in
left-to-right order.

Using the one-to-one correspondence between leftmost derivations and
parse trees, we can rewrite the joint probability $p(T,S)$ as:
\beqn p(T,S) = \prod_{i=1}^m p(r_i | t_i^- ).\eeqn
In a P-CFG, the probability of an expansion at node $i$ depends only
on the identity of the non-terminal $N_i$, i.e.,
$p(r_i|t_i^-)=p(r_i)$. Thus
\beqn p(T,S) = \prod_{i=1}^m p(r_i).\eeqn
So, in a P-CFG, the derivation order does not affect the probabilistic
model\footnote{Note the abuse of notation since we denote by $p(r_i)$
the conditional probability of rewriting the non-terminal $N_i$.}.

A less crude approximation than the usual P-CFG is to use a decision
tree to determine which aspects of the leftmost derivation have a
bearing on the probability of how node $i$ will be expanded.  In other
words, the probability distribution $p(r_i|t_i^-)$ will be modeled by
$p(r_i|E[t_i^-])$ where $E[t]$ is the equivalence class of the history
$t$ as determined by the decision tree.  This allows our probabilistic
model to use any information anywhere in the partial derivation tree
to determine the probability of different expansions of the $i$-th
non-terminal.  The use of decision trees and a large bracketed corpus
may shift some of the burden of identifying the intended parse from
the grammarian to the statistical estimation methods.

In this experiment, only a restricted implementation of this model is
explored, one in which only the path from the current node to the root
of the derivation along with the index of a branch (index of the child
of a parent) is examined to build equivalence classes of histories.

\subsection{The Grammar}

The grammar used in this experiment is a broad-coverage, feature-based
unification grammar. The grammar is context-free but uses unification
to express rule templates for the the context-free productions.  For
example, the rule template:
\begin{equation}
\left[ \begin{array}{c}
        pos=NP \\ number=x
       \end{array}
\right]
\rightarrow
\left[ \begin{array}{c}
        pos=Det \\ number=unspec
       \end{array}
\right]
\left[ \begin{array}{c}
        pos=N \\ number=x
       \end{array}
\right]
\end{equation}
where the second feature, $number,$ can take on three possible values:
singular, plural, or unspecified $(unspec).$ This rule template
corresponds to 3 different context-free productions.  The rule
template might also include a restriction on the values of $x$ based
on values of other features in the production.

The grammar has 21 features whose range of values maybe from 2 to
about 100 with a median of 8.  There are 672 rule templates of which
400 are actually exercised when we parse a corpus of 15,000 sentences.
The number of productions that are realized in this training corpus is
several hundred thousand.

For a complete description of this grammar, see Black, Garside, and
Leech\cite{blackbook}.

\subsubsection{P-CFG}

A non-terminal in the above grammar is a feature vector.  For the
purposes of parameterizing a P-CFG model of this grammar, several
non-terminals are grouped into a single class called a {\em mnemonic.}
A mnemonic is represented by the one non-terminal that is the least
specified in its class.  For example, the mnemonic VB0PASTSG*
corresponds to all non-terminals that unify with:

\begin{equation}
\left[ \begin{array}{c}
        pos=v \\ verb-type=be \\ tense-aspect=past
       \end{array}
\right]
\end{equation}

These mnemonics are used to label a parse tree and also to estimate a
P-CFG, where the probability of rewriting a non-terminal is given by
the probability of rewriting the mnemonic. From a training set, a CFG
is induced from the actual mnemonic productions that are elicited in
parsing the training corpus.  Using the inside-outside algorithm, a
P-CFG can be estimated from a large corpus of text.  Using the
bracketed sentences from the Lancaster treebank, the inside-outside
algorithm can be adapted to reestimate the probability parameters
subject to the constraint that only correct parses, i.e. parses whose
constituent structure matches the bracketing in the treebank,
contribute to the reestimation.  From a training run of 15,000
sentences, 87,704 mnemonic productions were observed, with 23,341
mnemonics of which 10,302 were lexical.  Running on a test set of 760
sentences, 32\% of the rule templates were used, 7\% of the lexical
mnemonics, 10\% of the constituent mnemonics, and 5\% of the mnemonic
productions actually contributed to parses of test sentences.

\subsection{The HBG Model}

Unlike the P-CFG model which assigns a probability to a mnemonic
production, the HBG model assigns a probability to a rule template.

For the HBG model, in place of the mnemonics, there are 50 syntactic
categories and 50 semantic categories.  Each non-terminal of the
grammar has been assigned a syntactic ($Syn$) and a semantic ($Sem$)
category.  Also, each rule production associates a primary lexical
head, denoted by $\mbox{H}_1$, and a secondary lexical head, denoted
by $\mbox{H}_2$, with each constituent.\footnote{The primary lexical
head $\mbox{H}_1$ corresponds (roughly) to the linguistic notion of a
lexical head.  The secondary lexical head $\mbox{H}_2$ has no
linguistic parallel.  It merely represents a word in the constituent
besides the head which contains predictive information about the
constituent.} When a rule is applied to a non-terminal, it indicates
which child will generate the primary lexical head and which child
will generate the secondary lexical head.

The HBG model estimates the probability of a parse tree and a
sentence, $p(T,S)$ as the product of the probabilities of every
constituent node in the sentence, where a constituent node contains a
unique value for $Syn$ and $Sem,$ a rule $R$ which is used to expand
the node, and a primary and secondary lexical head:
\beqn
p(T,S) = \prod_{[Syn,Sem,R,H_1,H2]\in T}
p([Syn, Sem, R, H_1, H_2] \given [Syn, Sem, R, H_{1}, H_{2}]_p, I_{pc}),
\eeqn
where $[]_p$ refers to the information at the parent of the current node
and $I_{pc}$ refers to the index of the node in its parent's list of
children (e.g. in VP$\rightarrow$~V~NP, the $I_{pc}$ of the V node is 0
and the $I_{pc}$ of the NP node is 1).

In this experiment, the probability of a node is approximated by the
following five factors:
\begin{enumerate}
\item $p$($Syn$ \given $R_p, I_{pc}, H_{1p}, Syn_p, Sem_p$)
\item $p$($Sem$ \given $Syn, R_p, I_{pc},H_{1p}, H_{2p}, Syn_p, Sem_p$)
\item $p$($R$ \given $Syn, Sem, R_p, I_{pc}, H_{1p}, H_{2p}, Syn_p, Sem_p$)
\label{rule}
\item $p$($H_1$ \given $R, Syn, Sem, R_p, I_{pc}, H_{1p}, H_{2p}$)
\item $p$($H_2$ \given  $H_1, R, Syn, Sem, R_p, I_{pc}, Syn_p$)
\end{enumerate}
While a different order for these predictions is possible, the
experiment used only this one.

The rule probability component of this model is the only model
estimated using decision trees.  The other four components are {\it
n-gram} models with the usual {\it deleted interpolation} for
smoothing.

Following the techniques described in Chapter~\ref{DTREEchapter},
binary classification trees are generated by hand for the syntactic
categories, the semantic categories, and the rule templates.  The
binary classification tree for the lexical head values are
automatically clustered using the bigram mutual information clustering
algorithm in Brown et.al.\cite{ngrams}.  Given the bitstring of a
history, a decision tree is grown to model the probability that a rule
will be used for rewriting a node in the parse tree.

Since the grammar produces parses which may be more detailed than the
treebank, the decision tree is built using a training set constructed
in the following manner.  Using the grammar with the P-CFG model, the
most likely parse that is consistent with the treebank is determined,
and the resulting sentence-tree pair is identified as an event.  Note
that the grammar parse will also provide the lexical head structure of
the parse.  Using the resulting data set a decision tree is grown by
classifying histories to locally minimize the entropy of the rule
template.

\subsection{Experimental Results}

To evaluate the performance of a grammar and an accompanying model,
two types of measurements are used:
\begin{itemize}
\item the {\it any-consistent} rate,
defined as the percentage of sentences for which the correct parse is
proposed among the many parses that the grammar provides for a
sentence.
\item the {\it Viterbi} rate defined as the percentage of sentences
for which the most likely parse is consistent.
\end{itemize}
The {\it any-consistent} rate is a measure of the grammar's coverage
of linguistic phenomena. The {\it Viterbi} rate evaluates the
grammar's coverage with the statistical model imposed on the grammar.
The goal of probabilistic modeling is to produce a {\it Viterbi} rate
close to the {\it any-consistent} rate.

The {\it any-consistent} rate of the grammar is 90\% when evaluating
consistency based on both structure and labels and 96\% when only
unlabeled brackets are considered.  These results are obtained on 760
sentences from 7 to 17 words in length.  The {\it parse base} of the
grammar is 1.35 parses/word.  The {\it parse base} of a grammar is
defined as the geometric mean of the number of proposed parses on a
per word basis, to quantify the ambiguity of the grammar.  This
translates to about 23 parses for a 12-word sentence.  The unlabeled
{\it Viterbi} rate of the P-CFG model stands at 64\% and the labeled
{\it Viterbi} rate is 60\%.

While the {\it Viterbi} rate is believed to be close if not the
state-of-the-art performance, there is room for improvement by using a
more refined statistical model to achieve the labeled {\it
any-consistent} rate of 90\% with this grammar.  There is a
significant gap between the labeled {\it Viterbi} and {\it
any-consistent} rates: 30 percentage points.

Using the HBG model as described above, the {\it Viterbi} rate is
increased to 75\%.  This is a reduction of 37\% in error rate.

\begin{figure}[htb]
\begin{center}
\begin{tabular}{rc}
 & Accuracy \\
\hline
P-CFG & 59.8\% \\
\hline
HBG & 74.6\% \\
\hline
Error Reduction & 36.8\% \\
\hline
\end{tabular}
\end{center}
\caption{Parsing accuracy: P-CFG vs. HBG}
\end{figure}

In developing the HBG model, the experiments also explored similar
models of varying complexity.  One discovery made during this
experimentation is that models which incorporated more context than
HBG performed slightly worse than HBG.  This suggests that the current
training corpus may not contain enough sentences to estimate richer
models.  Based on the results of these experiments, it appears likely
that significantly increasing the size of the training corpus should
result in a corresponding improvement in the accuracy of HBG and
richer HBG-like models.

\subsection{Conclusions from the HBG Experiment}

The HBG experiment provided further evidence that statistical models
could be used to deal with ambiguity resolution in natural language
parsing.  Further, it showed that significant amounts of lexical
information could be incorporated into these models, although it also
suggested that more data would be necessary to realize the full
disambiguating potential of the lexical information.

This experiment also tested the application of decision tree
technology to the parsing modeling problem.  These algorithms had
previously been applied to language modeling and speech recognition
problems, but they had never been attempted on a problem as varied and
complex as the parsing problem.  The HBG experiment confirmed the
hypothesis that decision tree models could provide effective estimates
of a probabilistic parsing model.

Given the encouraging results of the {$\cal P$}earl model and HBG
model, the stage was set to try a far more ambitious experiment.  Both
of these experiments attempted to find models which could select a
correct parse from the limited set of parses generated by an elaborate
context-free grammar.  But, if there were no grammar at all, could a
statistical model select a correct parse from the space of {\em all}
possible parse trees?  That is, could a statistical model be trained
to select a correct parse tree for a sentence from the space of all
{$n$}-ary labeled trees with the same number of leaves as words in the
sentence?  This is the question addressed by remainder of this
dissertation.

\chapter{SPATTER Parsing\label{SPATTERchapter}}

The previous chapter recounted an experiment in which combining an
elaborate rule-based grammar with decision tree models succeeded in
improving the accuracy of a statistical parser.  In that experiment,
the grammar serves as a filter for unreasonable parses, based on the
grammarian's knowledge about the language.  The decision tree models
combine to rank the remaining analyses.

But why use the grammar to filter out parse trees in the first place?
If the statistical models are accurate enough to distinguish correct
grammatical parses from incorrect grammatical parses, certainly they
should be able to discern correct grammatical parses from ludicrous
ungrammatical ones.

There are good reasons to question the necessity of highly restrictive
grammars for parsing when statistical information is available, some
of which were explored in the experiments discussed in
Chapter~\ref{PRELIMchapter}.

Another strong motivation for eliminating complicated rule systems
from parsing is the enormous investment of time and resources in
grammar development without any guarantee of success.  To illustrate
this point, let us examine the grammar writing process:

\begin{quotation}
A grammarian examines sentences, skillfully extracts the
linguistic generalizations evident in the data, and writes grammar
rules which cover the language.  The grammarian then evaluates the
performance of the grammar, and upon analysis of the errors made by
the grammar-based parser, carefully refines the rules, repeating this
process on and off for about a decade.
\end{quotation}

For the best and most accurate robust rule-based grammars, a decade is
not an exaggeration of the time necessary to complete a grammar.  The
grammar refinement process is extremely time-consuming and difficult,
and has not yet resulted in a grammar which can be used by a parser to
analyze accurately a large corpus of unrestricted text.  Instead of
writing grammars, one can develop corpora of hand-analyzed sentences
({\em treebanks}) with significantly less effort.  With the
availability of treebanks of annotated sentences, one can view NL
parsing as simply {\em treebank recognition.}

In this chapter, I introduce an approach to parsing natural language
which restates the parsing problem as pattern recognition, where the
pattern to be recognized is a linguistic analysis of a sentence.  This
approach divides the parsing problem into two separate tasks: {\em
treebanking,} defining the annotation scheme which encodes the
linguistic content of the sentences and applying it to a corpus, and
{\em treebank recognition,} generating these annotations automatically
for new sentences.

The annotation scheme used to represent the linguistic content of the
sentence is now the primary contribution of the grammarian (although
as we will see much later, the grammarian has another important role
to play).  The treebank can contain whatever information is deemed
valuable by the grammarian, as long as it can be applied consistently
and efficiently to a large number of sentences.  If one views the
treebank as a black box which, given a sentence, assigns a unique
analysis to that sentence, then the goal of treebank recognition is to
produce the same analysis of a sentence that the treebank would
generate.

The solution to the treebank recognition problem described in the
remainder of this dissertation is called SPATTER (Statistical PATTErn
Recognizer).  The SPATTER parser uses probabilistic models to predict
the part-of-speech labels, parse tree edges, and constituent labels
for a given sentence.  The parsing process is divided into a sequence
of actions which add structure to the sentence in a bottom-up fashion.
The probabilistic models used to predict each action condition their
distributions on the information made available by the sequence of
actions preceding the current action.  Thus the derivational order of
the actions leading to an analysis affects the probability assigned to
the parse.  In Chapter~\ref{PRELIMchapter}, this type of model is
referred to as a history-based grammar model.  A derivational model,
which assigns a distribution to the possible derivations of a parse
tree, is self-organized in the process of training the other models.
The probabilistic models in SPATTER are estimated using statistical
decision trees, and these distributions are refined using two
different applications of the expectation-maximization (E-M)
algorithm.

In this chapter, I define and motivate the treebank recognition
problem (TRP), discuss the knowledge representation issues involved in
SPATTER parsing, and describe in detail the statistical models used in
this work.


\section{The Treebank Recognition Problem}

The purpose of introducing the treebank recognition problem (TRP) is
to remove a significant obstacle impeding progress in the area of
natural language (NL) parsing.  One of the most difficult aspects of
NL parsing is that, to the casual observer, or to a dedicated parsing
specialist for that matter, it is not clear exactly what the parsing
problem is.  To wit, parsing has something to with assigning syntactic
constituent structure to sentences.  But this simplistic definition
does little more than replace one unknown term (parsing) with another
(syntactic constituent structure).  Disregarding the issue of what
``syntactic constituent structure'' means, many would argue that a
parse is much more than syntactic structure alone.  To some, parsing
involves assigning semantic labels to constituents as well.  Still
others expect parse trees to indicate predicate-argument relations,
and to resolve anaphoric and elliptical references.  To make the task
even more difficult, linguists have divergent views on the definitions
of semantics and predicate-argument relations.  It is difficult to pin
down precise definitions to many of the features of a sentence that
parsers are expected to identify.  In summary, the natural language
parsing problem is an ill-defined task.

In contrast, the TRP is {\em posed} as a well-defined task.  The
annotation scheme used to analyze the treebank sentences {\em defines}
the expected output of the parser.  For instance, if a linguist can
consistently annotate sentences with predicate-argument structure,
then it is the goal of the treebank recognition problem to identify
that structure in new sentences accurately.

The most significant advantage the TRP has over the parsing problem is
in evaluation of different approaches and implementations.  For a
given treebank, divided into training data and unseen test data, the
best solution to the TRP is the one which assigns THE correct analysis
to the most test sentences.  Comparative evaluation of solutions to
the parsing problem has been hindered by the claim that there is no
one {\em correct} parse, i.e.  any reasonable analysis of a sentence
is correct, where a ``reasonable'' analysis is one supported by any
one of the diverse and numerous linguistic theories available on a
given day.  The TRP explicitly rules out this argument.  For better or
for worse, the one and only {\em correct} parse of sentence is the one
which the treebank assigns to that sentence.  Of course, the treebank
may have internal inconsistencies. While statistical training methods
can overcome some inconsistencies in the training data, annotation
schemes which border on the random will be difficult to reproduce
accurately.  And errors in the test data are unavoidable.  But,
assuming the description task, as discussed in the introduction,
results in a labeling scheme which can be consistently assigned to
text by humans, the inconsistencies in both the training and testing
sentences should be negligible.


\section{Representing Parts of Parse Trees}

The initial development of SPATTER was guided by the principle that
the best representation to start with is the most straightforward.
The rationale behind this principle is that, if experiments based on a
straightforward representation failed, error analysis might reveal
obvious flaws in the representation which could be corrected.  But, if
solving the TRP is reduced to a search for the best representation of
a parse tree, then there is little hope for success.

\subsection{Four Elementary Features}
The treebank can be viewed as a collection of n-ary branching trees,
with each node in a tree labeled by either a non-terminal label, a
part-of-speech label, or a word token.  For the purposes of pattern
recognition, these trees must be described precisely, in terms of
elementary components.

\begin{centering}
\begin{figure}[tbhp]
\centerline{\psfig{figure=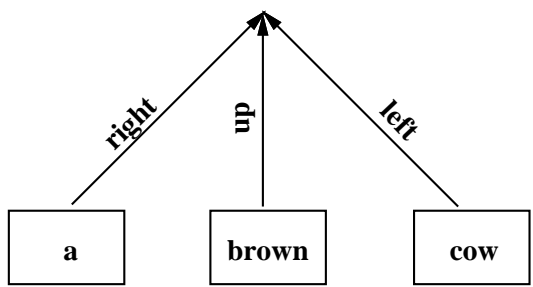}}
\caption{Representation of constituent and labeling of extensions in
SPATTER.\label{abrowncow}}
\end{figure}
\end{centering}

Grammarians elevate constituents to the status of elementary units in
a parse, especially in the case of rewrite-rule grammars where each
rewrite rule defines a legal constituent.  If a parse tree is
interpreted as a geometric pattern, however, a constituent is no more
than a set of edges which meet at the same tree node.  For instance,
the noun phrase, ``a brown cow,'' consists of an edge extending to the
right from ``a,'' an edge extending to the left from ``cow,'' and an
edge extending straight up from ``brown'' (see figure
\ref{abrowncow}).

The parse tree is encoded in terms of into four elementary components,
or {\em features}: words, tags, labels, and extensions.\footnote{The
idea of decomposing the components of a parse tree into words, tags,
labels, and extensions was originally proposed by Fred Jelinek, Bob
Mercer, and Salim Roukos.}  Each feature has a fixed vocabulary, with
each element of a given feature vocabulary having a unique
representation.

The input to the treebank recognizer is a sentence, which is a
sequence of tokens.  The first feature, the {\em word} feature,
defines the mapping from these tokens onto a fixed vocabulary of
words, including an ``unknown'' word for all tokens which do not map
onto a word in the vocabulary.  The mapping is almost trivial, with
most tokens mapping to the vocabulary item with the same spelling.
However, because the capitalization in a sentence sometimes contains
useful information, e.g. in titles, and sometimes doesn't, e.g. at the
beginning of a sentence, the capitalization of the tokens is altered
according to a deterministic algorithm before the tokens are mapped
onto the word vocabulary.

The next feature assigned, the {\em tag} feature, represents the
part-of-speech information for each word in the sentence.  The tag
vocabulary is defined to be the set of all part-of-speech labels which
are assigned to words in the treebank training data.

Once a leaf node has been assigned a tag feature value, it is extended
in some direction towards a parent node.  The direction of this
extension is represented by the {\em extension} feature.  This feature
takes on one of 4 possible values, {\em left, right, up,} or {\em
unary}.  A {\em left} extension value corresponds to the initial node
of a constituent, and a {\em right} value corresponds to the final
node of a constituent.  The {\em up} value is assigned to the nodes
which are between the initial and final nodes of a constituent.  The
{\em unary} value means that the parent of the node has only one
child.  So, in the noun phrase, ``a brown brown brown brown cow,'' the
nodes corresponding to all of the {\bf brown}'s in the sentence are
assigned the extension value {\em up}.\footnote{Since each node of the
parse tree must be assigned an extension value, there must be a fifth
extension value for the root node.  I call this value {\em root}.
However, since the root node of each parse tree is assigned the same
label ({\bf GOD}) and the label value is assigned before the extension
value, the {\em root} feature value is actually redundant.  It is
assigned to a node with the {\bf GOD} label with probability 1.
Therefore, discussions of the extension feature omit mention of the
{\em root} value.}

A constituent is defined in terms of the extension feature values of
consecutive nodes.  A node with extension value {\em right} followed
by any number of nodes with extension value {\em up} followed by a
node with extension value {\em left} is a constituent.  Any node which
is assigned the {\em unary} extension value is also a constituent.
Whenever the parser detects a sequence of nodes which corresponds to a
constituent, it generates a new parent node and begins the process of
assigning feature values to that node.  The first feature value
assigned to an internal node in a parse tree is a {\em label} feature
value, which encodes information about the constituent beneath that
node.

\begin{centering}
\begin{figure}[tbhp]
\psfig{figure=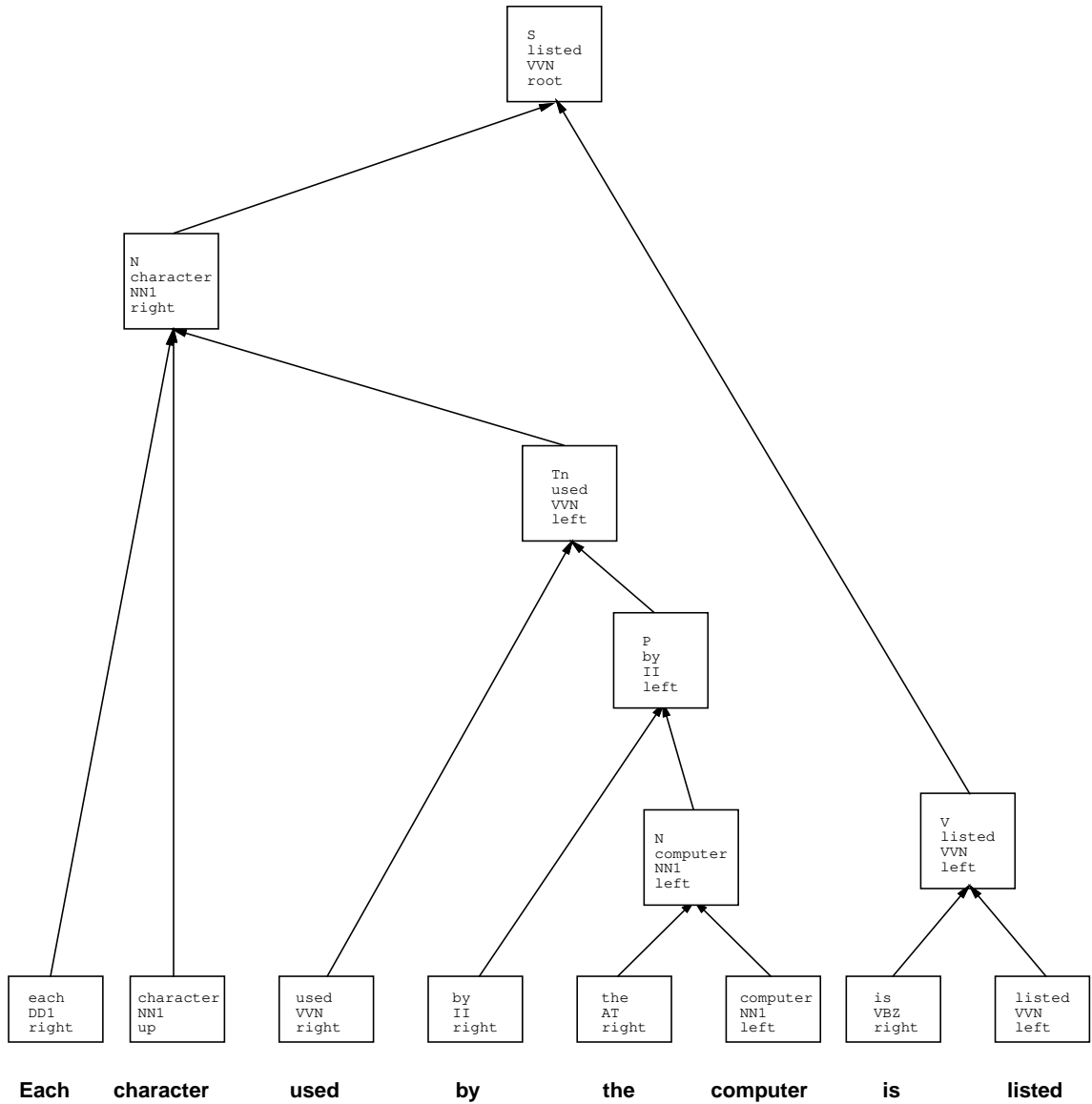,width=6in}
\caption[Treebank analysis encoded using feature values.]{Treebank
analysis encoded using feature values. Each internal node contains,
from top to bottom, a {\em label, word, tag,} and {\em extension}
value, and each leaf node contains a {\em word, tag,} and {\em
extension} value.\label{fullparse}}
\end{figure}
\end{centering}

\subsection{Propagating Lexical Information}

The constituent labels in a treebank are not very descriptive.  It
would be difficult for human treebankers to assign constituent labels
to text consistently, accurately, and efficiently if the labels
conveyed subtle nuances of meaning.  For example, the Lancaster
Treebank uses only 17 constituent labels, whereas the grammar
described in Black, Garside, and Leech\cite{blackbook} covering the
same domain assigns over 13,000 unique non-terminal categories.

Given the dearth of information in the label set, it would be
difficult to model the assignment of feature values by asking
questions primarily of the {\em label} values.  Questions about the
words and part-of-speech tags are more informative, but it is
infeasible, and not necessarily even useful, to ask questions about
all {\em word} and {\em tag} values from a constituent.

To augment the information available at constituent nodes, a word,
along with its corresponding part-of-speech tag, is selected from each
constituent to act as a lexical representative.  Thus, at each
internal node in the parse tree, there are {\em word} and {\em tag}
feature values which are deterministically assigned, as well as {\em
label} and {\em extension} values which are predicted by
models.

The lexical representative from a constituent loosely ({\em very}
loosely) corresponds to the linguistic notion of a head word.  For
example, the lexical representative of a noun phrase is the rightmost
noun, and the lexical representative of a verb phrase is the leftmost
non-auxiliary verb.  However, the correlation to linguistic theory
ends there.  The set of deterministic rules which select the
representative word from each constituent, called the {\em Tree Head
Table}, was developed in the better part of an hour, in keeping with
the philosophy of avoiding excessive dependence on rule-based methods.
If the performance of these methods depend on the precise word
selection rule set, there is little hope of success.

\begin{centering}
\begin{table}[tbhp]
\begin{verbatim}
Nr right-to-left Nr NNT1 NNT2 RR ...
V  left-to-right V VV0 VVC VVD VVG ...
N  right-to-left N NN NNJ NNU NP NN2 ... J ... Nn
S  right-to-left S V Ti Tn Tg N J Fa ...
Tg right-to-left Tg VVG VBG VDG VHG V
Ti right-to-left Ti VVI VDI VVN VDN VHI VHD VBI V TO
Tn right-to-left Tn VVN VDN VHD V
...
\end{verbatim}
\caption[Subset of the Tree Head Table.]{Subset of the Tree Head
Table, which determines the lexical representative from each
constituent, based on the label of the constituent and the tags and
labels of the elements of the constituent.\label{TreeHeadTable}}
\end{table}
\end{centering}

Table~\ref{TreeHeadTable} presents a subset of the mapping rules in
the Tree Head Table used in SPATTER.  The first column is the label of
the constituent whose word and tag features are being assigned.  The
second column indicates whether the children of this node are
processed left-to-right or right-to-left.  The remainder of each line
consists of an ordered list of tag and label values which might occur
as the children of the constituent label in the first column.  The
elements in this list are assigned priorities, with the first element
getting priority 1, the second element priority 2, and so on.  The
Tree Head assignment algorithm identifies the lexical representative
of a constituent as the lexical representative of the child whose
label (or tag if the child is not a constituent) has the lowest
priority value.  In the case of two children having the same priority
value, if the parent label is marked as left-to-right, then the
leftmost child is selected, otherwise the rightmost child is selected.

As an example, let's step through the Tree Head assignment algorithm
for the sentence:
\begin{verbatim}
[S [N I_PN N] [V really_RR [V like_VBZ [N ice_NN cream_NN N] V] V] S]
\end{verbatim}
Initially, each node is assigned its corresponding word value, e.g.
node 1 is assigned ``I,'' node 2 is assigned ``really,'' etc.
Working bottom-up, the first tree head assignment is for the noun
phrase, ``I.''  Since the noun phrase has only one child, and it's
tag, {\bf PN}, occurs in the priority list for {\bf N}, ``I'' is
selected as the lexical head.  Note that if {\bf PN} did not occur in
the priority list, then there would be no lexical head for the
constituent.  Now, consider the noun phrase ``ice cream.''  Both
``ice'' and ``cream'' are potential head's of this noun phrase, and
both have the same priority value.  According to the Tree Head Table,
{\bf N}'s are processed right-to-left, so the rightmost {\em NN} is
selected as the lexical head, in this case ``cream.''  The algorithm
selects ``like'' as the head of ``like ice cream,'' and also selects
``like'' as the head of ``really like ice cream.''  Finally, it
selects ``like'' as the head of the sentences, since {\bf V} has a
lower priority value than {\bf N} for constituents labeled with {\bf
S}.

To illustrate the mapping between a treebank parse and the feature
value representation scheme, consider the following sentence from the
IBM computer manuals domain, as annotated in the Lancaster Treebank:

\begin{verbatim}
[N Each_DD1 character_NN1
   [Tn used_VVN [P by_II [N the_AT computer_NN1 N] P] Tn] N]
[V is_VBZ listed_VVN V] ._.
\end{verbatim}

The node feature value representation of this treebank analysis is
shown in Figure~\ref{fullparse}.

\subsection{Representing Features as Binary Questions}

In SPATTER, a combination of binary statistical decision tree (SDT)
models assign a probability distribution on the space of possible
parse trees.  The exact form of these models is discussed in a later
section, but the use of binary SDTs brings up an interested
representation problem.  Binary SDTs require that the contexts, or
histories, be represented in terms of answers to binary questions.
Questions like ``Are the words {\em and} or {\em or} in the
sentence?'' have convenient binary answers.  But questions like ``What
is the word at the current node?'' or ``How many children does the
current node's constituent have?'' have more than 2 answers.  This
problem is solved by creating a {\em binary classification tree} for
the answers to each question.\footnote{This work contains references
to three different types of trees: parse trees, decision trees, and
classification trees.  It is important to keep in mind which type of
tree is being discussed.  The parse trees are constructed using
decision tree models.  The answers to the questions asked by the
decision tree models are encoded using classification trees.  Now, if
you were a tree, which kind would you be?}

\subsubsection{Binary Classification Tree Feature Representations}

SDTs ask questions about the contextual history which help them
predict the future more accurately, i.e. which reduce the entropy
(uncertainty) of the future.  In the context of SPATTER, these
questions are about the feature values of nearby nodes in the parse
tree.  To simplify the mathematical algorithms used in training SDTs,
it is very useful to enforce a binary structure on the decision trees.
In a binary decision tree, only binary questions can be asked of the
history.  Of course a question which has four possible answers can be
encoded as a sequence of binary questions.

\begin{centering}
\begin{figure}[tbhp]
\centerline{\psfig{figure=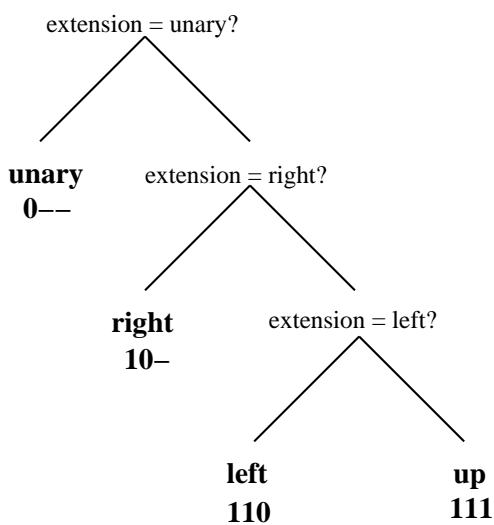}}
\caption[BCT for extension feature values with three
questions.]{Classification tree for extension feature values with
three binary questions.  The hyphen (-) in the binary encoding
indicates that the value of that bit can be either 0 or 1.  The value
of the bit in those cases can be selected to maximize the information
content of that bit with respect to the value being
predicted.\label{threequestions}}
\end{figure}
\end{centering}

\begin{centering}
\begin{figure}[tbhp]
\centerline{\psfig{figure=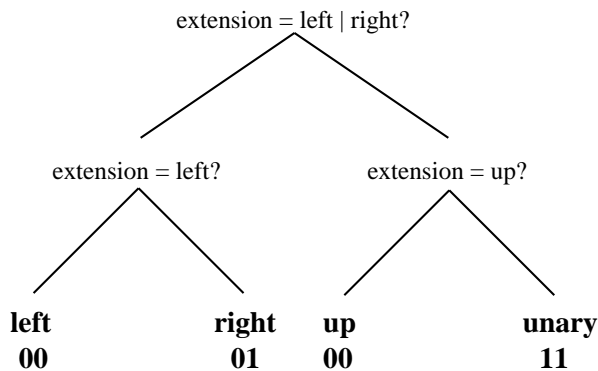}}
\caption[BCT for extension feature values with two
questions.]{Classification tree for extension feature values with two
binary questions.\label{twoquestions}}
\end{figure}
\end{centering}

Consider a 4-valued question which asks about the an extension feature
value.  Without the binary question constraint, one can ask a question
like ``What is the extension feature value of the parse node to the
left of the current node?''  With this constraint, one can ask the
same question using the sequence of questions shown in figure
\ref{threequestions}.  However, other sequences of questions might be
more predictive of the future.  For instance, one might want to ask if
the feature value is {\em left} or {\em right} first, indicating a
constituent boundary, as shown in figure \ref{twoquestions}.

Each of these sequences of questions corresponds to a binary
classification tree on the extension vocabulary, with each extension
feature value assigned a binary representation.  The word, tag, and
label feature vocabularies are also assigned binary representations
using the same scheme.

Ideally, the classification trees used to represent the features
should be constructed so that the set of values beneath each node in
the classification tree provide information about the future.  For
instance, if one were trying to predict the label of a constituent by
asking about the extension value of the node to the right of that
constituent, then the representation for extensions suggested by
figure \ref{threequestions} would be particularly useful if, in
general, the distribution of the labels of constituents next to {\em
unary} constituents are significantly different from those next to
nodes extended in any other direction --- either to the {\em left}, to
the {\em right,} or {\em up.}

While bit representations based on distributional behavior are ideal,
it is computationally expensive to create representations which
produce this attractive predictive behavior.  Previous work on n-gram
class models (Brown et.al.\cite{ngrams}) illustrates a method for
distributionally clustering words based on bigram mutual information.
However, this clustering method generates classes which reduce
uncertainty about the next word, but do not necessarily reduce
uncertainty about syntactic structure in a sentence.  It is not clear
how to extrapolate these methods to clustering constituent labels and
extensions, since these feature values cannot easily be ordered, from
left to right, as bigrams in the same way as words or part-of-speech
tags in a sentence.  Also, since a given node feature value is used to
predict all four feature values elsewhere in the parse tree, it would
be best to produce different binary representations for {\em each}
type of decision made by the models.

In the interests of expediency, the classification trees for the
word and tag features are generated using the mutual information
clustering algorithm, and the remaining feature value classification
trees are generated by hand.  For the hand-generated classes, some
redundancy is encoded in the bit representations, with the hope that
each statistical decision tree will find some sequence of questions to
ask which will predict the future well.

\subsubsection{More Questions?}

The previous section discusses predicting node feature values by
asking binary questions about feature values of nearby nodes.
However, this information does not easily capture generalizations
about the local context.  For instance, the number of children nodes
in a constituent is useful when assigning a constituent label, but
this information isn't readily accessible from the node feature values
without asking a number of questions about the specific {\em
extension} values of the node's children.  Also, the span of a
constituent, i.e. the number of words contained within the
constituent, also might alter the distributions of {\em label} and
{\em extension} feature value assignments.  This information is
extremely difficult to extract by asking only about feature values.

There are many such questions about the structure of the parse tree
which might provide predictive information for the decision tree to
ask about.  However, using the principle that the best representation
to start with is the most straightforward, only the most obvious
aspects of a parse tree are encoded as questions: the number of
children of a node, the span of a node, and the number of constituent
nodes in the sentence.\footnote{Note that the answer to this final question,
about the number of constituent nodes, changes as the parser adds
structure to the sentence.}

\subsection{A Feature for Conjunction}

After analyzing preliminary parsing results, it was clear that the
SPATTER was not able to predict the scope of conjunctions very
accurately.  The precise reason is not clear, but a contributing
factor is that there is no single feature which represents the fact
that two or more constituents will be conjoined.  And while there are
tell-tale signs that a conjunction is going to occur in a sentence,
e.g. the words {\em and} or {\em or}, or a comma (,) or semi-colon
(;), the cues for the boundaries of the conjuncts are not obviously
encoded in the four feature values or the other questions described
above.\footnote{Now try to parse that sentence automatically!}

Coincidentally, the Lancaster Treebank annotation scheme encodes
information about conjunction which was not used in the initial
experiments.  The treebank indicates which constituents are conjoined
by appending a symbol to the end of the constituent label.  This
information can be viewed as one bit of information attached to each
constituent, in other words, a fifth feature of the parse tree.  This
binary-valued node feature, the {\em conjunction} feature, takes on a
true value only when the constituent node is going to be a part of a
conjoined phrase.

\section{Derivations of Parse Trees}

Since the statistical models in SPATTER are sensitive to the order in
which feature values are assigned, it would be ideal to consider all
possible orderings of feature value assignments.\footnote{The idea of
using derivations of a parse tree was originally proposed by Fred
Jelinek, Bob Mercer, and Salim Roukos.}  However, with five feature
values assigned at each internal node and four feature values assigned
at each leaf, there are far too many different derivations of a single
parse tree to consider.  If derivations were completely unconstrained,
a parse tree with 20 nodes would have over $10^{150}$ (over 100
factorial) possible derivations.

To constrain this space, the only derivations which will be
considered are bottom-up derivations.  In a bottom-up derivation, a
node is not constructed until all of the node feature values beneath
it are assigned.  One simple derivation is the bottom-up leftmost
derivation, in which the parsing process begins with the leaf nodes,
and feature values are filled out in order from left to right.  But
there are many other derivations which might allow more predictive
information to be available at crucial points in the parsing process.
For instance, the extension feature value of a noun phrase node which
is sandwiched between a verb and a prepositional phrase determines the
prepositional phrase attachment.  In the bottom-up leftmost
derivation, the noun phrase extension would be predicted before any
information was known about the prepositional phrase and its argument
noun phrase.  In other derivations both constituent nodes would be
within the extension model's five node window.  On the other hand, not
all derivations provide information by deviating from the bottom-up
leftmost derivation.  Predicting the part-of-speech tag of the last
word in a long sentence is unlikely to affect the distribution of the
part-of-speech tag of the first word in the sentence.

To further constrain the number of derivations while still allowing
informative derivations to be considered, a {\em derivational window
constraint} (DWC) is used.\footnote{The idea of a derivational window
constraint was first suggested by Fred Jelinek, and the idea was
developed by myself, Jelinek, Bob Mercer, Salim Roukos, Adwait
Ratnaparkhi, and Barbara Gates.} Under a DWC of size $n,$ the first
$n$ possible feature value assignments, from left to right, are
considered.  The nodes at which these candidate feature value are to
be assigned are called {\em active} nodes.

As an example, at the start of the parsing process, when only the word
feature values are assigned, a window constraint of 3 would allow any
one of the first three part-of-speech tags to be assigned, and thus
the first three nodes in the sentence are active.  Let's say the
second word is tagged.  Then, the window constraint permits either the
first or third word to be tagged or the second word to be extended,
and again the first three nodes are active.  The fourth node in the
sentence does not become active until one of the first three nodes is
completed (i.e. all of its feature values are assigned).  When a
sequence of completed nodes are combined to form a new constituent
node, the new node is then active again.

Even with a DWC of 2 applied to all of the features, there are too
many possible derivations to consider.  To make the number of
derivations manageable, the DWC is set to 1 for all but the extension
and tag feature assignments.  This means that initially the tag
feature value can be assigned for the first 2 words.  However, as soon
as an internal node is created (i.e. a constituent is recognized), the
label and conjunction feature values for that node are assigned
immediately, without considering other derivational possibilties.

It is difficult to explain how the window constraint works in any more
detail without stepping through an example.  The DWC will be made more
clear in the next section, when the SPATTER parsing algorithm is
applied to an example sentence.

\section{SPATTER Parsing: The Algorithm}

\begin{centering}
\begin{figure}[tbhp]
\begin{verbatim}
create parse node array INITNODE
let Word feature value for INITNODE[i] = ith word in sentence
create sets of node arrays STATES, PARSES
add INITNODE to STATES
while (STATES is non-empty)
  remove a node array NODE from STATES
  if (NODE array has only one node and NODE[0] is completed)
    add NODE to PARSES
  else
    let FEATURE[i] = the next unassigned feature for NODE[i]
    let ACTIVE[i] = number of active nodes to left of ith node
    if for some X, ((ACTIVE[X] < DWC) and
                    ((FEATURE[X] == Label, Word, or Conj) or
                     (NODE[X] is an internal node and
                      FEATURE[X] == Tag)))
      AssignFeature(FEATURE[X], NODE, X)
    else
      for X from 0 to DWC-1
        AssignFeature(FEATURE[X], NODE, X)

Procedure AssignFeature(FEATURENAME, NODE, X):
  for each value V for feature FEATURENAME
    let NEWNODE = copy of NODE
    set value of FEATURENAME to V for NEWNODE[X]
    if ((NEWNODE[X] is completed) and
        (<NEWNODE[X-A], ..., NEWNODE[X+B]> forms a constituent))
      replace <NEWNODE[X-A], ..., NEWNODE[X+B]>
              with an empty node whose children are
              NEWNODE[X-A], ..., NEWNODE[X+B]
    add NEWNODE to STATES
\end{verbatim}
\caption[The SPATTER parsing algorithm.]{The SPATTER parsing
algorithm.  When the algorithm terminates, the set of complete parse
trees for the input sentence is in PARSES.\label{spatalg}}
\end{figure}
\end{centering}

In SPATTER parsing, a parse tree is represented as a connected,
single-rooted graph with feature values at each node, one of which
indicates the geometry of the parse (the {\em extension} feature).
Since the probability of a feature value assignment at a particular
node is conditioned on the information available at other nodes in the
partially-constructed tree, the probability of a parse tree derived in
a certain order is different from the probability of the same parse
tree derived in a different order.

\subsection{An Example\label{EXAMPLEsection}}

In this section, I step through the parsing algorithm in
Figure~\ref{spatalg} using the sentence from Figure~\ref{fullparse}:
\begin{verbatim}
each character used by the computer is listed
\end{verbatim}

First, the initial parse node array INITNODE is allocated with eight
nodes, one for each word in the sentence.  The node feature value of
the $i$th node of INITNODE is set to the $i$th word in the sentence,
with the word {\bf each} assigned to INITNODE[0], {\bf character}
assigned to INITNODE[1], etc.  This initial state is added to the
STATES set.

The algorithm's main loop begins by trying to advance this initial
state, now called NODE.  Since NODE is not completed, the algorithm
tries to advance the first two active nodes of NODE.  The FEATURE
array is set to the value $\{$Tag, Tag, Tag, Tag, Tag, Tag, Tag, Tag$\}$ and
the ACTIVE array is set to the value $\{0,1,2,3,4,5,6,7\}.$ Using
$DWC=2,$ the tag feature for NODE[0] and NODE[1] are assigned.  This
means that for all possible tags $t$, a new state is generated with
the tag feature value of NODE[0] set to $t$, and another new state is
generated with the tag feature value of NODE[1] set to $t.$ Since
there are 196 part-of-speech tags, a total of 392 new states are
generated in this step.  All of these states are added to the STATES
set.

The main loop continues with one of these states.  The order in which
states are expanded are determined by the stack decoder algorithm,
described in section~\ref{STACKDECODEsection}.  For the sake of the
example, assume that the next state extended is the state which has
the tag feature value for NODE[1] set to the correct tag, {\bf NN1.}
For this state, the FEATURE array is set to the value
$\{$Tag, Extend, Tag, Tag, Tag, Tag, Tag, Tag$\}$ and the ACTIVE array is set
to the value $\{0,1,2,3,4,5,6,7\}.$ For this state, the tag feature
value of NODE[0] and the extend feature value of NODE[1] are assigned.
Since there are 4 possible extension values, 200 new states are
generated in this step.

Consider the four states added in this step which had their NODE[1]
extend feature value assigned.  Using the terminology of the
AssignFeature procedure in Figure~\ref{spatalg}, the NEWNODE[1] for
these four states is completed.  However, only one of these states
contains a sequence of completed nodes which forms a constituent.  The
state for which the NEWNODE[1] extend feature was assigned the value
{\bf unary} contains a constituent consisting of the word {\bf
character.} For this state, a new parse node is created with
unassigned feature values, and this new empty node replaces
NEWNODE[1], with the old completed node becoming this node's child.

Consider what happens when the state being advanced is the one where
the tag feature value of NODE[1] is set to {\bf NN1} and the extension
feature value of NODE[1] is set to {\bf up.} In this case, the FEATURE
array is set to the value $\{$Tag, NONE, Tag, Tag, Tag, Tag, Tag, Tag$\}$ and
the ACTIVE array is set to the value $\{0,0,1,2,3,4,5,6\}.$ Since
NODE[1] is completed, the active nodes which are advanced in this step
are NODE[0] and NODE[2].

\subsection{Managing the Search: Stack Decoding\label{STACKDECODEsection}}

\begin{figure}[tbhp]
\centerline{\psfig{figure=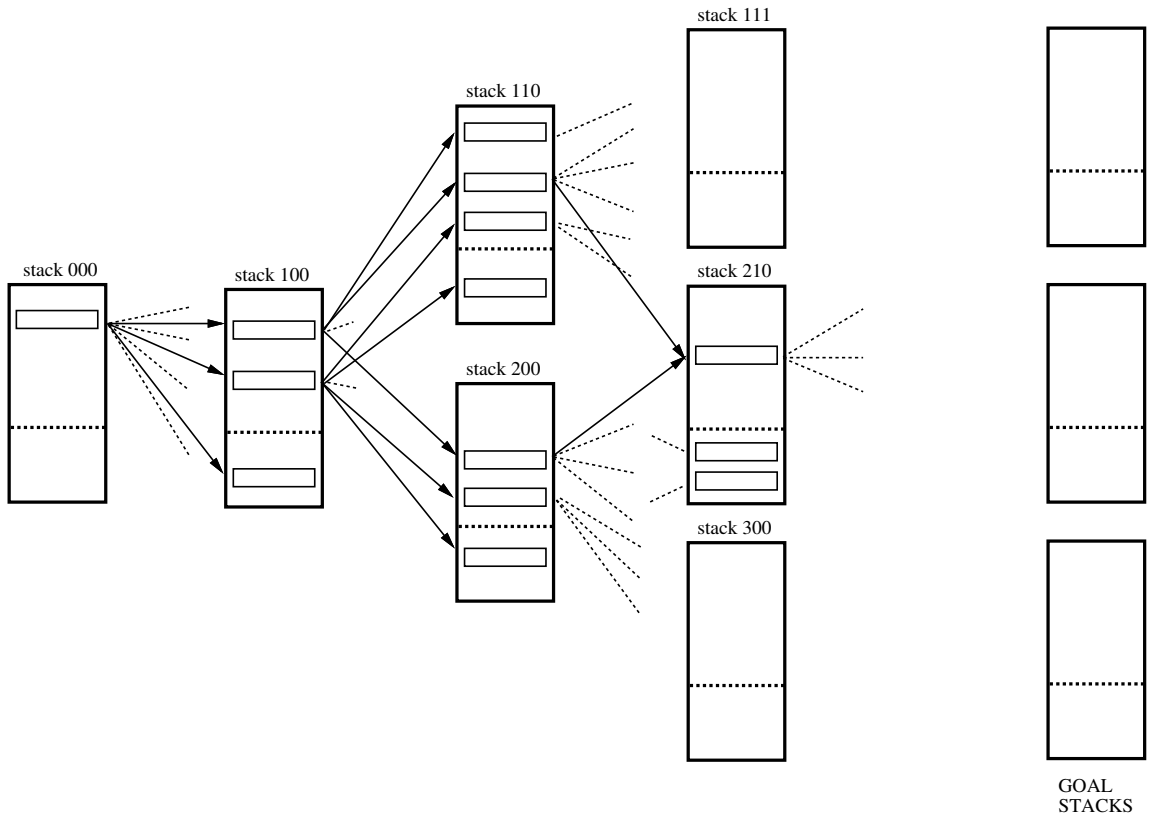,width=6in}}
\caption{Example stack search space for SPATTER stack decoding
algorithm.\label{Stacks}}
\end{figure}

In SPATTER parsing, a state is defined as a sequence of {$n$}-ary
labeled trees which together span the sentence.  Since all possible
feature values are generated at each node with some probability, the
search space for this parser is immense.  Even with the heuristic
constraints, such as the DWC and the Tree Head Table, the search space
is still far too large to search exhaustively.  To complicate matters
further, because the parser pursues different derivations of the same
parse, the state space is a graph instead of a tree.

The search algorithm which SPATTER uses to explore the search graph
and prune intelligently is the stack decoding algorithm.  The stack
decoding algorithm was introduced by Jelinek in 1969 \cite{Jelinek:69}
for solving the graph search problem in speech recognition.  Other
discussions of stack decoding in speech recognition are found in Bahl
et.al.\cite{Bahl:83} and Paul\cite{Paul:90}.

Like other AI search procedures, the stack decoding algorithm uses a
scoring function to evaluate a state based on the path from the
initial state.  In SPATTER, the state evaluation function is the
product of the probability of each decision along the path to that
state, according to the decision tree models.  Since the stack
decoding algorithm is a {\em graph} search algorithm, the state
evaluation function must provide a mechanism for evaluating states
with multiple paths from the initial state.  In some problems, it
makes sense to assign such a state the value of the maximum or minimum
value path.  However, since SPATTER includes a derivational model, the
probability, and thus the score, of a state with multiple derivations
is the sum of the probabilities of all of the derivations of the
state.

Unlike standard tree search algorithms, the stack decoding algorithm's
scoring function does not attempt to provide a total order on the
state space.  Since different states result from different numbers of
decisions, the probability of a state with very little structure would
generally be higher than that of a nearly completed parse tree.  Thus,
a probability-based search procedure would nearly exhaustively explore
the state space nearest to the initial state before expanding states
near the goals.

The stack decoding algorithm avoids this problem by using a {\em stack
index function} to assign each state to a stack.  State probabilities
are only compared to one another if states are assigned to the same
stack.  The stack index function used in SPATTER reflects the number
of tagging, labeling, and extension decisions made in that state.
Thus, states will only be compared to one another if they have been
constructed by the same number of tagging, labeling, and extension
decisions.

The stack search space in SPATTER parsing is illustrated in
Figure~\ref{Stacks}.  The initial state, as described in
section~\ref{EXAMPLEsection}, is assigned to stack 000, reflecting the
fact that no decisions have been made in that state.  Since the only
action permitted in the initial state is a tagging decision, every
state generated by the initial state is assigned to stack 100.  States
in stack 100 can be continued with either a tagging action or a
labeling action, so states extended from stack 100 states can be
assigned to either stack 200 or stack 110, etc.  Note in the figure
that a state in stack 210 has been generated by a state in stack 110
and from another state in stack 200.

The most important aspect of the stack decoding algorithm is its
method of pruning.  It prunes states by assigning each state a
threshold based on the scores of the states which are {\em alive} in
each stack.  The health of a state is determined by the status of its
progeny, and a stack thresholding parameter, $\lambda.$ Each stack is
assigned a parameter $\lambda$ between 0 and 1.  Frequently, the
$\lambda$ for each stack is the same.

The pruning process begins with the stacks which are on the {\em
frontier}, i.e. stacks which have only states with no descendents.
All of the states in a frontier stack are marked as alive.  The score
$p_{max}$ of the highest scoring state in the stack is determined, and
only states which have a score greater than or equal to
$\lambda~p_{max}$ are advanced.  The remainder of the states in a
stack are pruned for this iteration.  Next, the parent states of all
of the unpruned states states are marked as ``alive.''  This pruning
procedure is applied recursively to a stack $S$ once all of the stacks
containing states generated by states in $S$ are pruned.
Figure~\ref{Stacks} indicates the pruning threshold with a dashed line
separating unpruned states above from pruned states below.  Once the
pruning algorithm reaches the initial stack (000) and all of the
unpruned states have been advanced, the algorithm begins again at the
new frontier.

A major difference between the stack decoding algorithm and standard
tree search algorithms is that, while tree pruning algorithms
generally eliminate states permanently, the stack decoding algorithm
only prunes states temporarily.  If a state generates unfruitful
progeny, it will not remain alive.  Once a high probability state dies
due to low probability descendents, the pruning threshold for that
state's stack is lowered, re-animating previously pruned states.

While stack decoders work best when there is no permanent pruning, it
may be necessary to implement an upper limit on the number of states
allowed in each stack, especially if the state space is very large and
memory is a limited resource.  It is unlikely, however, that pruning
the least probable state in a large stack will affect the search
results.  Consider the situation where this permanent pruning leads to
a search error.  In this case, if the lowest probability state were
not pruned, the search procedure would likely take a very long time
expanding the more probable states first.  Assuming there is some
limit on the search processing time, the search procedure would likely
run out of time before getting around to processing the lowest
probability state.

Note in Figure~\ref{Stacks} that there are multiple goal stacks.  Goal
states in SPATTER consist of a single-rooted labeled parse trees
spanning the entire sentence.  Since different parses of the same
sentence can have different numbers of constituents, goal states will
not all be assigned to the same stack.  Thus, there will be multiple
goal stacks when the SPATTER search procedure is terminated.
According to the stack decoding procedure, there is no justification
for comparing the scores of states from different goal stacks.
However, the SPATTER parsing algorithm selects the highest probability
completed parse tree, regardless of the amount of structure.  This
{\em does} favor simpler analyses over parses with more constituent
structure.  This favoritism biases the parser against adding {\em
unnecessary} structure, but it does not prevent the construction of
complex analyses if simpler ones are not likely according to the
models.

\chapter{Probabilistic Models in SPATTER\label{MODELchapter}}

The SPATTER parser assigns probability to a parse tree $T$ given the
sentence $S,$ ${\cal P}(T \given S).$  This is referred to as a {\em
parsing} model, which is distinguished from a {\em generative} model,
i.e. one which assigns probability to a parse tree and a sentence,
${\cal P}(T,S).$

The most common type of probabilistic parsing model is the P-CFG model
(e.g. Sharman, Jelinek and Mercer \cite{Jelinek:90}, Black, Garside,
and Leech \cite{blackgram}), which assigns a probability to each rule
in a context-free grammar and computes the probability of the parse
tree by assuming that each grammar rule application is independent of
all other rule applications in the sentence.  As illustrated in
Magerman and Marcus \cite{Pearl}, Black et.al. \cite{HBG}, and
Magerman and Weir\cite{Picky}, probabilistic parsers are much more
accurate when their models incorporate lexical information from the
context, and when the applications of the models are not assumed to be
independent.  The development of history-based grammar models is based
on the premise that each decision potentially affects future decisions
in the parsing process.

The statistical models in SPATTER can ask questions about all feature
value assignments made prior to the current parse action.  Due to
constraints on resources, both computational and physical, the models
are limited to asking questions about two constituent nodes to the
left of the current node, two constituent nodes to the right of the
current node, and up to four children of the current node.

\section{Notation}

In the remainder of this section, the following notational scheme is
used.  $w_i$ and $t_i$ refer to the word corresponding to the $i$th
token in the sentence and its part-of-speech tag, respectively.  $N^k$
refers to the 4-tuple of feature values at the $k$th node in the
current parse state, where the nodes are numbered from left to right.
$N_l^k,$ $N_w^k,$ $N_t^k,$ $N_c^k,$ and $N_e^k$ refer, respectively,
to the label, word, tag, conjunction, and extension feature values at
node $k.$ $N^{c_j}$ refers to the $j$th child of the current node
where the leftmost child is child 1.  $N^{c_{-j}}$ refers to the $j$th
child of the current node where the rightmost child is child 1.  The
symbol $Q_{etc}$ refers to miscellaneous questions about the current
state of the parser, such as the number of nodes in the sentence and
the number of children of a particular node.

\section{Probabilistic Models for Node Features}

The probability distribution for each feature value is estimated using
conditional models in the form of statistical decision trees.  The
decision tree models are conditioned on information from a five node
window, including the node processed and its children.  Since the
ordering of feature value assignments is not fixed, some feature value
slots in the nodes within this window will be empty, i.e. they may not
have been assigned yet.  If a node feature value being queried by the
decision tree is unassigned, the decision tree is licensed to ask
about that same feature value from the nearest child of that node.  If
the node has no children, a canonical NULL feature value is returned,
indicating that there is no information available.

\begin{figure}[tbhp]
\centerline{\psfig{figure=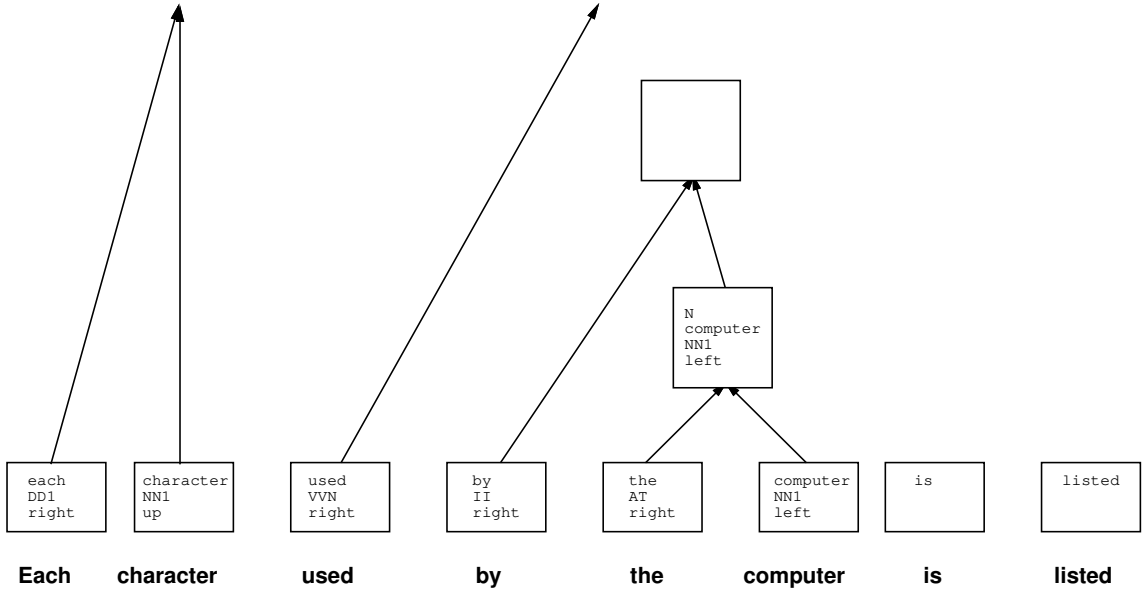,width=6in}}
\caption{Partially-constructed SPATTER parse tree.\label{partialparse}}
\end{figure}

In this section, each of the conditional decision tree models used in
SPATTER is defined.  Using the partially-constructed parse tree in
Figure~\ref{partialparse}, examples are provided of the specific
information made available to the parsing models.

\subsection{The Tagging Model}

The tag feature value prediction is conditioned on the two words to
the left, the two words to the right, and all information at two nodes
to the left and two nodes to the right.

\beqn {\cal P}(t_i \given context) \approx
{\cal P}(t_i \given w_i w_{i-1} w_{i-2} w_{i+1} w_{i+2} t_{i-1}
t_{i-2} t_{i+1} t_{i+2} N^{k-1} N^{k-2} N^{k+1} N^{k+2}) \eeqn

In Figure~\ref{partialparse}, consider the tag feature value
assignment decision for the word {\bf is.} In this case, the decision
tree model can ask questions about the word being tagged,
$w_i=\mbox{\bf is},$ the two words to the left, $w_{i-2}=\mbox{\bf
the}$ and $w_{i-1}=\mbox{\bf computer},$ and the two words to the
right, $w_{i+1}=\mbox{\bf listed}$ and $w_{i+2}=\mbox{\bf NULL}.$ It
also can ask about the tags which have been previously assigned in
this five word window, namely $t_{i-2}=\mbox{\bf AT}$ and
$t_{i-1}=\mbox{\bf NN1}.$ It can ask about any feature value
information contained in the two parse nodes to the left, in this case
the leaf node containing the word {\bf used} ($N^{k-2}$) and the node
containing the phrase {\bf by the computer} ($N^{k-1}$) for which none
of the feature values have yet been assigned.  Since none of the
feature values are known for node $N^{k-1},$ the decision tree can
instead ask questions about its rightmost child, the {\bf N} node
corresponding to the noun phrase {\bf the computer.} It can also ask
about the information in the two parse nodes to the right ($N^{k+1}$
and $N^{k+2}$), but in this case, there is only one parse node to the
right of the current node, the leaf node containing the word {\bf
listed.} Finally, the decision tree can ask miscellaneous questions
about any of the nodes in this five node window, including about the
number of children each of these nodes has, and about the presence or
absence of certain key words or punctuation marks.

\subsection{The Label Model}

The label feature value prediction is conditioned on all information
from two nodes to the left and two nodes to the right, on all
information from the two leftmost and two rightmost children of the
current node, and on miscellaneous questions about any of these nodes
or about words in the sentence.

\beqn {\cal P}(N_l^k \given context) \approx
{\cal P}(N_l^k \given N^{k-1} N^{k-2} N^{k+1} N^{k+2} N^{c_1} N^{c_2}
N^{c_{-1}} N^{c_{-2}} Q_{etc})\eeqn

In Figure~\ref{partialparse}, consider the label feature value
assignment decision for the node containing the phrase {\bf by the
computer.} In this case, the decision tree model can ask questions
about any feature value information contained in the two parse nodes
to the left and to the right, in this case the leaf nodes containing
the words {\bf character} ($N^{k-2}$), {\bf used} ($N^{k-1}$), {\bf
is} ($N^{k+1}$), and {\bf listed} ($N^{k+2}$).  It can also ask about
the two children of this node, the node containing the word {\bf by}
($N^{c_1}$ and $N^{c_{-2}}$) and the node containing the noun phrase
{\bf the computer} ($N^{c_{-1}}$ and $N^{c_2}$).  Finally, the
decision tree can ask miscellaneous questions about any of these
nodes, including about the number of children each of these nodes has,
and about the presence or absence of certain key words or punctuation
marks.

\subsection{The Extension Model}

The extension feature value prediction is conditioned on the node
information at the node being extended, all information from two nodes
to the left and two nodes to the right, and the two leftmost and two
rightmost children of the current node (these will be redundant if
there are less than 4 children at a node).

\beqn {\cal P}(N_e^k \given context) \approx
{\cal P}(N_e^k \given N_w^k N_t^k N_l^k N_c^k N^{k-1} N^{k-2} N^{k+1}
N^{k+2} N^{c_1} N^{c_2} N^{c_{-1}} N^{c_{-2}})\eeqn

\subsection{The Conjunction Model}

The conjunction feature value prediction is conditioned on the node
information at the node being extended, all information from two nodes
to the left and two nodes to the right, and the two leftmost and two
rightmost children of the current node (these will be redundant if
there are less than 4 children at a node).  However, questions about
the conjunction bit of adjacent nodes are omitted from the set of
candidate questions.

\beqn {\cal P}(N_c^k \given context) \approx
{\cal P}(N_c^k \given N_w^k N_t^k N_l^k N^{-1} N^{-2} N^1 N^2 N^{c_1}
N^{c_2} N^{c_{-1}} N^{c_{-2}})\eeqn

\subsection{The Derivation Model}

In initial experiments, the active node selection process was modeled
by a uniform (${\cal P}(active)=1/n$) model.  However, different
derivations are considered in the parser, at great computational
expense, in order to allow better derivations to supersede the
leftmost bottom-up derivation when appropriate.  When a uniform
distribution on the active node selection is assumed, all derivations,
better or worse, contribute to the probability of a parse.  In
experiments this characteristic was preventing good derivations, in
which the information available suggested the correct parse, from
distinguishing themselves from less good derivations, in which the
information available was either inconclusive or misleading.  The
solution to this problem is to model the active node selection,
conditioning the prediction on the current node information and the
node information available within the five node window.

\beqn {\cal P}(active \given context) \approx
{\cal P}(active \given Q_{etc} N^k N^{k-1} N^{k-2} N^{k+1} N^{k+2})
\eeqn

\subsection{The Parsing Model}

The overall model used in SPATTER is defined in terms of the previous
approximations.

First, the probability of a parse tree given the sentence is the sum
over all derivations of that parse tree:

\beqn {\cal P}(T \given S) = \sum_d {\cal P}(T,d \given S) \eeqn

The probability of a derivation of a parse tree is the product of the
each of the feature value assignments in that derivation and the
probability of each active node selection made in that derivation:

\beqn {\cal P}(T,d \given S) = \prod_{N \in T, j < |d|}
{\cal P}(active=N \given context(d_j)) {\cal P}(N_x \given
context(d_j)) \eeqn

where $x$ ranges over all feature values predicted at a node.


\section{Expectation Maximization Training}

The most obvious way to estimate a probability distribution from a
corpus of examples is to count the occurrences of the phenomena of
interest and to estimate the probability of an event to be the
frequency of the event divided by the total number of events.
However, this is not a very good way to train a model which considers
as much contextual information as the models in SPATTER.

One reason why this is true is that there is a strong tendency to
overtrain models when using empirical estimates directly.  The SPATTER
parsing models consider so much contextual information that almost
every new sequence of words would introduce new parameters into the
model, and the models would be doing no more than memorizing the parse
for each sentence.  A model trained in this way would perform very
poorly on new sentences.  The overtraining problem is treated by
smoothing the decision trees with a separate {\em held-out} training
set using an expectation-maximization (E-M) algorithm.

However, smoothing issues aside, the SPATTER parsing model certainly
cannot be estimated by direct frequency counts because the model
contains a {\em hidden} component: the derivation model.  The order in
which the treebank parse trees were constructed is not encoded in the
treebank, but the SPATTER parser assigns probabilities to specific
derivations of a parse tree.  The training process must discover which
derivations assign higher probability to the treebank parses, and
favor those derivations over others.  The decision tree which assigns
a distribution to the possible {\em active} nodes directly assigns
weights to different derivations.  The reestimation procedure which
tries to maximize the probability of the parse trees of a corpus is
called the {\em forward-backward} algorithm.

\subsection{Forward-Backward Reestimation}

The {\em forward-backward} (F-B) algorithm, which is a special case of
the expectation-maximization algorithm, reestimates the parameters of
a model $P_i$ to generate a new model $P_{i+1}$ which is guaranteed to
assign a higher probability to the training corpus than the original
model, $P_{i+1}(C) \geq P_i(C),$ when certain assumptions about the
model hold.  For proofs concerning this and other properties of the
forward-backward algorithm, see Poritz\cite{forbac}.

The intuition behind the F-B algorithm is that when there are multiple
paths from an initial state to a goal state, the different paths
should contribute to the model according to their relative
probabilities.  For example, let's say there are two different
derivations, $d_1$ and $d_2,$ of a parse tree of a sentence, $T,$
where $P(T)=P(d_2,T)+P(d_1,T).$ If the goal is to maximize the
probability of the parse tree $P(T),$ then it would make sense to try
to increase the probabilities assigned to the actions in $d_1$ and
$d_2.$ However, the reestimation procedure is complicated by the
constraint that the probabilities of all actions at a given state must
sum to 1.  It may be impossible to increase the probabilities of all
of the actions in both derivations.  Consider the first step in each
derivation.  If the probability of the first step of $d_1$ is set to
1, then the probability of the first step of $d_2$ must be zero, since
both derivations begin with the same initial state.  The F-B algorithm
manages the constraints of probabilistic models while improving the
probability of the corpus from one iteration to the next.

\subsection{Forward-Backward in SPATTER\label{FB}}

To train the SPATTER parser, all legal derivations of a parse tree
(according to the derivational window constraint) are computed.  Each
derivation can be viewed as a path from a common initial state, the
words in the sentence, to a common final state, the completed parse
tree.  These derivations form a lattice of states, since different
derivations of the same parse tree inevitably merge.  For instance,
the state created by tagging the first word in the sentence and then
the second is the same state created by tagging the second word and
then the first.  These two derivations of this state have different
probability estimates, but the state can be viewed as one state for
future actions, since it represents a single history.

Adjacent states in this lattice differ by only a single feature value
assignment.  For a state $s,$ let $s_h$ represent a state which
precedes $s$ in the state lattice, and let $f(s_h,s)$ be the feature
value assignment which was made to get from state $s_h$ to state $s.$
The probability of a state $s$ is computed by
\beqn P(s) = \sum_{s_h} P(s_h)P(f(s_h,s)|s_h).\eeqn

The construction of the state lattice and assignment of transition
probabilities according to the current model is called the forward
pass.  The probability of a given state, $P(s),$ is referred to as
$\alpha(s).$ The backward probability of a state, referred to as
$\beta(s),$ is calculated according to the following recursive
formula:
\beqn \beta(s_h) = \sum_s \beta(s) P(f(s_h,s)|s_h) \eeqn
where the backward probability of the goal state is set equal to the
forward probability of the goal state, $\beta(s_{goal})=\alpha(s_{goal}).$
The count associated with a feature value assignment, $f(s_h,s),$ is
\beqn count(f(s_h,s)) = \frac{\beta(s) \alpha(s_h)
P(f(s_h,s)|s_h)}{\alpha(s_{goal})}. \eeqn
This value, $count(f(s_h,s)),$ is the contribution of the event
$f(s_h,s)$ to the distribution which predicts the feature value $f$
given the history $h.$  The reestimate of the probability of a feature
value assignment given a history is the ratio of the total count for
that feature value assignment and the total count of all feature value
assignments given that history:
\beqn P_{new}(f|h) \approx \frac{count(f(s_h,s))}{\sum_{s'} count(f(s_h,s'))}.
\eeqn

\subsection{Decision Trees and the Forward-Backward Algorithm}

Each leaf of decision tree represents the distribution of a class of
histories.  The parameters of these distributions can be updated using
the F-B algorithm as described in chapter \ref{FB}.

In addition to increasing the probability of a corpus, the algorithm
assigns weights to the events in a corpus, where the weights represent
the relative contribution of the event to the probability of the
corpus.  This attribute of the F-B algorithm is useful for growing the
decision tree models used in SPATTER.

Initially, the models in SPATTER are assumed to be uniform.
Accordingly, each event in each derivation contributes equally to the
process which selects questions to ask about the history in order to
predict each feature value.  However, the uniform model is certainly
not a very good model of feature value assignments.  Additionally,
since some derivations of a parse tree are better than others, the
events generated by the better derivations should contribute more to
the decision tree-growing process.  The decision trees grown using the
uniform assumption collectively form a parsing model, $M_1.$ The F-B
count for each event in the training corpus using $M_1$ can be used to
grow a new set of decision trees, $M_2.$ The decision trees in $M_2$
are constructed in a way which gives more weight to the events which
contributed most to the probability of the corpus.  However, there is
no guarantee that $M_2$ is a better model than $M_1.$ It isn't even
guaranteed that the probability of the training corpus according to
$M_2$ is higher than the probability according to $M_1.$ However,
based on experimental results, the use of F-B counts in the
construction of new decision trees is effective in acquiring a better
model of the data.

\subsection{Training Algorithm\label{SPATTERtraining}}

There is no way of knowing {\em a priori} which combination of the
previously mentioned applications of the forward-backward algorithm
will produce the best model.  It might be best to grow an initial set
of decision trees, perform F-B training on these trees for a few
iterations, and then grow new trees.  It might be best to grow more
than two sets of trees, generating model $M_3$ from the F-B counts of
$M_2,$ and $M_4$ from the F-B counts of $M_3.$

Growing the decision tree models takes almost 2 days on one machine,
and so the number of variations which can be attempted is limited.
After initial experimentation, the following sequence of training
steps proved effective:

\begin{enumerate}
\item Grow initial decision trees ($M_1$) based on uniform models.
\item Create $M_2$ by pruning trees in $M_1$ to a maximum depth of 10.
\item Grow decision trees ($M_3$) from F-B counts from $M_2.$
\item Perform F-B reestimation for leaves of decision trees in $M_3.$
\end{enumerate}

There is no obvious justification for the decision to prune the trees
in step two to a depth of 10, as opposed to a depth of 5 or a depth of
20.  If computational resources permit it, the F-B reestimation may
also be applied to $M_1,$ with the resulting model compared to the
fully trained $M_3$ model.  Also, steps two and three may be applied
to $M_3$ to generate a new model $M_5,$ which may then be trained
using F-B.

\chapter{Evaluation Methodology\label{EVALUATIONchapter}}

Natural language parsing is a means to an end.  As such, the output of
a natural language parser is difficult to evaluate outside of the
context of a natural language processing system.  Natural langauge
researchers find themselves in a Catch-22: to solve the NL problem,
they need a good parser, but in order to identify a good parser, they
need an NL system.

\section{Parsing Performance Measures}

One of the difficulties of evaluating parsers is that different
approaches to the parsing problem have different strengths and
weaknesses which can not be represented completely in a single scalar
value. Some parsers identify more information than is encoded in a
treebank, such as word senses, the semantic roles of constituents, or
the referents of pronouns.  While this information may be useful if
not crucial to an NL parsing system, it is difficult to evaluate the
accuracy of these features, much less the impact of these capabilities
on the parsing task.

The treebank recognition problem was defined in
Chapter~\ref{SPATTERchapter} to limit the scope of the parsing problem
in order to more easily evaluate alternative solutions.  However,
proponents of other parsing methods argue that their approaches don't
lend themselves to the strict evaluation criterion of the TRP, exact
match.  I argue this point later in this section.

Here I consider the relative merits of a number of evaluation
criteria.  First, I comment on the use of test entropy as a measure of
performance in a statistical parser.  Then I discuss the various
applications of the {\em crossing-brackets} measure, along with the
constituent-based measures of {\em precision} and {\em recall.}
Finally, I consider the strictest measure of accuracy, the exact-match
criterion.

\subsection{Entropy as a Predictor of Performance}

Entropy provides the second best measure of the expected relative
performance of different models, second only to applying the model to
a real recognition task.  In speech recognition, it would be ideal to
evaluate every experimental language model by plugging it into a
speech recognizer and judging the improvements made in the recognition
accuracy.  However, this is infeasible.  Instead, language models
distinguish themselves by assigning a higher probability to a given
test corpus than existing models.  Only if a model achieves a
significant reduction in entropy over the state-of-the-art will it be
tried out in a recognizer.

Statistical training algorithms use entropy as their objective
function, trying to improve a model in a way which increases the
expected likelihood of a training corpus.  Expectation-Maximization
algorithms are frequently used in training precisely because they {\em
guarantee} improvements in training entropy.  The hypothesis upon
which these procedures depend is that improvements in training entropy
will lead to improvements in test entropy.

While this hypothesis is frequently true, it is not guaranteed.
Overtraining can lead to the construction of a model which essentially
memorizes the training data, resulting in a training entropy close to
0.  However, an overtrained model will yield high entropy on a test
corpus, especially if the training corpus is small compared to the
size of the domain.

Test entropy is not a perfect measure of performance.  It is possible
to decrease the entropy of a data set without increasing the
recognition accuracy rate of the model on that data.  Consider two
models which assign probability to a sequence of independent binary
random variables.  For each event, both models select the choice which
has a probability greater than 0.5.  The first model is a perfect
model which assigns probability 0.51 to the correct choice every time.
The second model ``improves'' this model by assigning probability 1 to
the correct choice for half of the events and probability 0.49 to the
correct choice for the other half.  The second model has a lower test
entropy than the first, since it has 0 entropy for half of the events
and only a negligibly smaller entropy for the other half.  However,
the first model has a 100\% recognition rate, whereas the second
model's recognition rate is only 50\%.  This is an extreme example,
but training algorithms will frequently sharpen the probability of
high probability events, increasing the probability of an event from
0.9 to 0.999, resulting in a decrease in entropy which is unlikely to
change recognition performance at all.

Test entropy has other weaknesses as well.  Since a test corpus
generally contains only a small sample of events from the domain, a
small decrease in test entropy may not be significant. Also, test data
is frequently generated in a way makes it artificially similar to the
training data.  Thus, an overtrained model will report lower entropy
on the test corpus than it would achieve in a true test of the model
on a recognition task.

Despite these problems, test entropy is useful for gauging incremental
progress on a statistical modeling problem.  Significant test entropy
reduction from a new model typically leads to recognition
improvements.  But it is important to verify these improvements by
decoding with the new model.

\subsection{The Misguided Crossing-Brackets Measure}

The crossing-brackets measure was introduced during the PARSEVAL
workshop on parser evaluation at the University of Pennsylvania in
1990 (Black et.al.\cite{CBpaper}).  This measure was never intended to
be used in isolation; it was one of three measures which together
evaluate the performance of a parser.  The workshop participants
reached a consensus that the constituent-based measures of
crossing-brackets rate, precision, and recall, sufficiently
represented the performance of their parsers in comparison to a
skeletal treebank.

A single crossing-bracket error is a constituent in a parse tree which
contains parts of two different constituents from a treebank analysis
without completely containing either.  For instance, consider the
following treebank analysis and corresponding parser analysis:
\begin{verbatim}
Treebank: [ [ A B ] [ C D ] [ E F ] G ]
Parse Tree: [ [ A B C ] D [ E F G ]
\end{verbatim}
The parse constituent ``[~A~B~C~]'' is an instance of
crossing-brackets violation, since B and C are not in the same
constituent in the treebank analysis, and neither of the treebank
constituents containing B or C are completely contained in the parse
constituent.  On the other hand, ``[~E~F~G~]'' is not a
crossing-brackets violation, even though it is not represented in the
treebank analysis.

The three constituent-based measures are calculated by the following
formulas:
\begin{description}
\item[Crossing-Brackets] --- total number of crossing-brackets
violations in the sentence,
\item[Precision] --- $\frac{\mbox{\# of parse constituents which
exactly match treebank constituents}}{\mbox{total \# of parse
constituents}},$
\item[Recall] --- $\frac{\mbox{\# of treebank constituents which
exactly match parse constituents}}{\mbox{total \# of treebank
constituents}}.$
\end{description}

Without {\em precision} and {\em recall,} the constituent-based
crossing-brackets measure is a very weak measure of parsing accuracy.
First, a trivial analysis with no constituent structure yields a
perfect score on this measure.  And assigning extra structure which
does not violate constituent boundaries results in an artificially
higher crossing-brackets score.  But even if a parser assigns the same
number of constituents as the treebank, or more, the crossing-brackets
measure does not identify classes of errors including incorrect
prepositional-phrase attachment, inappropriate internal noun-phrase
structure, and omitting important constituents such as the main verb
phrase.

Another criticism of the workshop evaluation measures is that they
have completely side-stepped the issue of constituent labeling and
part-of-speech tagging.  It undoubtedly would have been difficult to
reach a consensus about a constituent label set or part-of-speech tag
set at the workshop, considering the difference between the level of
detail supplied by some grammars compared to a skeletal treebank.
However, by ignoring tagging and labeling errors, inaccurate parses
which misidentify the main verb and its arguments might still get low
crossing-bracket error scores.  And a natural language processing
system will not be able to use a parser effectively unless it is
identifying the categories of constituents along with the constituent
structure.

The PARSEVAL measures provide very crude evaluations of parsing
performance.  It was the estimation of the PARSEVAL workshop
participants that the state-of-the-art in parsing was at such a low
level that these crude measures were sufficient to differentiate good
parsers from bad.  While that was probably true 3 years ago, the level
of parsing performance, particularly for those using statistical
modeling techniques, has improved to the point that these measures
are no longer informative in isolation.  They may provide an upper
bound on performance, since a parse cannot be correct if it has
crossing-brackets errors, but a high score on these measures does not
necessarily indicate a solution to the parsing problem.

\subsection{The Exact Match Criterion}

The exact match criterion is a much stricter evaluation of a parser's
performance than the PARSEVAL statistics.  This measure considers a
parse tree correct if and only if every constituent, label, and tag in
the parse tree matches those in the treebank analysis.  This, of
course, requires that the parser generate parses using the same
constituent label and part-of-speech tag sets that were used to
annotate the corpus.

In some ways, the exact match criterion appears {\em too} stringent.
The measure doesn't take into account the fact that treebanks are
internally inconsistent.  The internal consistency of the Lancaster
treebank, used as training data for the experiments reported in this
chapter, has been measured at a little higher than 50\%.  This means
that if the same sentence is analyzed twice by treebankers, there is
only a 50\% chance that both analyses will be identical.  The internal
consistency of the UPenn treebank has been measured at 23\%.

These consistency measures certainly appear to be damning for the
exact match criterion, but they really are not.  It is most important
that the test data be internally consistent; training algorithms
should be robust enough to tolerate some inconsistency in the training
data.  Since test sets are considerably smaller than training sets, it
is feasible to have multiple treebankers annotate the test sentences,
and select the consensus analysis as the correct one.  Using this
method, the Lancaster treebankers achieved over 90\% consistency on
test data.  Certainly, it would improve matters to have a higher level
of quality control in the treebanking process.  But perfect
treebanking should not be expected, nor could it be achieved.  Humans
can not be expected to be completely consistent when there are perhaps
dozens of treebankers analyzing hundreds of thousands of sentences
over the course of a weeks, months, or years.  However, as long as
test data has a consistency rate much higher than the accuracy rate of
state-of-the-art parsers, then exact match measures can be useful for
evaluating parsing performance.

Aside from the internal consistency of a treebank, there is still a
strong argument against the exact match criterion.  Adding or omitting
constituent structure in a way which does not significantly alter the
meaning of the analysis should not be considered as much of an error
as completely misanalyzing a sentence.  However, the exact match
criterion does not differentiate between these two types of errors.

There are many different levels at which the structure of a sentence
can be analyzed.  The Lancaster treebank attempts to indicate some
internal noun-phrase structure inconsistently, and it leaves other
detailed structure out completely.  The UPenn treebank omits all
internal noun-phrase structure, including the internal structure of
multiple conjoined noun phrases.  However, statistical learning
algorithms, including the decision tree methods discussed here, can
overcome the inconsistencies in the treebank and accurately predict
internal noun-phrase structure as well as other details which are not
annotated in the treebank.  They also can make blatant mistakes by
adding extra structure which completely change the meaning of the
analysis.

\subsection{An Argument for the Exact Match Criterion}

Most of the drawbacks of the exact match criterion are actually flaws
in the treebanking process, not in the evaluation measure itself.  No
treebank will ever be perfect; there is an upper bound on the
consistency and accuracy with which humans can annotate data.
However, as discussed above, the consistency of test data can be
brought up to acceptable levels.

The exact match criterion measures something concrete: the parser's
ability to make generalizations about the syntactic structures in the
training data and apply this knowledge to new data consistently.  If a
parser can do this, by whatever means, it can achieve a high score on
this measure.  This may require renaming constituent labels and
part-of-speech tags in a grammar to match those of the treebank.  But,
after all, these are only symbols to the parser.  They could be
renamed A1 through A100 and still represent the same information.

The alternative to using exact match is to continue to accept human
judgment of parsing performance as a scientific measure.  A grammarian
or one of his colleagues scanning a few hundred parses and reporting
the percentage which the evaluator {\em feels} are correct is not a
very reliable measure.  The human evaluator is performing a task even
more difficult than the treebanker's task, and we have already seen
measures of the reliability of individual annotators.  But, even if
the evaluator is internally consistent in his judgments, different
evaluators are used in every experiment, and these evaluators'
standards and judgments are not available for review.  These standards
and judgments {\em are} available for the exact match criterion, in a
treebanker's style manual, and to a certain extent in the treebank
itself.

Restating the parsing problem as treebank recognition does not really
change the problem; it merely levels the playing field.  The only
difference between the traditional parsing problem and treebank
recognition is that the minimal label and tag sets are predetermined.
A grammar can assign more precise labels and indicate semantic, but it
will only be evaluated on what is contained in the treebank.
Only when parsers are speaking the same language, generating parse
trees using the same labels at some level, will we be able to compare
the performance of different parsing systems.

\section{Rules of Experimentation}
Not only do statistics represent a possible solution to natural language
processing problems, but they also offer insight into experimentation
procedures.  There are a number of rules which must be observed in
order for experiments to have statistical validity.

\subsection{Test Data}

Violations of test data etiquette are prevalent in the natural
language community.  The single most important rule of test data is
that {\em under no circumstance whatsoever} should one ever, ever,
{\em ever} look at one's test data.  This precludes not only
physically eyeing the data, but also extracting vocabularies from it,
collecting distributional information from its annotations, modifying
grammar rules based on it, or gaining {\em any information at all}
from the test data that would not be available in a real test of the
system.  Any violations of these rules put into question any results
from experiments on the test data.

Another violation of test data etiquette is more subtle: separating
training and test data from a corpus by random sampling.  It goes
without saying that one should not test on data which was used in
training.  Recently, a number of parsing papers, including two of my
own \cite{Pearl} \cite{Picky}, have reported results using test sets
were randomly sampled from a corpus, using the remainder of the corpus
as training material.  While this technique seems benign, it actually
ensures that the test data will be as statistically similar to the
training data as possible.  This improves the perceived test
performance of any training algorithm which is susceptible to
overtraining, as most statistical methods are.

Unless a program can automatically adapt its models in real time based
on recent test data to improve its performance on new test data, it is
also inappropriate to use test data which is known to be contiguous
with the training data.  For example, the experiments described in
this chapter are based on a large training set and a non-contiguous
test set.  When the same experiments were repeated by training on the
first 90\% of the training data and testing on the final 10\%, the
accuracy results were significantly higher.  While, according to the
standard definition this new experiment is a fair test, it is clear
that the results are biased by the proximity of the test data to the
training data.  And since the training algorithms used in this work
are not incremental, the better results are not reflective of the
actual performance of the parser.

Finally, it is important not to test too many times on the same test
data.  It is possible to fit solutions to a particular test set, even
if the only result from an experiment is a single accuracy rate.  This
is a danger particularly when different experiments involve changes to
a rule base.  If a grammarian keeps changing a grammar's rule base
until the grammar performs better on a particular test set, it is not
necessarily true that the rules changed in the ``improved'' grammar
are any more of an improvement than the rule changes made in previous
experiments.  It only means that the changed rules which led to the
increased accuracy rate are rules which are used in a correct parse in
that particular test set.

\subsection{Training Data}
Training data is the critical source of information which will drive
any acquisition process, be it automatic or manual.  It is important
that this data set be an unbiased random sample of the typical events
in the domain being modeled.

\subsubsection{Data Collection}
An example of biased data collection comes from the ARPA ATIS project
\cite{MADCOW:92}.  This project involves building a spoken language
interface to an airline reservation system.  Data was collected at a
number of ARPA research sites using a standard Wizard of Oz setup,
where speakers interact with a what they believe is a computer program
but which is actually a human being responding electronically.  At one
of the sites, the participants were given an instruction sheet with
information about the tasks they were to perform, {\em along with
example queries.} When the data was accumulated from all of the sites,
it became clear that the aforementioned site had collected biased
data.  Unlike the data collected at other sites, which contained a
great deal of variability between speakers, this site's data was
largely uniform.  Nearly all of the speakers had asked about the few
things mentioned in the example queries, and had even asked about them
using the same sentence structure.

Another area in which data collection methodology is critical is in
treebanking.  Treebanking and corpus annotation is a subtle task which
can yield strikingly different results depending on the method used.
The Lancaster and UPenn treebanks are supposed to contain generally
the same information, skeletal syntactic analyses.  However, they were
constructed using different philosophies, and the annotators for the
treebanks did their work using very different software environments.

The Lancaster researchers stressed consistency over efficiency.  The
Lancaster treebankers started with completely unannotated sentences
and filled in all of the structure using a simple editor.  Working
with the IBM parsing researchers, they iteratively improved the
consistency of the data, reannotating data sets over and over until
the data was of sufficient quality to train and evaluate a statistical
parser.

The UPenn researchers put a premium on getting data to the research
community as soon as possible.  They used automatic parsing and
tagging programs as preprocessors and had annotators correct the
automatically annotated data.  There was also a high turnover rate in
the early years of the treebanking project.  These factors resulted in
a treebank which had low internal consistency.  The early versions of
the treebank released to the research community included segments of
data done by very poor annotators, data which has since been
reannotated because of its low quality.  The data also contained
unmatched brackets; and since sentence boundaries were not clearly
indicated, there was no systematic way to determine where the missing
brackets should be.  There were also sentences which included markers
and symbols output by the automatic parsing preprocessor.  Based on
the number and types of errors, these sentences clearly had not been
corrected by treebankers.

\subsubsection{Using Data Representative of the Problem}

Recently, there have been a number of papers, such as Schabes and
Pereira\cite{Schabes:92}, Brill \cite{Brill}, and Bod\cite{Bod},
citing work based on parsing using sentences tagged for part of
speech.  While it is true that part-of-speech taggers have very low
error rates, some below 3\%, parsing from {\em manually assigned} tags
is not the same as parsing from {\em automatically assigned} tags.
Automatic taggers make systematic errors that might seriously affect
the performance of a parser trained on only correct tag assignments.
Since a parser will be tested on data which is completely
automatically processed (a human tagger will not be a component of an
NL system), the parser should be trained on automatically tagged data.
State-of-the-art automatic tagging programs are available throughout
the research community, so accessibility to automatically tagged data
should not be an issue.  Certainly, training and testing on correct
tag sequences will lead to higher accuracy results.  But such
experiments are artificial and do not reflect the performance of the
parser on a real test.

\subsubsection{Hand-modifying Training Data}

In order to ensure reproduceability of experiments, one should not to
hand-modify training data.  This issue was discussed with regards to
test data, but it is nearly as important with regards to training
data.  Training data should be generated in a way which is
documentable and reproduceable.  If humans are generating data, it
must be clear how to generate similar data.  Otherwise, the results
cannot be reproduced by other members of community.

Altering training data can also cause confusion.  As was mentioned
earlier, the UPenn treebank has a very low internal consistency rate.
In his recent paper \cite{Bod}, Bod reports achieving a 96\% exact
match accuracy rate parsing UPenn treebank data from the ATIS domain.
Yet, a statistical analysis of this data reveals that, due to the
internal inconsistency of the training data, there was an upper bound
of 70\% accuracy on a fair test using that data.\footnote{These
consistency experiments were performed by Salim Roukos, Adwait
Ratnaparkhi, and Todd Ward at IBM.} Bod later revealed that he had in
fact hand-corrected his training (and test) data.  Thus, all of the
experiments we had planned to reproduce his results and to compare
them directly to other parsers on the same data were useless.

\section{Conclusions}

The bottom line in experiment methodology is that a system should be
trained in an environment that is comparable to the environment in
which a working version of the system would be trained, and it should
be tested in a way which is {\em identical} to the way the system will
be used.

\chapter{Experiment Results\label{RESULTchapter}}

In the absence of an NL system, I have performed parsing experiments
which attempt to evaluate the output of the parser directly.  These
experiments explore the parser's performance and how this performance
is affected by variations in the training process and in the
configuration of the parser.

The parser output in these experiments are evaluated by the following
criteria: exact match of structure, labels, and tags ; exact match of
structure and labels; exact match in the top 5 ranked parses; and
exact match in the top 20 ranked parses.  Training and test entropies
for all of these experiments are listed in
Appendix~\ref{ENTROPYappendix}.

The domain of these experiments, IBM Computer Manuals domain, is
described in section~\ref{TREEBANKsection}.

\section{Parser Configuration}

The SPATTER parser uses the tag and label vocabularies dictated by the
Lancaster treebank annotations.  The word vocabulary is also
determined by the Lancaster training data.\footnote{Words are selected
from the training data by frequency so that approximately 5\% of the
words in the data will be outside of this fixed vocabulary.  This is
to allow the training algorithms to acquire a model for unknown
words.} SPATTER also uses vocabularies which contain the possible
answers to the questions which the decision trees can ask.  Each of
the items in these vocabularies has a corresponding binary
representation.  These vocabularies and their binary encodings are
listed in Appendix~\ref{VOCABULARYappendix}.

The Tree Head Table used in these experiments is included in
Appendix~\ref{THTappendix}.

The parser also uses a dictionary to constrain the set of
part-of-speech tags which a word can be assigned.  This dictionary is
automatically generated by listing all of the tags which a word is
assigned in the training data.

\section{Significance of Results}

The statistical significance of the differences between the
performances of two models X and Y can be determined by counting how
many sentences X gets right and Y gets wrong $(c_{12})$ and how many
sentences Y gets right and X gets wrong $(c_{21})$.  If the hypothesis
is that X and Y have the same accuracy rate, then $c_{12}$ and
$c_{21}$ should be equal.  Further, $c_{12}$ and $c_{21}$ should be
distributed according to a binomial distribution with $p=\frac{1}{2}.$
Thus, the probability that the two models are equivalent is the same
as the probability of getting $c_{12}$ heads when tossing a fair coin
$c_{12}+c_{21}$ times.  When $c_{12}+c_{21}$ is large, the binomial
distribution can be estimated using the DeMoivre-Laplace
approximation, which is much easier to compute than the binomial.

\section{Basic Experiment}

The idea behind the basic experiment is to construct the best parser
possible and to perform experiments using the same training and test
data as the experiment reported in Black, Garside, and
Leech\cite{blackbook}, which I refer to as the P-CFG experiment.  The
basic experiment consists of constructing SPATTER's decision tree
models as described in Chapter~\ref{MODELchapter} using the treebank
data, and training these models nearly to convergence.\footnote{I
don't allow the training algorithms to converge largely because of the
computational constraints of current hardware.  However, further
training is more likely to result in overtraining than in performance
improvements.}

The training procedure for the basic experiment is described in
section~\ref{SPATTERtraining}.  The first set of decision trees were
grown, pruned to a depth of 0, and new trees were grown based on the
pruned trees.  This is effectively the same as growing the second set
of trees based on a unigram model extracted from the training data.
Since the active node decision tree model is self-organized, it is
pruned to a depth of 20 before smoothing, since the nodes nearer to
the leaves are likely to be overtrained on the training sentences.
Nine iterations of the forward-backward algorithm were applied to the
second set of trees, and the final model was smoothed for 20
iterations using held out data.

Not all of the sentences in the Lancaster treebank training set were
used in the training and smoothing processes.  Only sentences between
3 and 30 words in length were considered.  Also, any sentence which
contained a constituent with more than 8 children was deemed too
superficially annotated to be trained on.  When applying the models,
the parser is also constrained not to generate a constituent with more
than 8 children.\footnote{Treebankers are instructed to skip sentences
which are incomplete or make no sense to them.  Sentences with
constituents made of more than 8 children likely fall into this
category.}

The parser was trained on the first 28,000 sentences of the Lancaster
treebank training set and smoothed using the next 2,800 sentences.
The next 100 sentences were used to generate experimental test
entropies during the smoothing process.

The test set included 1,473 sentences, whose lengths range from 3 to
30 words, with a mean length of 13.7 words.  These sentences are the
same test sentences used in the experiments reported for the P-CFG
parser in Black, Garside, and Leech\cite{blackbook}.  Since
\cite{blackbook} only reports results using the sentence-based
crossing-brackets measure, I am reporting the same measure for the
sake of comparison.  On sentences of 25 words or less, SPATTER Model
1.9 has a 78\% accuracy rate, as compared to the 69\% accuracy rate of
the P-CFG.

\subsection{Interpreting the Results}

Table~\ref{RESULTStable} contains the results of the basic experiment.
It includes the results obtained by performing 0 and 9 iterations of
the F-B algorithm on both sets of trees.

\begin{table}[tbhp]
\begin{center}
\begin{tabular}{|l|c|c|c|c|r|}
\hline
Experiment & EXACT & EXNOTAG & EXTOP5 & EXTOP20 & Perp.\\
\hline
Model 1.0 & 37.5\% & 45.6\% & 50.2\% & 59.4\% & 1270\\
Model 1.9 & 37.7\% & 45.2\% & 49.7\% & 59.8\% & 1425\\
Model 2.0 & 36.6\% & 44.2\% & 50.8\% & 60.7\% & 1049\\
Model 2.9 & 37.2\% & 44.7\% & 50.4\% & 60.5\% & 1241\\
\hline
\end{tabular}
\end{center}
\caption[Battery of results from the basic experiment.]{Battery of
results from basic experiment.  The designation ``Model x.y''
corresponds to the results from decision tree set $x$ trained using
$y$ iterations of the F-B algorithm.\label{RESULTStable}}
\end{table}

The columns correspond to the percentage of sentences in the test data
for which the SPATTER output satisfies the following criteria: exact
match of structure, labels, and tags (EXACT); exact match of structure
and labels (EXNOTAG); exact match in the top 5 ranked parses (EXTOP5);
and exact match in the top 20 ranked parses (EXTOP20).  The last
column reports the test perplexity of the grammar, which represents
the average number of parses for each test sentence according to the
model.

\begin{table}[tbhp]
\begin{center}
\begin{tabular}{|c|c|c|c|c|}
\hline
Model $X$ & Model $Y$ &
$\#\frac{X=~T}{Y\not=~T}$ &
$\#\frac{Y=~T}{X\not=~T}$ &
$p(X=Y)$\\
\hline
1.0 & 1.9 & 11 & 16 & 0.22 \\
1.0 & 2.0 & 80 & 75 & 0.37 \\
1.0 & 2.9 & 78 & 80 & 0.47 \\
1.9 & 2.0 & 81 & 71 & 0.23 \\
1.9 & 2.9 & 80 & 77 & 0.44 \\
2.0 & 2.9 & 23 & 30 & 0.21 \\
\hline
\end{tabular}
\end{center}
\caption[Significance analysis of EXACT results from the basic
experiment.]{Significance analysis of EXACT results from the basic
experiment.  Column 3 indicates the number of sentences for which
model $X$ achieved an exact match and model $Y$ did not.  Column 4
indicates the number of sentences for which model $Y$ achieved an
exact match and model $X$ did not.\label{SIGtable}}
\end{table}

\begin{table}[tbhp]
\begin{center}
\begin{tabular}{|c|c|c|c|c|}
\hline
Model $X$ & Model $Y$ &
$\#\frac{X\simeq~T}{Y\not\simeq~T}$ &
$\#\frac{Y\simeq~T}{X\not\simeq~T}$ &
$p(X\simeq~Y)$\\
\hline
1.0 & 1.9 & 15 & 13 & 0.42\\
1.0 & 2.0 & 97 & 87 & 0.25\\
1.0 & 2.9 & 97 & 93 & 0.41\\
1.9 & 2.0 & 94 & 86 & 0.30\\
1.9 & 2.9 & 99 & 97 & 0.47\\
2.0 & 2.9 & 30 & 36 & 0.27\\
\hline
\end{tabular}
\end{center}
\caption[Significance analysis of EXNOTAG results from the basic
experiment.]{Significance analysis of EXNOTAG results from the basic
experiment.  Column 3 indicates the number of sentences for which
model $X$ achieved an exact match except for tags and model $Y$ did
not.  Column 4 indicates the number of sentences for which model $Y$
achieved an exact match except for tags and model $X$ did not.
\label{SIGENTtable}}
\end{table}

Table~\ref{SIGtable} shows the significance analysis of the EXACT
results.  The first two columns show the models being compared.
Column three contains the number of sentences for which model $X$
achieved an exact match with the treebank but model $Y$ did not.
Column four contains the number of sentences for which model $Y$
achieved an exact match bin model $X$ did not.  As discussed earlier,
these counts should be distributed according to a binomial
distribution with $p=\frac{1}{2}$ and $N$ equal to the sum of columns
three and four.  Column five indicates the probability that the two
models have the same accuracy rate according to the experiment.  Based
on Table~\ref{SIGtable}, none of the experiments reveals a significant
difference between the models.  Table~\ref{SIGENTtable} shows the same
significance analysis for the EXNOTAG results.

Based on both of these tables, any two models have a non-negligible
probability of having the same accuracy rate.  Thus, none of the
differences in the performances of the models are statistically
significant.

However, there is a significant trend in test perplexity.  The Model 2
trees have a lower test entropy than the Model 1 trees.  The
perplexity steadily increases as the trees are trained more.  This
increase is evidence of overtraining.  Note that the performance
appears to degrade as perplexity decreases, the exact opposite of what
one might expect.  Of course no conclusions can be drawn, since the
performance differences are not significant enough.

\section{Variations on the Theme\label{Variations}}
I also performed a number of experiments to explore the different
parameter settings and design decisions in the basic configuration of
SPATTER.  Table~\ref{VARIATIONStable} describes the conditions of each of
the experiments.

\begin{table}[tbhp]
\begin{description}
\item[A] Parse with no derivational model, using only the bottom-up
leftmost derivation.
\item[B] Parse with no conjunction feature.
\item[C] Use a stopping rule, pruning decision trees to 1 bit-event
of significance.
\item[D] Prune decision trees to 5 bit-events of significance.
\item[E] Train using only half of the training data (14,000
sentences).
\item[F] Parse assuming the correct tag for each word is known.
\item[G] Parse from tags, assuming the correct tag sequence for the
sentence is known but the words are not.
\item[H] Use a flexible tagging dictionary, allowing the highest
probability tag each time a tag is assigned, regardless of whether the
tagging dictionary allows it.
\item[I] Use a flexible tagging dictionary, allowing the 5 highest
probability tags.
\end{description}
\caption{Descriptions of experiments A-H.\label{VARIATIONStable}}
\end{table}

Due to computational constraints, the decision trees in experiments B,
E, F, G, H, and I were not trained.  Thus it makes sense to compare
the results of these experiments to the results of the Model 1.0
experiment.  Experiments C and D were trained for 9 iterations.  Since
there is no hidden component to experiment A's models, it makes no
sense to train them.  Thus experiments A, C, and D should be compared
to the most completely trained models, Model 1.9 and Model 2.9.

\subsection{Experiment A: No Derivational Model}

\begin{table}[tbhp]
\begin{center}
\begin{tabular}{|l|r|r|r|r|r|}
\hline
Experiment & EXACT & EXNOTAG & EXTOP5 & EXTOP20 & Perp.\\
\hline
A         & 35.8\% & 43.0\% & 52.8\% & 64.2\% & 536\\
\hline
\end{tabular}
\end{center}
\caption[Results of experiment A.]{Results of experiment A (no derivation
model).\label{Atable}}
\end{table}

\begin{table}[tbhp]
\begin{center}
\begin{tabular}{|c|c|c|c|}
\hline
Model $X$ & Model $Y$ & $p(X=Y)$ & $p(X\simeq~Y)$\\
\hline
A & 1.9 & 0.03 & 0.06\\
A & 2.9 & 0.03 & 0.07\\
\hline
\end{tabular}
\end{center}
\caption[Significance analysis of results from experiment
A]{Significance analysis of EXACT and EXNOTAG results from experiment
A.\label{SIGAtable}}
\end{table}

Experiment A illustrates the value of the derivational model.  Based
on Table~\ref{SIGAtable}, there is statistically significant
improvement in performance with the derivational model.  I suspect the
improvement would be more significant with better questions.  Since
the questions asked are superficial, they are not able to take full
advantage of the added information provided by the different
derivations.

The EXTOP5 and EXTOP20 results point out one drawback of the
derivation model.  Because the space of all possible parses is so many
orders of magnitude smaller than the space of all possible derivations
of all possible parses, the stack decoding search algorithm prunes
away the correct parse more often with the derivation model than
without it.

\subsection{Experiment B: No Conjunction Feature}

\begin{table}[tbhp]
\begin{center}
\begin{tabular}{|l|r|r|r|r|r|}
\hline
Experiment & EXACT & EXNOTAG & EXTOP5 & EXTOP20 & Perp.\\
\hline
B         & 34.2\% & 41.6\% & 48.5\% & 59.2\% & 1555\\
\hline
\end{tabular}
\end{center}
\caption[Results of experiment B.]{Results of experiment B (no conjunction
feature).\label{Btable}}
\end{table}

\begin{table}[tbhp]
\begin{center}
\begin{tabular}{|c|c|c|c|}
\hline
Model $X$ & Model $Y$ & $p(X=Y)$ & $p(X\simeq~Y)$\\
\hline
B & 1.0 & 0.005 & 0.001\\
\hline
\end{tabular}
\end{center}
\caption[Significance analysis of results from experiment
B]{Significance analysis of EXACT and EXNOTAG results from experiment
B.\label{SIGBtable}}
\end{table}

Experiment B illustrates the improvements achieved by implementing the
conjunction feature.  The conjunction feature was a response to the
poor performance of a previous version of SPATTER on sentences
with conjunctions.

While the parser's performance on sentences with conjunctions improves
with the conjunction feature, a significant percentage of sentences
with conjoined phrases are still misanalyzed.  Table~\ref{CONJtable}
shows the percentage of conjoined phrases which are correctly
identified by SPATTER using the conjunction feature.  These results
are from a hand analysis of 250 test sentences.

\begin{table}[htbp]
\begin{center}
\begin{tabular}{|l|r|r|r|r||r|}
\hline
Conjoined Phrase & Noun & Verb & Sentence & Other & Total\\
\hline
Correct/Total & 22/39 & 17/24 & 7/17 & 2/4 & 48/84\\
\% Correct & 56 & 71 & 41 & 50 & 57\\
\hline
\end{tabular}
\end{center}
\caption[Conjunction error analysis.]{Percentage of conjoined phrases
correctly identified by SPATTER using the conjunction
feature.\label{CONJtable}}
\end{table}

Most of the conjunction successes were in sentences where there was
very little or no local ambiguity.  When the conjoined phrases are
long and there are nearby phrases that seem reasonable to conjoin,
then SPATTER usually strongly favors the wrong attachment decision.
This behavior is symptomatic of the failure of SPATTER to capture some
classes of long distance dependencies.  This is probably due to the
simplicity and local nature of the decision tree questions asked.

\subsection{Experiments C and D: Pruning the Decision Trees}

\begin{table}[tbhp]
\begin{center}
\begin{tabular}{|l|r|r|r|r|r|}
\hline
Experiment & EXACT & EXNOTAG & EXTOP5 & EXTOP20 & Perp.\\
\hline
C         & 36.8\% & 44.5\% & 50.5\% & 59.9\% & 1085\\
D         & 37.0\% & 45.8\% & 49.3\% & 60.0\% & 1023\\
\hline
\end{tabular}
\end{center}
\caption[Results of experiments C and D.]{Results of experiments C and
D (pruning decision trees to 1 and 5 bit-events of
significance).\label{CDtable}}
\end{table}

\begin{table}[tbhp]
\begin{center}
\begin{tabular}{|c|c|c|c|}
\hline
Model $X$ & Model $Y$ & $p(X=Y)$ & $p(X\simeq~Y)$\\
\hline
C & 1.9 & 0.14 & 0.19\\
C & 2.9 & 0.21 & 0.25\\
D & 1.9 & 0.22 & 0.35\\
D & 2.9 & 0.29 & 0.31\\
\hline
\end{tabular}
\end{center}
\caption[Significance analysis of results from
experiment C and D]{Significance analysis of EXACT and EXNOTAG results
from experiment C and D.\label{SIGCDtable}}
\end{table}

Experiments C and D explore the effect of pruning the decision trees
on performance.  In experiment C, the decision trees are pruned to 1
bit-event of significance, and in experiment D, to 5 bit-events.  The
results show slight but not statistically significant degradations in
performance with the pruned models.  However, the perplexities of
these models are much higher, indicating that the fully grown trees
provide better models than the pruned trees.

This result is somewhat surprising, since nodes which are pruned in C
and D are based on splits which are not statistically significant.
The smoothing appears be accomplishing its goal of statistically
pruning the children of those low-count nodes which should not have
been split, and giving weight to the children of those which were
correctly split.

\subsection{Experiment E: Training on Half the Data}

\begin{table}[tbhp]
\begin{center}
\begin{tabular}{|l|r|r|r|r|r|}
\hline
Experiment & EXACT & EXNOTAG & EXTOP5 & EXTOP20 & Perp.\\
\hline
E         & 32.5\% & 39.9\% & 48.4\% & 60.2\% & 122\\
\hline
\end{tabular}
\end{center}
\caption[Results of experiment E.]{Results of experiment E (trained with 1/2
training
data).\label{Etable}}
\end{table}

\begin{table}[tbhp]
\begin{center}
\begin{tabular}{|c|c|c|c|}
\hline
Model $X$ & Model $Y$ & $p(X=Y)$ & $p(X\simeq~Y)$\\
\hline
E & 1.0 & $10^{-8}$ & $10^{-9}$\\
\hline
\end{tabular}
\end{center}
\caption[Significance analysis of results from experiment
E.]{Significance analysis of EXACT and EXNOTAG results from experiment
E.\label{SIGEtable}}
\end{table}

Experiment E measures the impact of cutting the training data size in
half.  The performance of the parser degrades significantly, lending
credence to the hypothesis that there is not enough training data to
train the SPATTER models with the current questions.  It would be
informative to see the results of training on twice as much data, but
there is not yet a large enough treebank to try this experiment.

\subsection{Experiments F and G: Parsing from Tags}

\begin{table}[tbhp]
\begin{center}
\begin{tabular}{|l|r|r|r|r|r|r|}
\hline
Experiment & EXACT & EXNOTAG & EXTOP5 & EXTOP20 & Perp.\\
\hline
F         & 46.2\% & 46.2\% & 64.8\% & 74.2\% & 81\\
G         & 50.8\% & 50.8\% & 68.2\% & 78.3\% & 54\\
\hline
\end{tabular}
\end{center}
\caption[Results of experiments F and G.]{Results of experiments F (parsing
from words and correct
tags) and G (parsing from correct tags only).\label{FGtable}}
\end{table}

\begin{table}[tbhp]
\begin{center}
\begin{tabular}{|c|c|c|c|}
\hline
Model $X$ & Model $Y$ & $p(X=Y)$ & $p(X\simeq~Y)$\\
\hline
F & 1.0 & $10^{-14}$ & 0.12\\
G & 1.0 & $10^{-22}$ & $10^{-5}$\\
\hline
\end{tabular}
\end{center}
\caption[Significance analysis of results from experiments F and
G.]{Significance analysis of EXACT and EXNOTAG results from
experiments F and G.\label{SIGFGtable}}
\end{table}

Experiments F and G mimic the experiments performed by Brill
\cite{Brill} and by Bod \cite{Bod}, respectively.  Experiment F parses
assuming the words in the sentence and their correct tags are known.
Experiment G parses assuming the correct tags are known, but the
decision trees are not permitted to ask about the words in the
sentence.

The very surprising result here is that the parser performs
significantly better when it is not allowed to ask about the words in
the sentence!  This seems to go against intuition.

But this result is not saying that lexical information is unimportant
in parsing.  Parsing from correct tags is an artificial problem.
Determining the correct tag for each word is a significant part of the
disambiguation problem.  Once the part-of-speech tags are determined,
the statistical algorithms train better when the input is tags, since
the space of inputs is much smaller.  But the words would certainly be
needed to determine the correct part-of-speech tags.

Another possible explanation for these results is that the binary
representations for words, which are determined by word bigram mutual
information, do not provide sufficiently informative questions for the
decision tree growing algorithms.  Thus, by giving the decision trees
access to the word bits, the models are overtrained.

It is also surprising that the EXNOTAG performance does not improve
significantly when the correct tag is known.  This suggests that if
the parser does not get a sentence correct, it would not have gotten
the sentence correct even if it had known the correct tag for each
word.

\subsection{Experiments H and I: Using a Flexible Tag Dictionary}

\begin{table}[tbhp]
\begin{center}
\begin{tabular}{|l|c|c|c|c|r|}
\hline
Experiment & EXACT & EXNOTAG & EXTOP5 & EXTOP20 & Tag Error\\
\hline
1.0       & 37.5\% & 45.6\% & 50.2\% & 59.4\% & 3.1\%\\
H         & 36.5\% & 45.5\% & 49.6\% & 59.1\% & 3.5\%\\
I         & 36.5\% & 45.7\% & 49.5\% & 58.3\% & 3.6\%\\
\hline
\end{tabular}
\end{center}
\caption[Results of experiments H and I.]{Results of experiments H and
I (allowing top 1 and top 5 highest probability tags), including
part-of-speech tagging error rate.\label{HItable}}
\end{table}

\begin{table}[tbhp]
\begin{center}
\begin{tabular}{|c|c|c|c|}
\hline
Model $X$ & Model $Y$ & $p(X=Y)$ & $p(X\simeq~Y)$\\
\hline
H & 1.0 & 0.0005  & 0.31\\
I & 1.0 & 0.001   & 0.5\\
\hline
\end{tabular}
\end{center}
\caption[Significance analysis of results from experiments H and
I.]{Significance analysis of EXACT and EXNOTAG results from
experiments H and I.\label{SIGHItable}}
\end{table}

Experiments H and I examines the trade-off created by using the tag
dictionary.  If the tag dictionary were complete, it would be
appropriate to allow only tags listed as legal for a word.  However,
since the tag dictionary is very crude, automatically generated by the
word-tag pairs in the training data, it could prevent the correct
analysis from being generated, regardless of the probability assigned
that analysis.  The point of this experiment is to determine if this
phenomenon is occurring.

Using the flexible tag dictionary, there is a significant difference
in the EXACT score, but a much less significant difference in the
EXNOTAG score.  Also, the tagging error rate increases when the
flexible tag dictionary is used (see Table~\ref{HItable}).  These
results suggest that the flexible tag dictionary is allowing the
incorrect tag more often than it is compensating for a gap in the tag
dictionary.  Based on this result, it is best to use the tag
dictionary.  Nonetheless, it would be a simple task to improve the tag
dictionary, either using more tagged data or using an on-line
dictionary as a knowledge source.

\section{Summary of Results}

It is difficult to draw conclusions from these experiments.  The
differences in performance among most of the parsing models were not
statistically significant.  Small but measurable improvements were
achieved by adding the conjunction feature model and the derivation
model, but variations in the decision tree stopping rule and the
configuration of the tag dictionary did not affect the parsing
results.  This suggests that the arbitrary decisions made in
configuring the parser are less important than improvements in the
modeling decisions.  However, some of the results might have been
different if the models had been trained to convergence.

While experiments show that the derivational model improves
performance, it seems that its power is not fully exploited by the
questions being asked.  But there are no methods yet for quantifying
the overall value of a question in a decision tree, so it would be
difficult to evaluate the different types of questions objectively.
It also seems likely that the derivational model cannot be completely
self-organized.  Since it is started using a uniform distribution, it
has complete freedom to move in any direction in the space of models
which will increase the probability of the corpus.  Self-organized
models rarely approach global optima without an informed starting
point, and this is probably true of the hidden derivational model.

Other conclusions one can draw from these experiments are that more
training data and more linguistically motivated questions might
improve the results.  Of course these observations are no more than
rules of thumb when using these statistical methods and decision
trees.

In trying to compare SPATTER's performance to the state-of-the-art, it
is hard to identify the best results in parsing. Some of the best
parsers are incorporated into good natural language processing
systems, where the boundary between parsing and understanding is
blurred.  It is a serious weakness of this work that the parser is not
applied to a natural language processing task.  But based on accuracy
measures reported at the PARSEVAL meetings in 1990 and 1992, which
included many of these parsers, this work is a significant improvement
over the 30\% - 60\% sentence accuracy results using the
already-maligned crossing-brackets measure.  However, it remains to be
seen whether the skeletal treebank annotations generated by SPATTER
are as useful as the more elaborate markings of rule-based grammars.

\chapter{Open Questions\label{OPENchapter}}

This work offers far more questions than it answers.  For instance,
what can be done to improve the parse tree representations to
facilitate the statistical modeling process?  How can one eliminate or
automate the acquisition of the knowledge used in SPATTER?  And how
does one evaluate the effectiveness of the decision tree modeling
techniques employed here?

The answers to these questions elude me, but in this chapter I at
least present the issues involved and speculate on ways to pursue
them.

\section{Parse Tree Representation}

One of the early design decisions made in SPATTER was to represent the
parse tree {\em only} at a sub-constituent level.  Combinations of
decision tree questions can be used to discover information about
constituents.  But since the decision tree growing algorithm is
greedy, it fails to find informative {\em combinations} of questions
unless each question individually is very informative.

\begin{figure}[htbp]
\begin{verbatim}
Treebank:
[V press [N the [Nn carriage restore Nn] key N]
         [Ti to advance [N the paper N]
                        [P to [N the next form N] P] Ti] V]
Probability = 0.023

Parse:
press
[N the carriage N]
[V restore [Fn [N key N]
               [Ti to advance
                  [N the paper N]
                  [P to [N the next form N] P] Ti] Fn] V]
Probability = 0.029
\end{verbatim}
\caption[SPATTER parse output for a test sentence.] {The treebank
analysis and SPATTER parse output for a sentence from the computer
manuals domain.\label{TBfigure}}
\end{figure}

SPATTER occasionally generates nonsensical constituents in order to
make a context look like a familiar history.  Figure~\ref{TBfigure}
gives an example of this.  SPATTER correctly identifies the
infinitival phrase in the sentence, but it misidentifies the word
``restore'' as a verb.  In order to accommodate this interpretation, it
labels the noun phrase ``key'' followed by the infinitival phrase as a
relative clause!  Although the decisions which construct this phrase
have low probability, their combined probability is not low enough to
overcome the low probability of the correct part-of-speech tag
assignment.  As indicated in the figure, SPATTER finds both analyses
but assigns the incorrect one a higher probability.

SPATTER might be improved by representing constituents, but how should
this be implemented?

One solution might be to model multiple levels of constituent
structure, as Bod does in \cite{Bod}.  The main problem with this
approach is that, given the limited amount of training data, it is
likely that this approach would overtrain on the constituents in the
treebank, and closely mimic a P-CFG.  This proposals is difficult to
pursue because there is too little training data to model the space of
subtrees accurately.

Another possibility is to try to discover automatically the classes of
subtrees which are not modeled well by the decision trees, and to
model them using other statistical methods.  These subtrees could be
identified by generalizing those contexts in the treebank which have
high entropy according to the decision tree models.

\section{Knowledge Engineering}

One of the principles of this work is that manual knowledge engineering
should be kept at a minimum.  The majority of the knowledge for
parsing should be encoded in the treebank.  The pattern recognition
training algorithms should extract whatever knowledge is needed from
the data.

In general, this principle was adhered to in the implementation of
SPATTER; however, there were a few exceptions.  The Tree Head Table is
a blatant violation of this principle.  The development of this
knowledge base was automated as much as possible, using the treebank
constituents to propose candidates and to verify the completeness of
this table so that every constituent is assigned a head.  However, it
would be preferable to acquire this knowledge completely automatically
from the treebank.

It might be possible to acquire the information in this table by
implementing it as another hidden component of the model.  In other
words, instead of using a table to select the head word from the
children of a constituent, each possible head word could be selected
according to some probability distribution, and all paths could be
pursued.

There are two problems with this approach.  Allowing the parser to
select a constituent's head word from any of its children
significantly increases the size of the search space.  It also
increases the number of parameters in SPATTER's models.  Since there
is too little training data to train the model as it stands,
increasing the size of the model without increasing the size of the
training set is unlikely to improve matters.

The hand-coded binary classification trees used as decision tree
questions are more examples of manual knowledge engineering in
SPATTER.  This information could probably be acquired using
statistical methods similar to the bigram mutual information
clustering used to discover the word classes.  However, it is not
clear what measures to use to effectively cluster the objects, such as
extensions and constituent labels, in a parse tree.

\section{Statistical Decision Tree Modeling}

There are a number of open questions in the area of statistical
decision tree modeling.  The need for improvement in the decision tree
algorithms, especially the smoothing, has already been discussed in
detail in section~\ref{DTREEproblems}.  A problem which has not yet
been discussed is that of quantifying the value of decision tree
questions.

At a given node in the decision tree, a question can be evaluated by
the entropy reduction achieved by asking that question.  But what is
the {\em overall} value of a question?  SPATTER {\em allows} questions
about any node in the parse tree, even any combination of nodes.  But
only a tiny fraction of the possible questions are actually considered
because of computational constraints of current machines.  It would be
useful to be able to rank candidate questions in order to eliminate
worthless questions and replace them with more useful ones.

One measure of the usefulness of a question is the total entropy
reduction achieved by the question, combining the incremental
reductions from all of the nodes at which the question was selected.
It would be more useful, however, to find a measure which does not
depend on a fully grown decision tree.

\chapter{Conclusions\label{CONCchapter}}

In this dissertation, I have presented a first attempt at statistical
decision tree parsing.  By some measures, in terms of both technology
and performance, it is an improvement over the state-of-the-art in
parsing.

The state-of-the-art in statistical parsing technology includes P-CFGs
trained using the Inside-Outside algorithm (Schabes and Pereira
\cite{Schabes:92}, Kupiec \cite{Kupiec}, and Black, Garside, and
Leech\cite{blackbook}, parsers which generate unlabeled bracketing
using correct tag sequences as input (Brill \cite{Brill}, Schabes and
Pereira \cite{Schabes:92}, and grammar induction strategies which
attempt to acquire grammars by extracting context-free productions
from treebanks.  None of these parsing techniques considers lexical
information in its models, with the exception of probabilistic
lexicalized tree-adjoining grammar (Schabes and Waters\cite{Schabes}),
which has yet to be implemented and tested on a large scale.  In
contrast, the SPATTER parser incorporates into its models as much
lexical information as the decision tree algorithms deem useful, and
uses a hidden derivational model to maximize the amount of information
available to make the more difficult disambiguation decisions.  The
parser considers an immense space of possible parses, and uses the
stack decoding algorithm from speech recognition to search this space.

It must be stressed that this work is only a {\em first
attempt} at applying speech recognition technology to the natural
language parsing problem.  The natural language processing community
must distance itself from the toy problems it has addressed in the
past.  Parsing technology has improved to the point that it can and
should be evaluated on stricter measures.  Evaluating unlabeled
structure or structure generated from tagged data should no longer be
considered acceptable.  Parsers are capable of analyzing and labeling
syntactic structure at a reasonable accuracy rate.  Reporting results
on toy experiments only serves to mislead the community.

The speech recognition community has demonstrated that solutions to
difficult problems can be found by addressing a real problem, and not
by creating an artificial task and solving it instead.  Only by
formalizing the parsing problem and agreeing to use objective measures
of progress will the parsing community make progress on the parsing
problem.  The treebank recognition problem and the exact match
criterion represent first steps in this direction.

\appendix
\chapter{Training and Test Entropies\label{ENTROPYappendix}}

Appendix~\ref{ENTROPYappendix} contains training and test entropies
for all of the models described in Chapter~\ref{RESULTchapter}.  The
training entropies reported here are the experimental entropies after
each iteration of the forward-backward algorithm.  The experimental
entropies are estimated based on 2,800 sentences.  These sentences
were {\em not} among those sentences used in growing the decision tree
models.  The test entropies are estimated based on 100 sentences which
were not used in growing or smoothing the models.

Following the tables of training and test entropies, there are tables
reporting the perplexity estimates for each component model (label,
tag, extension, and conjunction).  These perplexity estimates are
based on the same 2,800 sentences from which the entropy results were
generated.

\begin{table}
\begin{center}
\begin{tabular}{|r|r|r|r|r|}
\hline
Iter. & 1.0 & 1.9 & 2.0 & 2.9 \\
\hline
 1 & 2692.4 & 2957.5 & 2540.3 & 2901.2 \\
 2 & 362.1 & 384.4 & 354.4 & 375.7 \\
 3 & 233.8 & 247.1 & 230.1 & 242.2 \\
 4 & 195.1 & 205.5 & 192.7 & 202.3 \\
 5 & 177.1 & 186.1 & 175.2 & 183.8 \\
 6 & 166.8 & 174.9 & 165.3 & 173.2 \\
 7 & 160.1 & 167.6 & 158.9 & 166.4 \\
 8 & 155.4 & 162.4 & 154.4 & 161.5 \\
 9 & 151.9 & 158.5 & 151.0 & 158.0 \\
10 & 149.1 & 155.5 & 148.5 & 155.2 \\
11 & 147.0 & 153.1 & 146.4 & 153.0 \\
12 & 145.3 & 151.2 & 144.8 & 151.3 \\
13 & 143.9 & 149.6 & 143.5 & 149.8 \\
14 & 142.8 & 148.3 & 142.4 & 148.6 \\
15 & 141.8 & 147.2 & 141.4 & 147.6 \\
16 & 141.1 & 146.3 & 140.6 & 146.7 \\
17 & 140.4 & 145.5 & 140.0 & 145.9 \\
18 & 139.8 & 144.8 & 139.3 & 145.2 \\
19 & 139.2 & 144.2 & 138.8 & 144.6 \\
20 & 138.8 & 143.7 & 138.3 & 144.1 \\
\hline
Test & 1259.1 & 1204.7 & 917.2 & 1046.9 \\
\hline
\end{tabular}
\end{center}
\caption[Entropies for smoothing models 1.0, 1.9, 2.0, and
2.9.]{Training entropies after each iteration of smoothing algorithm
for basic configuration models 1.0, 1.9, 2.0, and 2.9.  Smoothing
entropies are based on 2,800 sentences.  Last row contains test
entropies, based on 100 sentences.}
\end{table}

\begin{table}
\begin{center}
\begin{tabular}{|r|r|r|r|r|r|r|r|}
\hline
Iter. & A & B & C & D & E & F & G \\
\hline
 1 & 2699.6 & 1267.3 & 1723.7 & 1400.9 & 32592.7 & 65.1 & 50.4 \\
 2 & 648.0 & 262.3 & 392.6 & 392.9 & 3008.3 & 36.6 & 26.5 \\
 3 & 420.5 & 167.5 & 255.1 & 267.8 & 1590.4 & 30.2 & 22.6 \\
 4 & 340.6 & 136.0 & 209.0 & 222.9 & 1192.3 & 27.4 & 21.2 \\
 5 & 300.3 & 120.6 & 186.6 & 200.2 & 1012.2 & 25.9 & 20.6 \\
 6 & 276.1 & 111.6 & 173.6 & 186.5 & 911.1 & 25.0 & 20.2 \\
 7 & 259.9 & 105.7 & 165.0 & 177.4 & 847.2 & 24.4 & 19.9 \\
 8 & 248.5 & 101.5 & 159.0 & 171.0 & 803.5 & 24.0 & 19.8 \\
 9 & 240.0 & 98.4 & 154.6 &  166.1 & 771.9 & 23.8 & 19.6 \\
10 & 233.5 & 95.9 & 151.1 & 162.4 & 748.2 & 23.5 & 19.5 \\
11 & 228.4 & 94.0 & 148.4 & 159.4 & 730.0 & 23.4 & 19.4 \\
12 & 224.3 & 92.5 & 146.2 & 157.0 & 715.9 & 23.2 & 19.4 \\
13 & 221.0 & 91.2 & 144.4 & 155.0 & 704.6 & 23.1 & 19.3 \\
14 & 218.3 & 90.3 & 142.8 & 153.3 & 695.4 & 23.0 & 19.2 \\
15 & 216.1 & 89.5 & 141.5 & 151.9 & 687.7 & 22.9 & 19.2 \\
16 & 214.2 & 88.8 & 140.4 & 150.6 & 681.1 & 22.9 & 19.2 \\
17 & 212.6 & 88.2 & 139.5 & 149.5 & 675.4 & 22.8 & 19.1 \\
18 & 211.3 & 87.7 & 138.7 & 148.6 & 670.4 & 22.7 & 19.1 \\
19 & 210.1 & 87.3 & 138.0 & 147.8 & 666.0 & 22.7 & 19.1 \\
20 & 209.1 & 86.9 & 137.4 & 147.1 & 662.1 & 22.7 & 19.1 \\
\hline
Test & 1430.9 & 829.0 & 781.2 & 891.1 & 5986.4 & 87.6 & 79.1\\
\hline
\end{tabular}
\end{center}
\caption[Entropies for smoothing models from experiments A -
G.]{Training entropies after each iteration of smoothing algorithm for
models A - G.  Smoothing entropies are based on 2,800 sentences.  Last
row contains test entropies, based on 100 sentences.}
\end{table}

\begin{table}
\begin{center}
\begin{tabular}{|r|r|r|r|r|}
\hline
Iter. & 1.0 & 1.9 & 2.0 & 2.9 \\
\hline
 1 & 1.07741 & 1.07603 & 1.07745 & 1.07654 \\
 2 & 1.05061 & 1.04988 & 1.05111 & 1.05051 \\
 3 & 1.04592 & 1.04510 & 1.04642 & 1.04584 \\
 4 & 1.04420 & 1.04334 & 1.04468 & 1.04412 \\
 5 & 1.04324 & 1.04238 & 1.04370 & 1.04318 \\
 6 & 1.04260 & 1.04175 & 1.04304 & 1.04255 \\
 7 & 1.04215 & 1.04130 & 1.04257 & 1.04210 \\
 8 & 1.04181 & 1.04097 & 1.04222 & 1.04176 \\
 9 & 1.04155 & 1.04072 & 1.04195 & 1.04150 \\
10 & 1.04134 & 1.04052 & 1.04173 & 1.04129 \\
11 & 1.04117 & 1.04035 & 1.04156 & 1.04112 \\
12 & 1.04104 & 1.04022 & 1.04141 & 1.04098 \\
13 & 1.04092 & 1.04011 & 1.04129 & 1.04086 \\
14 & 1.04082 & 1.04001 & 1.04119 & 1.04076 \\
15 & 1.04074 & 1.03992 & 1.04110 & 1.04068 \\
16 & 1.04066 & 1.03985 & 1.04102 & 1.04060 \\
17 & 1.04060 & 1.03979 & 1.04096 & 1.04054 \\
18 & 1.04054 & 1.03973 & 1.04090 & 1.04048 \\
19 & 1.04049 & 1.03968 & 1.04085 & 1.04043 \\
20 & 1.04044 & 1.03964 & 1.04080 & 1.04038 \\
\hline
\end{tabular}
\end{center}
\caption[Perplexities for label model from basic
experiment.]{Perplexities for label model after each iteration of
smoothing basic configuration models 1.0, 1.9, 2.0, and 2.9.}
\end{table}

\begin{table}
\begin{center}
\begin{tabular}{|r|r|r|r|r|}
\hline
Iter. & 1.0 & 1.9 & 2.0 & 2.9 \\
\hline
 1 & 1.23947 & 1.24896 & 1.23740 & 1.24695 \\
 2 & 1.14769 & 1.15204 & 1.14651 & 1.15089 \\
 3 & 1.12747 & 1.13140 & 1.12651 & 1.13048 \\
 4 & 1.11891 & 1.12265 & 1.11806 & 1.12186 \\
 5 & 1.11425 & 1.11790 & 1.11347 & 1.11719 \\
 6 & 1.11130 & 1.11489 & 1.11059 & 1.11425 \\
 7 & 1.10922 & 1.11277 & 1.10858 & 1.11219 \\
 8 & 1.10766 & 1.11118 & 1.10708 & 1.11065 \\
 9 & 1.10644 & 1.10995 & 1.10592 & 1.10945 \\
10 & 1.10546 & 1.10897 & 1.10500 & 1.10850 \\
11 & 1.10467 & 1.10818 & 1.10425 & 1.10773 \\
12 & 1.10403 & 1.10754 & 1.10364 & 1.10710 \\
13 & 1.10351 & 1.10703 & 1.10314 & 1.10659 \\
14 & 1.10309 & 1.10662 & 1.10273 & 1.10616 \\
15 & 1.10275 & 1.10629 & 1.10238 & 1.10581 \\
16 & 1.10246 & 1.10603 & 1.10208 & 1.10551 \\
17 & 1.10221 & 1.10580 & 1.10182 & 1.10525 \\
18 & 1.10199 & 1.10561 & 1.10159 & 1.10502 \\
19 & 1.10181 & 1.10545 & 1.10138 & 1.10481 \\
20 & 1.10164 & 1.10531 & 1.10120 & 1.10464 \\
\hline
\end{tabular}
\end{center}
\caption[Perplexities for tag model from basic
experiment.]{Perplexities for tag model after each iteration of
smoothing basic configuration models 1.0, 1.9, 2.0, and 2.9.}
\end{table}

\begin{table}
\begin{center}
\begin{tabular}{|r|r|r|r|r|}
\hline
Iter. & 1.0 & 1.9 & 2.0 & 2.9 \\
\hline
 1 & 1.19404 & 1.19691 & 1.18946 & 1.19575 \\
 2 & 1.15802 & 1.16035 & 1.15489 & 1.15847 \\
 3 & 1.15085 & 1.15290 & 1.14803 & 1.15094 \\
 4 & 1.14807 & 1.14995 & 1.14543 & 1.14805 \\
 5 & 1.14669 & 1.14842 & 1.14416 & 1.14661 \\
 6 & 1.14592 & 1.14751 & 1.14345 & 1.14579 \\
 7 & 1.14544 & 1.14690 & 1.14301 & 1.14529 \\
 8 & 1.14512 & 1.14648 & 1.14273 & 1.14495 \\
 9 & 1.14489 & 1.14616 & 1.14254 & 1.14472 \\
10 & 1.14472 & 1.14591 & 1.14240 & 1.14454 \\
11 & 1.14459 & 1.14572 & 1.14229 & 1.14440 \\
12 & 1.14449 & 1.14556 & 1.14220 & 1.14428 \\
13 & 1.14441 & 1.14542 & 1.14213 & 1.14418 \\
14 & 1.14434 & 1.14530 & 1.14208 & 1.14409 \\
15 & 1.14428 & 1.14520 & 1.14203 & 1.14402 \\
16 & 1.14423 & 1.14511 & 1.14199 & 1.14395 \\
17 & 1.14419 & 1.14503 & 1.14195 & 1.14389 \\
18 & 1.14415 & 1.14496 & 1.14192 & 1.14384 \\
19 & 1.14411 & 1.14489 & 1.14189 & 1.14379 \\
20 & 1.14408 & 1.14484 & 1.14187 & 1.14375 \\
\hline
\end{tabular}
\end{center}
\caption[Perplexities for extension model from basic
experiment.]{Perplexities for extension model after each iteration of
smoothing basic configuration models 1.0, 1.9, 2.0, and 2.9.}
\end{table}

\begin{table}
\begin{center}
\begin{tabular}{|r|r|r|r|r|}
\hline
Iter. & 1.0 & 1.9 & 2.0 & 2.9 \\
\hline
 1 & 1.03001 & 1.02985 & 1.03111 & 1.03140 \\
 2 & 1.02469 & 1.02489 & 1.02617 & 1.02699 \\
 3 & 1.02316 & 1.02333 & 1.02465 & 1.02548 \\
 4 & 1.02247 & 1.02264 & 1.02394 & 1.02478 \\
 5 & 1.02209 & 1.02226 & 1.02356 & 1.02441 \\
 6 & 1.02185 & 1.02203 & 1.02332 & 1.02419 \\
 7 & 1.02168 & 1.02187 & 1.02317 & 1.02405 \\
 8 & 1.02155 & 1.02175 & 1.02306 & 1.02395 \\
 9 & 1.02145 & 1.02166 & 1.02298 & 1.02388 \\
10 & 1.02138 & 1.02157 & 1.02292 & 1.02383 \\
11 & 1.02131 & 1.02150 & 1.02288 & 1.02379 \\
12 & 1.02126 & 1.02142 & 1.02285 & 1.02376 \\
13 & 1.02121 & 1.02135 & 1.02282 & 1.02373 \\
14 & 1.02118 & 1.02128 & 1.02280 & 1.02371 \\
15 & 1.02114 & 1.02123 & 1.02279 & 1.02369 \\
16 & 1.02111 & 1.02118 & 1.02278 & 1.02367 \\
17 & 1.02109 & 1.02115 & 1.02277 & 1.02365 \\
18 & 1.02107 & 1.02112 & 1.02277 & 1.02364 \\
19 & 1.02105 & 1.02109 & 1.02277 & 1.02362 \\
20 & 1.02103 & 1.02107 & 1.02276 & 1.02361 \\
\hline
\end{tabular}
\end{center}
\caption[Perplexities for conjunction model from basic
experiment.]{Perplexities for conjunction model after each iteration
of smoothing basic configuration models 1.0, 1.9, 2.0, and 2.9.}
\end{table}

\begin{table}
\begin{center}
\begin{tabular}{|r|r|r|r|r|r|r|r|}
\hline
Iter. & A & B & C & D & E & F & G \\
\hline
 1 & 1.08095 & 1.07683 & 1.07939 & 1.07889 & 1.11101 & 1.07883 & 1.06071 \\
 2 & 1.05842 & 1.05658 & 1.06033 & 1.06027 & 1.07719 & 1.05938 & 1.04427 \\
 3 & 1.05344 & 1.05158 & 1.05548 & 1.05557 & 1.06843 & 1.05427 & 1.04069 \\
 4 & 1.05139 & 1.04948 & 1.05334 & 1.05350 & 1.06495 & 1.05194 & 1.03926 \\
 5 & 1.05016 & 1.04837 & 1.05208 & 1.05228 & 1.06316 & 1.05060 & 1.03848 \\
 6 & 1.04931 & 1.04767 & 1.05125 & 1.05147 & 1.06208 & 1.04973 & 1.03798 \\
 7 & 1.04866 & 1.04719 & 1.05067 & 1.05088 & 1.06137 & 1.04913 & 1.03763 \\
 8 & 1.04817 & 1.04684 & 1.05025 & 1.05045 & 1.06086 & 1.04869 & 1.03736 \\
 9 & 1.04778 & 1.04659 & 1.04993 & 1.05013 & 1.06048 & 1.04836 & 1.03716 \\
10 & 1.04747 & 1.04639 & 1.04969 & 1.04988 & 1.06018 & 1.04811 & 1.03701 \\
11 & 1.04721 & 1.04623 & 1.04949 & 1.04967 & 1.05995 & 1.04791 & 1.03688 \\
12 & 1.04700 & 1.04610 & 1.04933 & 1.04951 & 1.05975 & 1.04774 & 1.03677 \\
13 & 1.04682 & 1.04599 & 1.04920 & 1.04937 & 1.05959 & 1.04760 & 1.03668 \\
14 & 1.04667 & 1.04589 & 1.04909 & 1.04925 & 1.05944 & 1.04748 & 1.03660 \\
15 & 1.04654 & 1.04581 & 1.04899 & 1.04915 & 1.05932 & 1.04738 & 1.03654 \\
16 & 1.04642 & 1.04574 & 1.04890 & 1.04906 & 1.05921 & 1.04729 & 1.03648 \\
17 & 1.04632 & 1.04568 & 1.04882 & 1.04899 & 1.05910 & 1.04721 & 1.03642 \\
18 & 1.04623 & 1.04562 & 1.04876 & 1.04892 & 1.05901 & 1.04715 & 1.03638 \\
19 & 1.04615 & 1.04557 & 1.04870 & 1.04885 & 1.05893 & 1.04708 & 1.03634 \\
20 & 1.04608 & 1.04552 & 1.04864 & 1.04880 & 1.05885 & 1.04703 & 1.03630 \\
\hline
\end{tabular}
\end{center}
\caption[Perplexities for label model from experiments A - G.]{Perplexities for
label model
after each iteration of smoothing models from experiments A - G.}
\end{table}

\begin{table}
\begin{center}
\begin{tabular}{|r|r|r|r|r|r|r|r|}
\hline
Iter. & A & B & C & D & E & F & G \\
\hline
 1 & 1.25801 & 1.24207 & 1.24796 & 1.24717 & 1.38945 & 1 & 1 \\
 2 & 1.18214 & 1.16090 & 1.16551 & 1.16730 & 1.26025 & 1 & 1 \\
 3 & 1.16208 & 1.14017 & 1.14443 & 1.14642 & 1.22876 & 1 & 1 \\
 4 & 1.15296 & 1.13089 & 1.13505 & 1.13703 & 1.21452 & 1 & 1 \\
 5 & 1.14772 & 1.12564 & 1.12976 & 1.13169 & 1.20629 & 1 & 1 \\
 6 & 1.14430 & 1.12222 & 1.12632 & 1.12822 & 1.20094 & 1 & 1 \\
 7 & 1.14186 & 1.11979 & 1.12389 & 1.12576 & 1.19722 & 1 & 1 \\
 8 & 1.14003 & 1.11796 & 1.12206 & 1.12391 & 1.19447 & 1 & 1 \\
 9 & 1.13861 & 1.11651 & 1.12063 & 1.12247 & 1.19235 & 1 & 1 \\
10 & 1.13748 & 1.11534 & 1.11949 & 1.12133 & 1.19068 & 1 & 1 \\
11 & 1.13657 & 1.11438 & 1.11856 & 1.12040 & 1.18937 & 1 & 1 \\
12 & 1.13584 & 1.11360 & 1.11779 & 1.11963 & 1.18835 & 1 & 1 \\
13 & 1.13523 & 1.11297 & 1.11713 & 1.11898 & 1.18753 & 1 & 1 \\
14 & 1.13474 & 1.11245 & 1.11657 & 1.11842 & 1.18686 & 1 & 1 \\
15 & 1.13433 & 1.11203 & 1.11610 & 1.11794 & 1.18630 & 1 & 1 \\
16 & 1.13398 & 1.11167 & 1.11570 & 1.11753 & 1.18582 & 1 & 1 \\
17 & 1.13369 & 1.11137 & 1.11537 & 1.11718 & 1.18540 & 1 & 1 \\
18 & 1.13343 & 1.11112 & 1.11509 & 1.11688 & 1.18503 & 1 & 1 \\
19 & 1.13321 & 1.11090 & 1.11485 & 1.11662 & 1.18471 & 1 & 1 \\
20 & 1.13301 & 1.11070 & 1.11464 & 1.11639 & 1.18442 & 1 & 1 \\
\hline
\end{tabular}
\end{center}
\caption[Perplexities for tag model from experiments A - G.]{Perplexities for
tag model after each iteration of
smoothing models from experiments A - G.}
\end{table}

\begin{table}
\begin{center}
\begin{tabular}{|r|r|r|r|r|r|r|r|}
\hline
Iter. & A & B & C & D & E & F & G \\
\hline
 1 & 1.20265 & 1.18199 & 1.19157 & 1.18310 & 1.26553 & 1.17672 & 1.15695 \\
 2 & 1.18077 & 1.15187 & 1.16780 & 1.16854 & 1.21722 & 1.15388 & 1.13153 \\
 3 & 1.17225 & 1.14230 & 1.15961 & 1.16253 & 1.20420 & 1.14582 & 1.12520 \\
 4 & 1.16766 & 1.13767 & 1.15568 & 1.15931 & 1.19838 & 1.14177 & 1.12277 \\
 5 & 1.16482 & 1.13503 & 1.15349 & 1.15736 & 1.19515 & 1.13934 & 1.12160 \\
 6 & 1.16292 & 1.13337 & 1.15213 & 1.15607 & 1.19312 & 1.13779 & 1.12093 \\
 7 & 1.16157 & 1.13225 & 1.15123 & 1.15517 & 1.19173 & 1.13678 & 1.12051 \\
 8 & 1.16057 & 1.13143 & 1.15059 & 1.15452 & 1.19074 & 1.13607 & 1.12021 \\
 9 & 1.15981 & 1.13079 & 1.15012 & 1.15403 & 1.19002 & 1.13555 & 1.11998 \\
10 & 1.15922 & 1.13029 & 1.14976 & 1.15364 & 1.18947 & 1.13516 & 1.11980 \\
11 & 1.15874 & 1.12988 & 1.14947 & 1.15333 & 1.18904 & 1.13486 & 1.11965 \\
12 & 1.15836 & 1.12956 & 1.14923 & 1.15307 & 1.18869 & 1.13461 & 1.11953 \\
13 & 1.15805 & 1.12931 & 1.14903 & 1.15286 & 1.18841 & 1.13441 & 1.11942 \\
14 & 1.15780 & 1.12911 & 1.14887 & 1.15267 & 1.18817 & 1.13424 & 1.11934 \\
15 & 1.15758 & 1.12895 & 1.14872 & 1.15252 & 1.18796 & 1.13409 & 1.11926 \\
16 & 1.15741 & 1.12882 & 1.14859 & 1.15238 & 1.18779 & 1.13397 & 1.11920 \\
17 & 1.15726 & 1.12871 & 1.14848 & 1.15225 & 1.18764 & 1.13386 & 1.11914 \\
18 & 1.15713 & 1.12862 & 1.14838 & 1.15215 & 1.18751 & 1.13377 & 1.11909 \\
19 & 1.15702 & 1.12853 & 1.14829 & 1.15205 & 1.18739 & 1.13369 & 1.11905 \\
20 & 1.15693 & 1.12846 & 1.14821 & 1.15196 & 1.18729 & 1.13361 & 1.11902 \\
\hline
\end{tabular}
\end{center}
\caption[Perplexities for extension model from experiments A - G.]{Perplexities
for extension model after each iteration of
smoothing models from experiments A - G.}
\end{table}

\begin{table}
\begin{center}
\begin{tabular}{|r|r|r|r|r|r|r|r|}
\hline
Iter. & A & B & C & D & E & F & G \\
\hline
 1 & 1.03453 & N/A & 1.04217 & 1.04221 & 1.04287 & 1.02959 & 1.02460 \\
 2 & 1.03255 & N/A & 1.03888 & 1.03927 & 1.03743 & 1.02590 & 1.02069 \\
 3 & 1.03158 & N/A & 1.03720 & 1.03777 & 1.03540 & 1.02474 & 1.01930 \\
 4 & 1.03099 & N/A & 1.03620 & 1.03689 & 1.03433 & 1.02410 & 1.01860 \\
 5 & 1.03059 & N/A & 1.03556 & 1.03633 & 1.03368 & 1.02370 & 1.01819 \\
 6 & 1.03031 & N/A & 1.03511 & 1.03593 & 1.03323 & 1.02343 & 1.01791 \\
 7 & 1.03010 & N/A & 1.03478 & 1.03563 & 1.03290 & 1.02324 & 1.01771 \\
 8 & 1.02993 & N/A & 1.03452 & 1.03539 & 1.03265 & 1.02309 & 1.01755 \\
 9 & 1.02980 & N/A & 1.03431 & 1.03520 & 1.03244 & 1.02296 & 1.01742 \\
10 & 1.02969 & N/A & 1.03415 & 1.03504 & 1.03227 & 1.02286 & 1.01732 \\
11 & 1.02960 & N/A & 1.03400 & 1.03490 & 1.03213 & 1.02278 & 1.01723 \\
12 & 1.02952 & N/A & 1.03388 & 1.03478 & 1.03201 & 1.02271 & 1.01716 \\
13 & 1.02945 & N/A & 1.03377 & 1.03467 & 1.03190 & 1.02265 & 1.01709 \\
14 & 1.02939 & N/A & 1.03368 & 1.03458 & 1.03181 & 1.02259 & 1.01704 \\
15 & 1.02934 & N/A & 1.03360 & 1.03449 & 1.03172 & 1.02254 & 1.01699 \\
16 & 1.02930 & N/A & 1.03352 & 1.03441 & 1.03165 & 1.02250 & 1.01694 \\
17 & 1.02925 & N/A & 1.03345 & 1.03434 & 1.03158 & 1.02246 & 1.01691 \\
18 & 1.02922 & N/A & 1.03339 & 1.03428 & 1.03152 & 1.02242 & 1.01688 \\
19 & 1.02918 & N/A & 1.03334 & 1.03422 & 1.03146 & 1.02239 & 1.01685 \\
20 & 1.02915 & N/A & 1.03329 & 1.03416 & 1.03140 & 1.02236 & 1.01683 \\
\hline
\end{tabular}
\end{center}
\caption[Perplexities for conjunction model from experiments A -
G.]{Perplexities for conjunction model after each iteration of
smoothing models from experiments A - G.}
\end{table}

\chapter{SPATTER Vocabularies and Binary Encodings\label{VOCABULARYappendix}}

The 7,655 word vocabulary and their bitstrings, and the bitstrings for
the part-of-speech tag set are available from the author upon request.
To request electronic verisions of any of the SPATTER vocabularies,
send electronic mail to magerman@cs.stanford.edu.

The descriptions of the non-terminal label set and part-of-speech tag
set can be found in Black, Garside, and Leech\cite{blackbook}.

\section{Part-of-Speech Tag Vocabulary}

Here is the part-of-speech tag vocabulary, sorted by frequency of
occurrence in the Lancaster Computer Manuals Treebank:

\begin{verbatim}
NN1 AT .  II NN2 VVC VVN JJ , VVI AT1 CC NP1 VBZ PPY TO VM CS IO IF MC
VVZ DD1 VVG RR JB : ) ( VV0 VBI XX VBR NN CST APP$ IW &FO MC1 RP PPH1
DB NNT1 DD RT MD ZZ1 VVD " DAR DDQ DA VH0 DD2 VBDZ VHZ CSA II22 II21
RRQ VHI VDZ VD0 REX22 REX21 NNJ ; NNU EX VBN CSN NNT2 UH CF VDC CCB $
JJR RR21 RL RR22 PPHS2 ? LE DA2 VBG CSW VBDR MC-MC RG PPHO2 VDI -
CS22 CS21 PNQS RG22 RG21 ZZ2 VDN DB2 II33 II32 II31 RRR VDD PN NNS1 JK
PN1 DAT VHG PPX1 ... PPHS1 NNL1 JJT RR33 RR32 RR31 DDQ$ RGR NP VBC
NNU1 NNU2 VDG VHD CSW33 CSW32 CSW31 JA DD222 DD221 CS33 CS32 CS31 VBS
BTO22 BTO21 NNJ2 MF RGQ NNL2 CC33 CC32 CC31 RRQV VHC VVGK DA1 CC22
CC21 ! RGT RGA MC2 PPHO1 PPX222 PPX221 RPK PN122 PN121 VVS RRT NNO2
BCS22 BCS21 RA VVO VVC22 VVC21 VHS VBDS VB0 PPX122 PPX121 NPM1 NP2
NNSB2 NNSA1 NNJ1 NN122 NN121 MC222 MC221 JJ33 JJ32 JJ31 JJ22 JJ21 JBT
CC23 BDT
\end{verbatim}

\section{Non-terminal Label Vocabulary}
\begin{tabular}{ll}
\hline
Non-terminal & Bitstring\\
\hline
BDL &  10000000\\
GOD &  11000000\\
\*\* & 11000001\\
P &    11000011\\
Fc & 11000110\\
J & 11001011\\
G & 11010011\\
V & 11100000\\
S & 11100010\\
Si & 11100011\\
Fa & 11100100\\
Fr & 11100110\\
Fn & 11100111\\
Ti & 11110001\\
Tn & 11110101\\
Tg & 11110110\\
Nn & 11111000\\
N & 11111010\\
Nv & 11111100\\
Nr & 11111110\\
\hline
\end{tabular}

\section{Extend Feature Vocabulary}
\begin{tabular}{ll}
\hline
Extend Type & Bitstring\\
\hline
up   &  10011\\
left &  10110\\
right & 11000\\
unary & 11101\\
root  & 01000\\
BDE   & 00100\\
\hline
\end{tabular}

\section{NumChildren Question Vocabulary}

\begin{tabular}{ll}
\hline
\# of Children & Bitstring\\
\hline
0 & 11001\\
1 & 11101\\
2 & 11111\\
3 & 10111\\
4 & 10110\\
5 & 10100\\
$>$5 & 10000\\
\hline
\end{tabular}

\section{NumNodes Question Vocabulary}

\begin{tabular}{ll}
\hline
\# of Nodes & Bitstring\\
\hline
1 & 1000\\
2 & 0101\\
3 & 0100\\
4 & 0110\\
5 & 0111\\
6-10 & 0010\\
11-20 & 0000\\
$>$20 & 1100\\
\hline
\end{tabular}

\section{Span Question Vocabulary}

\begin{tabular}{ll}
\hline
Constituent Span & Bitstring\\
\hline
1 & 1110\\
2 & 1100\\
3 & 1101\\
4 & 1001\\
5 & 1011\\
6-10 & 0111\\
11-20 & 0010\\
$>$20 & 0000\\
\hline
\end{tabular}

\chapter{SPATTER Tree Head Table\label{THTappendix}}

\begin{description}
\item[Nr] {\em right-to-left} Nr NNT1 NNT2 RR RRR DAR RT DAT DA1 NN1 NN2 MC
\item[Nv] {\em right-to-left} Nv NNU1 NNU2 NNU NN1 NN2 MC MC1 MC-MC DA DAR JJR
N P
\item[V] {\em left-to-right} V VV0 VVC VVC21 VVC22 VVD VVG VVGK VVI VVN VVO VVS
VVZ VB0 VBC VBDR VBDS VBDZ VBG VBI VBN VBR VBS VBZ VD0 VDC VDD VDG VDI
VDN VDZ VH0 VHC VHD VHG VHI VHS VHZ VM Tg Nn
\item[N] {\em right-to-left} N NN NNJ NNU NP NN2 NNJ2 NNL2 NNO2 NNSB2 NNT2 NNU2
NP2 NN1 NN121 NNJ1 NNL1 NNS1 NNSA1 NNT1 NNU1 NP1 NPM1 ZZ1 ZZ2 \&FO UH
PPY PN PN1 PN122 PNQS PPH1 PPHO1 PPHO2 PPHS1 PPHS2 PPX1 PPX122 PPX222
PPY JA JB JBT JJ JJ21 JJ32 JJR JJT JK J DA DA1 DA2 DAR DAT DB DB2 DD
DD1 DD2 DD222 DDQ DDQ\$ MC MC-MC MC1 MC2 MC222 MD MF EX Nn
\item[S] {\em right-to-left} S V Ti Tn Tg N J Fa REX22 P UH
\item[Tg] {\em right-to-left} Tg VVG VBG VDG VHG V
\item[Ti] {\em right-to-left} Ti VVI VDI VVN VDN VHI VHD VBI V TO
\item[Tn] {\em right-to-left} Tn VVN VDN VHD V
\item[Fa] {\em right-to-left} Fa CS CS21 CS31 CSA CSN CST CSW CSW31 CF CCB LE
RRQV
\item[Fc] {\em right-to-left} Fc Fa
\item[Fn] {\em right-to-left} Fn S Ti N
\item[Fr] {\em right-to-left} Fr S
\item[G] {\em left-to-right} G N
\item[J] {\em right-to-left} J JJ JB JA JJ21 JJ31 JJR JJT JK VVN
\item[P] {\em left-to-right} P II II21 II32 IO IW IF
\item[Si] {\em right-to-left} Si S
\item[Nn] {\em right-to-left} Nn VVC N II RR RRQ RP UH RL CC : "
\end{description}

\end{document}